\documentclass[journal]{IEEEtran}
\ifCLASSOPTIONcompsoc
  \usepackage[nocompress]{cite}
\else
  \usepackage{cite}
\fi
\ifCLASSINFOpdf
\else
\fi
\usepackage[utf8]{inputenc} % allow utf-8 input
\usepackage[T1]{fontenc}    % use 8-bit T1 fonts
\usepackage{url}            % simple URL typesetting
\usepackage{amsfonts}       % blackboard math symbols
\usepackage{nicefrac}       % compact symbols for 1/2, etc.
\usepackage{microtype}      % microtypography
\usepackage{graphicx}
\usepackage{mathtools}
\usepackage{multirow}
\usepackage{subfigure}
\usepackage{tabularx}
\usepackage{amsmath,amsfonts,amssymb}
\usepackage{wrapfig,lipsum,booktabs}
\usepackage{xcolor}
\usepackage[pagebackref=true,breaklinks=true,letterpaper=true,colorlinks,bookmarks=false,citecolor=cyan]{hyperref}
\usepackage{balance}
\usepackage{bbm}
\DeclareRobustCommand\onedot{\futurelet\@let@token\@onedot}
\def\@onedot{\ifx\@let@token.\else.\null\fi\xspace}

\def\eg{\textit{e.g.}}
\def\ie{\textit{i.e.}}

\def\etal{\textit{et al.}}

\usepackage{subfigure}
\usepackage{xcolor}
\usepackage{color, soul}
\usepackage{colortbl}
\usepackage{booktabs}
\usepackage{multirow}
\usepackage[colorlinks]{hyperref}

\soulregister{\cite}7
\soulregister{\ref}7

\newcommand{\stdvu}[1]{\scriptsize{\color{darkgray}(#1)}}  % 方差字体

\definecolor{mypink}{rgb}{.99,.91,.95}

\begin{document}
\title{Unprejudiced Training Auxiliary Tasks Makes Primary Better: A Multi-Task Learning Perspective}

\author{ Yuanze Li,
Chun-Mei Feng,
Qilong Wang, 
Guanglei Yang$^*$,
Wangmeng Zuo,
\IEEEcompsocitemizethanks{
\IEEEcompsocthanksitem Yuanze Li, Guanglei Yang and Wangmeng Zuo are with the School of Computer Science and Technology, Harbin Institute of Technology (HIT), Harbin, China. E-mail: \{sqlyz,yangguanglei,wmzuo\}@hit.edu.cn.\protect
\IEEEcompsocthanksitem Chun-Mei Feng are with Institute of High Performance Computing (IHPC),
Agency for Science, Technology and Research (A*STAR), Singapore. E-mail: fengcm.ai@gmail.com.\protect
\IEEEcompsocthanksitem Qilong Wang are with the College of Intelligence and Computing, Tianjin University, Tianjin 300350, China. E-mail: qlwang@tju.edu.cn.\protect
\IEEEcompsocthanksitem $^*$ Corresponding author.\protect

% $^*$Corresponding authors. \protect

%E-mail: see http://www.michaelshell.org/contact.html
%\IEEEcompsocthanksitem J. Doe and J. Doe are with Anonymous University.
}% <-this % stops an unwanted space
% \thanks{Manuscript received April 19, 2005; revised August 26, 2015.}
}


\markboth{Submitted to IEEE Transactions on  Neural Networks and Learning Systems}%
{Shell \MakeLowercase{\textit{et al.}}: Bare Demo of IEEEtran.cls for Computer Society Journals}



\IEEEtitleabstractindextext{%
\begin{abstract}

Human beings can leverage knowledge from relative tasks to improve learning on a primary task. 
Similarly, multi-task learning methods suggest using auxiliary tasks to enhance a neural network's performance on a specific primary task.
However, previous methods often select auxiliary tasks carefully but treat them as secondary during training. 
The weights assigned to auxiliary losses are typically smaller than the primary loss weight, leading to insufficient training on auxiliary tasks and ultimately failing to support the main task effectively.
To address this issue, we propose an uncertainty-based impartial learning method
that ensures balanced training across all tasks.
Additionally, we consider both gradients and uncertainty information during backpropagation to further improve performance on the primary task.
Extensive experiments show that our method achieves performance comparable to or better than state-of-the-art approaches.
Moreover, our weighting strategy is effective and robust in enhancing the performance of the primary task regardless the noise auxiliary tasks' pseudo labels.

\end{abstract}
}

\maketitle
% \IEEEpeerreviewmaketitle
\IEEEdisplaynontitleabstractindextext

% \fi
\IEEEpeerreviewmaketitle

\section{Introduction}

\label{sec:intro}

Humans excel at acquiring new skills and consistently use past experiences to improve proficiency.
Similarly, advancements in machine learning have highlighted the capabilities of multi-task learning, which mimics the human ability to enhance performance and generalization in one task by concurrently learning different but related tasks~\cite{nddr_cnn, bottom_split, mtl_cluster_trans, npe_trans, cross_stitch, zhang2023controlvideo}.
In recent years, multi-task learning has gained prominence as an efficient tool, recognized for its cost-effective computation~\cite{zhang2015tcdcn}, high performance~\cite{mlrn}, and ability to traverse across modalities~\cite{unip_v2, must}.

In multi-task learning, auxiliary tasks are trained alongside the primary task, providing additional insights.
For instance, Liebel~\etal \cite{synMT} enhances depth prediction and instance segmentation tasks in autonomous driving by training simple auxiliary tasks, including predicting time, vehicle models, and weather conditions.
However, the relationship between the primary task and auxiliary tasks still lacks comprehensive exploration.
Standley \etal~\cite{2020_standley} pointed out that this relationship changes in real-time during training, and overly strong auxiliary tasks can impair the primary task's performance. 
Therefore, jointly optimizing the primary and auxiliary tasks remains a challenge.

\begin{figure}
    \centering
    \subfigure{
        \rotatebox[origin=c]{90}{\centering\scriptsize{OL-AUX}}
        \begin{minipage}{0.45\linewidth}
            \centering
            \includegraphics[width=0.993\textwidth,height=0.7in]{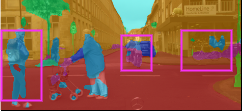}\\
        \end{minipage}%
    }%
    \subfigure{
        \begin{minipage}{0.45\linewidth}
            \centering
            \includegraphics[width=0.993\textwidth,height=0.7in]{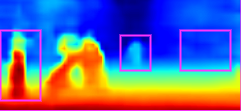}\\
        \end{minipage}%
    }%
    \vspace{-3mm}
    \setcounter{subfigure}{0}
    \subfigure[semantic segmentation]{
        \rotatebox[origin=c]{90}{\scriptsize{Ours}}
        \begin{minipage}{0.45\linewidth}
            \centering
            \includegraphics[width=0.993\textwidth,height=0.7in]{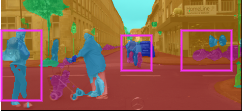}\\
            \vspace{1mm}
        \end{minipage}%
    }%
    \subfigure[disparity estimation]{
        \begin{minipage}{0.45\linewidth}
            \centering
            \includegraphics[width=0.993\textwidth,height=0.7in]{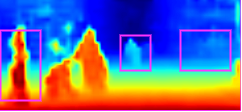}\\
            \vspace{1mm}
        \end{minipage}%
    }%

    \vspace{-3mm}
    \caption{Visual comparison between our impartial training framework and OL-AUX~\cite{ol_aux}. By sufficiently training the auxiliary task~(disparity estimation), our method achieves noticeably better results, as evidenced by the visualizations on the bottom row. Furthermore, we observe that by leveraging the knowledge provided by better auxiliary task, the predictions of the primary task~(semantic segmentation) are also improved. }
    \vspace{-6mm}
    \label{fig:teaser}
\end{figure}

To understand the relationship between primary and auxiliary tasks, current research can be categorized into heuristic-based and meta-learning-based approaches. 
OL-AUX~\cite{ol_aux} and GCS~\cite{gcs} use heuristic methods to select auxiliary tasks that closely resemble the primary task for training.
On the other hand, Auto-$\lambda$~\cite{autolambda} employs a meta-learning approach, optimizing a look-ahead meta-loss function that aggregates gradients from both primary and auxiliary tasks to ensure the primary task's continuous loss descent.
However, both approaches primarily focus on minimizing interference from auxiliary tasks, neglecting their proper training.
Inadequately trained auxiliary tasks negatively impact the primary task's performance, as shown in Fig.~\ref{fig:teaser}. 
On the Cityscapes benchmark, the OL-AUX model allocates just one-tenth of the training weight to depth prediction as an auxiliary task compared to the primary task, resulting in weaker performance for both (top row of Fig.~\ref{fig:teaser}) compared to models using unbiased training (bottom row of Fig.~\ref{fig:teaser}). 
This highlights the importance of properly training auxiliary tasks in multi-task learning frameworks.

To properly train both primary and auxiliary tasks, we propose a two-stage balanced multi-task learning framework. 
To ensure unprejudiced and effective training of auxiliary tasks, uncertainty weights are introduced to decouple the primary task from auxiliary tasks during the decoder stage. The optimizer dynamically adjusts each task's weight based on the loss function associated with these uncertainty weights, as described in~\cite{uw}.
These weights are based solely on the tasks' outputs, not the primary task's needs.
Consequently, all task decoders are trained independently, promoting impartial auxiliary learning.
In the encoder stage, the estimated task uncertainty from the decoder stage determines the auxiliary task weight. 
To ensure the primary task's dominance during training, the gradient norm is further employed to balance the natural gradient magnitude differences between tasks. 
Combining the above two stages, our impartial multi-task learning framework allows the multi-task model to train better auxiliary tasks, which in turn improves the generalization and prediction accuracy of the primary task.

Our method is evaluated on the NYUv2, Cityscapes, Pascal-Context, and CIFAR100 benchmarks.
In these experiments, each task could serve as the primary task, with the others as auxiliary tasks. 
The results demonstrate that our method effectively handles various types of primary and auxiliary tasks, including semantic segmentation, classification, and depth estimation.
Furthermore, we extended our evaluation to a new multi-task setting that collaborates with large-scale pre-trained models to provide pseudo-auxiliary tasks, eliminating the need for manual annotations.
This novel setting presents a fresh challenge for the robustness of multi-task learning, where the noisy auxiliary tasks might disrupt the training of the primary task due to a substantial domain gap. 
The results prove the method's robustness in handling these pseudo-auxiliary tasks.

In summary, the main contribution of this work includes:

\begin {itemize}

  \item We investigate existing multi-task learning methods and identify a lack of focus on auxiliary task training. We show that an imbalance in training between primary and auxiliary tasks leads to ineffective information from auxiliary tasks.
  \item We introduce uncertainty-based impartial learning during the decoder stage to ensure well-balanced training on auxiliary tasks. Additionally, during the encoder stage, we consider both gradients and uncertainty information during backpropagation to further enhance the primary task's performance.
  \item  Through extensive experiments on NYUv2~\cite{nyuv2}, Cityscapes~\cite{cityscapes}, Pascal-Context~\cite{pascal_context}, and Multi-CIFAR100~\cite{autolambda}, we demonstrate that our methods not only improve the primary task by enhancing auxiliary tasks but also perform well even in pseudo-auxiliary tasks.

\end {itemize}
\section{Related Work}
\label{sec:related_works}

\noindent \textbf{Parameter sharing in Multi-task Learning}
Recent multi-task learning (MTL) models are based on popular CNN and Transformer architectures.
These models typically utilize either hard parameter sharing or soft parameter sharing. 
In soft parameter sharing, each task maintains its own encoders and decoders~\cite{cross_stitch, nddr_cnn, mtl_trans}, whereas hard parameter sharing allows multiple tasks to share some model parameters. 
The newly proposed multi-task transformer~\cite{mtl_trans} is based on soft parameter sharing, advocating the use of task-specific transformers. It links cross-task transformers within the task-specific transformer to share information between tasks.
Given their effectiveness in minimizing the risk of overfitting and reducing storage costs, this paper focuses on hard parameter sharing.

\noindent \textbf{Task weighting in Multi-task Learning}
A multi-task model processes multiple tasks in parallel and generates individual outputs for each task, increasing time and computational resource efficiency.
However, owing to a significant optimization challenge, performance on joint learning of multiple tasks is even worse than single-task models.
Weighting tasks with specific values is a typical solution to this challenge. 
Several works~\cite{rlw, uw, imtl} focus on the relative improvement across all tasks. 
Ideally, a well-designed multi-task model treats all tasks equitably, achieving performance levels comparable to or better than those of single-task counterparts.
The main challenge in multi-task learning is resolving task conflicts and achieving task balance.
The Uncertainty Weight (UW) approach~\cite{uw} uses task-specific homoscedastic uncertainty as a scaling factor for loss to balance the joint training of multiple tasks.
Though this study incorporates a similar optimization objective, it is only applied to task-specific decoders to enhance learning of both primary and auxiliary tasks. 
Similarly, IMTL~\cite{imtl} aims to design loss weighting methods that ensure impartial training across all tasks.
Although IMTL shares similar insights, the proposed method prioritizes primary tasks rather than all tasks. 
It applies impartial training only to task-specific decoders. 
This approach avoids the complexity of matrix computations among task-specific gradients.
Miraliev \etal~\cite{RTM2024} trained the loss weights of both multiple tasks and objectives simultaneously, effectively facilitating MTL.
For instance, the Dice loss and IoU loss do not share the same weights in Semantic Segmentation. Furthermore, ATW~\cite{UMTNet2024}, based on Uncertainty Weight, considers the long-term rate of change in the loss of each task to balance the training speed across different tasks. 
Compared to ATW, Miraliev \etal train independent weights for multiple objectives within tasks, while ATW focuses more on utilizing long-term loss information rather than the relationships between tasks.

In MTL, task-weighting class methods can be viewed as the optimal weights of various tasks. PUW~\cite{PUM2023} builds on this concept, employing particle filters to address this complex search challenge.
Meanwhile, CMGEL~\cite{CMGEL2024} and MTLJD~\cite{TNNLS_2024_Evolution} utilize evolutionary algorithms. CMGEL is enhanced by carefully designed global and local metrics, allowing a closer approximation to the global optimal solution. 
PUW, MTLJD and CMGEL jointly strive to converge on the optimal combination in the weight space to maximize the performance of all tasks.
However, our method focuses on the training of auxiliary tasks to maximize the performance of the primary task.

When considering multi-task learning as a form of multi-objective optimization, it's important to recognize that each task often has its own set of parameters. 
Incorporating these task-specific parameters as part of objectives means that optimizing the encoder stage in multi-task learning frameworks can be seen as tackling a multi-objective optimization problem.
Within the encoder stage, the purpose is to achieve balance between primary and auxiliary tasks, rather than striving for equal training intensity across tasks.
To this end, our approach maintains the primary task while balancing the auxiliary tasks through mechanisms such as gradient magnitude and task uncertainty.
For instance, Faster-RCNN~\cite{faster_rcnn} employs various manually selected strategies to achieve balance in multi-objective optimization, including setting the balance weight $\lambda$ to a constant value of 10 or adjusting it according to the number of boxes or classes. Previous research~\cite{mtmo, paretomtl} identifies balanced solutions along the Pareto front, resulting in one task outperforming others in certain configurations. These findings further corroborate our perspective.

\noindent \textbf{Gradients in Multi-task Learning.}
Gradient is commonly used as an important indicator to balance task imbalance in multi-task learning, where Gradient norms are an essential factor.
In multi-task learning, the gradient norms of different tasks vary widely, ranging from 0.1 to over 100.
GradNorm~\cite{gradnorm} estimates an expected magnitude of gradient norms via the losses and their descending rate, where the well-learnt tasks~(with higher convergence speed) are set to relatively low weights. 
Meanwhile, tasks with slower convergence speeds are encouraged to learn faster with higher weights.
This work also uses gradient norm as an important indicator to adjust between tasks.
However, our approach simply re-norms the gradients of auxiliary tasks to match the magnitude of the primary task gradient norm.
PCGrad~\cite{pcgrad} and CAGrad~\cite{cagrad} use the direction of task-specific gradients to re-project the gradients, leading to a balanced situation where the cosine similarity between each pair of task-specific gradients is close to 0. 
However, these methods suffer from the high complexity of comparing and projecting gradients.

\noindent \textbf{Auxiliary tasks in Multi-task learning.}
Another setting of multi-task learning focuses only on a subset of tasks, typically just one primary task. 
A common practice is to categorize each task $t \in T = \{1, ..., K\}$ into two types: the primary task $t_{pri}$ and the auxiliary task $t_{aux}$.
To enhance the generalization of the primary task, the auxiliary task set $T_{aux}$ is incorporated into the training process for joint learning.

In this setting, Gradient Cosine Similarity~(GCS~\cite{gcs}) eliminates the cosine function to measure the similarity between primary and available auxiliary tasks and weights auxiliary tasks in a 'hard' way.
If an auxiliary task has negative cosine similarity with the primary tasks, this task would not be used to joint learning. 
Online Learning for AUXiliary tasks~(OL-AUX~\cite{ol_aux}) follows the similar idea, but provides softer weights. 
It gradually updates task-specific weights during training.
Auto-$\lambda$~\cite{autolambda} is proposed as a meta-learning method for multi-task learning. 
By formulating an additional optimization, the task-specific weights of both primary and auxiliary tasks are optimized to reach the lowest losses of primary tasks. 

\noindent \textbf{Transfer Learning}
Given large labeled datasets such as ImageNet~\cite{imagenet} and COCO, neural networks can learn various tasks, including image-level classification and dense prediction.
However, acquiring large datasets can be prohibitively expensive, resulting in the development of transfer learning. 
The most common use of transfer learning is fine-tuning pre-trained parameters on ImageNet~\cite{imagenet} for target tasks such as semantic segmentation or monocular depth estimation.
In this work, pseudo-auxiliary tasks are proposed by inferring pre-trained models on a multi-task dataset to create noisy but informative labels as auxiliary tasks, which can further examine the robustness of our method.
MuST~\cite{must} uses a similar approach to train a general pre-trained model with multi-task learning.
However, our work focuses on improving primary tasks through multi-task learning rather than transfer performance on downstream tasks.
The core contribution of this work is solving the optimization challenge during multi-task joint training. 
The results may also be applied to transfer learning, similar to MuST~\cite{must}.

\section{Methodology}
% 想了想还是主要用primary task来指代主任务吧。这样叫的人会多一些。
To capitalize on the knowledge gleaned from auxiliary tasks and enhance the primary task's learning, we propose the Impartial Auxiliary Learning (IAL) framework.
This multi-task learning framework is based on the pivotal observation that better-optimized auxiliary tasks lead to improved performance on the primary task.

This section is structured in the following manner: Firstly, Section \ref{sec:formulation} offers a precise definition of the multi-task learning problem with auxiliary tasks. Then, Section \ref{sec:framework} delves into the motivations behind this approach and suggests potential modifications to the encoder and decoder optimization stages. Lastly, Section \ref{sec:details} outlines the technical intricacies of the IAL framework and its implementation.

\begin{figure*}
    \centering
    \subfigure[Baseline]{
        \centering
        \includegraphics[width=0.45\linewidth]{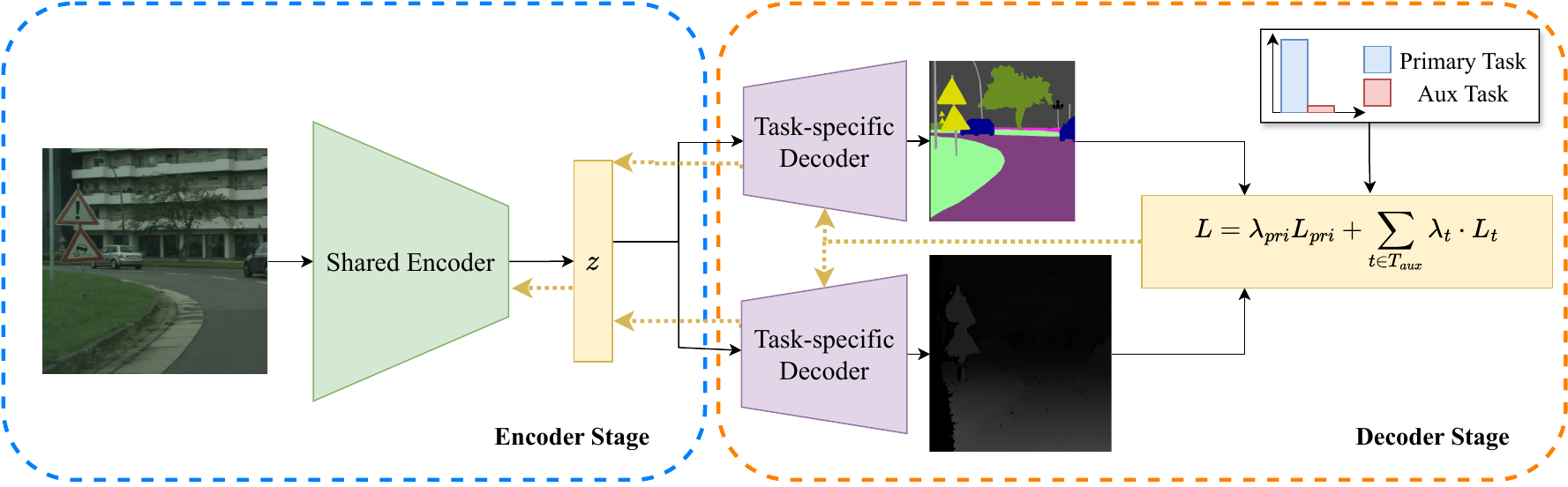}
        %\caption{Baseline}
    }
    %\hfill
    \subfigure[Ours]{
        \centering
        \includegraphics[width=0.45\linewidth]{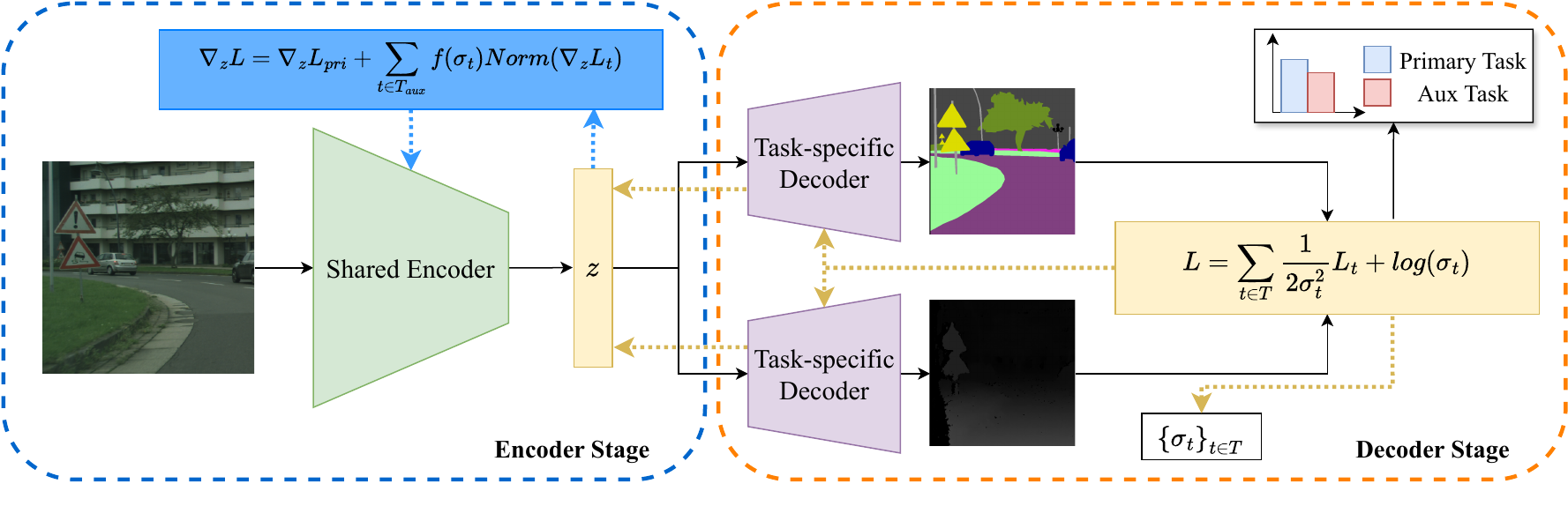}
        %\caption{Ours}
    }
    \vspace{-3mm}
    \caption{An comparison between our equally training framework and previous methods. (a)~Baseline methods, use a weight estimator to produce auxiliary weights according to their gradient norms, directions or loss magnitudes. All parameters are trained under one group of weights, and the auxiliary tasks are always set to lower weights leading to poor training problem. (b) An overview of our framework. For the decoder stage~(parameters surrounding by orange dash lines), we update the task-specific decoders with uncertainty weights~\cite{uw} which are changed only according to their corresponding losses. In this stage, the uncertainty $\sigma_t$ is estimated automatically. For the encoder stage~(shared encoder surrounding by blue dash lines), we use both gradients and uncertainty information to weight the auxiliary tasks. More details will be reported on Sec.~\ref{sec:details}. }
    \vspace{-6mm}
    \label{fig:overview}
\end{figure*}

\subsection{Problem Formulation}
\label{sec:formulation}

In a multi-task training framework, each task is equipped with a separate decoder $f_{dec}(\cdot)$, which has its parameters $\theta_t$, where $t \in T$.
However, all tasks share the same encoder $f_{enc}(\cdot;\theta_{shared})$, which maps images/samples $x$ into features $z=f(x;\theta_{shared})$ .
The task-specific decoders predict the corresponding outputs using the shared features, \ie, ${y'}_t=f_{dec}(z;\theta_t)$. 
Each task has its own loss function $\mathcal{L}_t=Loss({y'}_t,y_t)$ to minimize its own objective.
Therefore, a significant challenge is to optimize the training process across the task set $T=T_{pri} \cup T_{aux}$ to minimize the loss of the primary task.
To overcome this challenge, a set of weights $\{\lambda_1,\ldots,\lambda_K\}$ is defined on the task set $T$.
Each weight corresponds to a task, with the primary task's weight $\lambda_{pri}$ fixed at 1, while the weights of other auxiliary tasks $\{ \lambda_t|t\in T_{aux}\}$ are far lower than 1.
With these weights, the loss functions corresponding to each task $L_t$ are combined into a multi-task loss function through linear combination.

\begin{equation}
\mathcal{L}_{MTL}=\lambda_{pri} \mathcal{L}_{pri}+\sum\limits_{t\in T_{aux}} \lambda_t \mathcal{L}_t,
\label{eq:mtl_loss}
\end{equation}

Figure~\ref{fig:overview}(a) presents a widely used pipeline for multi-task learning that involves auxiliary tasks.
The multi-task loss is formed by taking a linear combination of pre-computed weights and task losses.
During back-propagation, task-specific decoder gradients are computed based on the multi-task loss, updating the decoder parameters. 
This process is denoted as the decoder stage.

Afterward, the gradients of each task are propagated to the shared features and continue to compute gradients for the shared encoder. At this point, the gradient of the shared encoder is:
\begin{equation}
\nabla_{\theta_{enc}}\mathcal{L}_{MTL}=\lambda_{pri} \nabla_{\theta_{enc}}\mathcal{L}_{pri}+\sum\limits_{t\in T_{aux}} \lambda_t \nabla_{\theta_{enc}}\mathcal{L}_t,
\label{eq:enc_grad}
\end{equation}

During the encoder stage of multi-task learning, the gradient is formed by using linear combinations to merge per-task gradients and weights.
% , similar to how the multi-task loss function is built. 
However, this can cause task conflicts and gradient cancellation due to the involvement of gradients and weights from all tasks. 
To avoid this, it is crucial to prioritize the primary task during this stage and filter out any potentially harmful auxiliary tasks.

In the overall multi-task loss function, the weight of the primary task is dominant. 
The weights of auxiliary tasks are generally determined by grid search or based on their correlation with the primary task.
The former often requires a large amount of repeated model training, resulting in unacceptable computational waste. 
The latter often assign lower weights to auxiliary tasks due to natural differences between tasks. This results in poor training and an inability to provide accurate and rich knowledge for the primary task.

\subsection{Proposed Solution: Impartial Auxiliary Learning}
\label{sec:framework}

A new training framework is proposed for properly training both primary and auxiliary tasks, as shown in Fig.~\ref{fig:overview}(b).
In the decoder stage, the main objective is to improve the performance of the decoders by providing proper training and attention to auxiliary tasks.
To achieve this goal, the concept of equal training from another branch of multi-task learning is introduced. 
The weight of each task depends only on itself, ensuring equal importance.  
Additionally, uncertainty-based weights are used to adjust each task's training pace according to its level of uncertainty. 
More practice training details are provided in the Experiments section.

After the decoder stage, the encoder stage presents a significant challenge due to conflicts and competition for resources among tasks sharing weights.
If this optimization issue is not addressed, multi-task models may underperform compared to single-task models.
Moreover, multi-task learning with auxiliary tasks poses an even more challenging and less explored problem: balancing the prioritization of the primary task during training while ensuring auxiliary tasks contribute sufficient knowledge to enhance generalization.

To prioritize the primary task during training, it is crucial to ensure it receives a proportional share of optimization. 
Existing methods often assign a weight of 1 to the primary task and less than 1 to each auxiliary task.
However, this approach can lead to issues when the gradient norms of each task differ significantly. 
Tasks with larger gradient norms may dominate optimization, negatively impacting the primary task's learning.
To address this, the gradient norms of all tasks are normalized to that of the primary task, effectively fixing its weight to 1. 
By shrinking the gradients of the auxiliary tasks, the Impartial Auxiliary Learning (IAL) framework ensures that the primary task remains dominant during training.

The next step is to adjust the weights of the auxiliary tasks so they have less impact during training.
The approach prioritizes better-optimized auxiliary tasks by assigning them higher weights. 
However, the challenge lies in evaluating the quality of training for each task, referred to as training levels, which are defined by the generalization performance of the tasks.
Previous methods, such as DTP (Dynamic Task Priority)~\cite{dtp}, used manually selected metrics to assess training levels, making fair cross-task comparisons difficult.
Instead, the proposed approach leverages uncertainty estimation obtained for each task to measure training levels in the decoder stage, avoiding hand-selected metrics used in DTP.
A simple mapping function is proposed to associate the uncertainty of a task with a weight, constrained within the range of [0, 1). 
This design serves two main purposes. 
Firstly, the weight assigned to each task does not exceed 1, ensuring the primary task's dominance in the learning process. 
Secondly, for auxiliary tasks with higher uncertainty, this weight limits their impact on the primary task while still allowing them to contribute meaningfully.

To sum up, the approach in the encoder stage prioritizes the primary task during training while promoting well-optimized auxiliary tasks to provide more insights. 
This leads to better generalization and accuracy of the primary task.

\subsection{Algorithm}
\label{sec:details}
In this section,  the specifics of the methodologies are elucidated, focusing on the two primary stages: the decoder and encoder stages. 
% As depicted in Fig.~\ref{fig:overview}(b), our technique enhances performance in both these stages. 
In the decoder stage, uncertainty-based task weights are employed to compute the multi-task loss function:
\begin{equation}
\mathcal{L}_{MTL}=\sum\limits_{t\in T} \frac{1}{2\sigma_t^2} \mathcal{L}_t + \log{\sigma_t},
\label{eq:uw_loss}
\end{equation}
Where the task-specific weight $ \lambda_t=\frac{1}{2\sigma_t^2}$ follows the form of Uncertainty Weight~\cite{uw} to effectively estimate task uncertainty. 
As in \cite{uw}, the uncertainty is brought from Bayesian modeling.
Aleatoric uncertainty captures uncertainty concerning information that the data cannot explain.
Aleatoric uncertainty can be divided into two categories: Data-dependent and Task-dependent uncertainty. 
In this setting, the focus is on task-dependent uncertainty, which stays constant across all data but varies between tasks. 
This work borrows related concept and estimates task-dependent uncertainty for auxiliary tasks.

The task-dependent uncertainty may change during the training process. 
So these weights which represent the task-dependent uncertainty are not pre-computed but updated using the multi-task loss as:
\begin{equation}
\nabla_{\sigma_t}\mathcal{L}_{MTL}=-\frac{1}{\sigma_t^3} \mathcal{L}_t+\frac{1}{x},
\label{eq:weight_grad}
\end{equation}
It's apparent that the update of a task's weight is contingent solely on its own loss, independent of any other tasks. 
This approach effectively circumvents the issue of auxiliary tasks acquiring exceedingly small weights, thereby mitigating any loss in performance due to the influence of the primary task or other auxiliary tasks.

Likewise, the task-specific decoder's parameters $ \theta_t$ are updated using this loss function, with gradients:
\begin{equation}
\nabla_{\theta_t}\mathcal{L}_{MTL}=\frac{1}{2\sigma_t^2} \nabla_{\theta_t}\mathcal{L}_t,
\label{eq:decoder_grad}
\end{equation}
This gradient is also only influenced by the loss of that particular task.

In the encoder stage, unlike previous methods in Fig.~\ref{fig:overview}(a), we adopt a new optimization objective which is similar to Eq.~\ref{eq:mtl_loss}:
\begin{equation}
\mathcal{L}_{MTL}=\mathcal{L}_{pri}+\sum\limits_{t\in T_{aux}} f(\sigma_t) \mathcal{L}_t,
\label{eq:ug_loss}
\end{equation}
Where the primary task weight is fixed at 1, the auxiliary task weights $f(\sigma_t), t\in T_{aux}$ are dependent on the uncertainty$\{ \sigma_t \}_{t\in T_{aux}}$ estimated during the decoder stage. 
This approach is grounded in the observation that superior auxiliary tasks assist better to the primary tasks.
According to this loss, the gradient of the shared features can be determined, and additional gradient normalization can be incorporated as:
\begin{equation}
\nabla_{z}\mathcal{L}_{MTL}=\nabla_{z}\mathcal{L}_{pri}+\sum\limits_{t\in T_{aux}} f(\sigma_t) Norm(\nabla_{z}\mathcal{L}_t),
\label{eq:z_grad}
\end{equation}

As mentioned before, gradients of the auxiliary tasks are normalized as follow:
\begin{equation}
Norm(\nabla_{z}\mathcal{L}_{t})=\frac{|\nabla_{z}\mathcal{L}_{pri}|}{|\nabla_{z}\mathcal{L}_t|} \nabla_{z}\mathcal{L}_t,
\label{eq:grad_norm}
\end{equation}
Through this operation, the gradient norms of the auxiliary tasks are aligned with the primary task in magnitude. Combined with the subsequent weighting, this effectively ensures the dominance of the primary task during training.

The task-specific weight function $f(\sigma_t)$ is a function related to task uncertainty, specifically:
\begin{equation}
f(x) = min(1.0, g(1-x)),
\label{eq:ug_weight}
\end{equation}
Where $g(\cdot)$ is a function that maps the uncertainty to the range $[0, +\inf)$. This function ensures that the weights lie within the [0,1) interval: high uncertainty leads to auxiliary task noise influencing primary task training, causing weights to decrease towards 0. Conversely, well-trained tasks are associated with low task uncertainty, and thus weights close to 1. In our experiments, these weights typically do not exceed 0.6.
Subsequently, with the shared feature gradients established, we can utilize the chain rule to compute the encoder gradients as follows:
\begin{equation}
\nabla_{\theta_{enc}}\mathcal{L}_{MTL} = \frac{d \mathcal{L}_{MTL}}{d z} \frac{d z}{d \theta_{enc}},
\label{eq:encoder_grad}
\end{equation}

Based on the above, we formally describe our framework:
\begin {enumerate}
  \item Forward the multi-task network with the samples $ \{(x, y_1, y_2, ..., y_K)\}_N $. 
  \item Decoder stage: compute multi-task loss based on Eq.~\ref{eq:uw_loss}. With the multi-task loss, the gradients of the uncertainty and the decoder parameters could be back-propagated with Eq.~\ref{eq:decoder_grad} and Eq.~\ref{eq:weight_grad}. Use the gradients to update the decoders and corresponding uncertainty. 
  \item Encoder stage: Compute gradients of the shared feature $z$ based on the new objective~(Eq.~\ref{eq:ug_loss}); use the chain rule to compute encoder gradients from shared feature gradients with Eq.~\ref{eq:encoder_grad}, and update the shared encoder.
\end {enumerate}
\section{Experiments}
\label{sec:exp}

\begin{table*}[!htbp]
\caption{Comparison of multi-task learning performance on \textbf{NYUv2} and \textbf{Cityscapes}, where the mean and standard deviation over $5$ random seeds for each measurement are reported. The best and $2$nd best performances are marked in \textbf{bold} and \underline{underlined}, respectively. All results are reimplemented. \label{table:dense_prediction}}
\vspace{-2mm}
\centering
\renewcommand\arraystretch{1.2}
\resizebox{\linewidth}{!}{
\begin{tabular}{c l cccc cccc}
\toprule
&\multirow{2}{*}[-2.0ex]{\textbf{Method}}& \multicolumn{4}{c}{\textbf{\texttt{NYUv2}}} & \multicolumn{4}{c}{\textbf{\texttt{Cityscapes}}}\\
\cmidrule(lr){3-6} \cmidrule(l){7-10}
& & \textbf{Depth} & \textbf{Segmentation} & \textbf{Normals} & $\Delta \textbf{MTL} $ 
                         & \textbf{Disparity} & \textbf{Part Seg.} & \textbf{Semantic Seg.} & $ \Delta \textbf{MTL} $  \\
  & & RMSE[m]($\downarrow$) & mIoU[\%]($\uparrow$) & Mean Error($\downarrow$) & [\%]($\uparrow$) 
  & aErr[m]($\downarrow$) & mIoU[\%]($\uparrow$) & mIoU[\%]($\uparrow$) & [\%]($\uparrow$)\\
\cmidrule(r){1-2} \cmidrule(lr){3-3}\cmidrule(lr){4-4} \cmidrule(lr){5-5} \cmidrule(lr){6-6}
\cmidrule(lr){7-7} \cmidrule(lr){8-8} \cmidrule(lr){9-9} \cmidrule(l){10-10}
 &\multirow{3}{*}{Single task}    & \multicolumn{1}{|c}{0.5877~\stdvu{±0.0006}}  & -       & -       & - 
                                & \multicolumn{1}{|c}{0.8403~\stdvu{±0.0009}}  & -       & -       & -  \\
 & & \multicolumn{1}{|c}{-}       & 43.58~\stdvu{±0.05}   & -       & - 
                                & \multicolumn{1}{|c}{-}       & 52.71~\stdvu{±0.06}   & -       & - \\
                                & & \multicolumn{1}{|c}{-}       & -       & 19.49~\stdvu{±0.03}   & - 
                                & \multicolumn{1}{|c}{-}       & -       & 56.22~\stdvu{±0.03}   & - \\

\midrule
\multirow{11}{*}{\rotatebox{90}{\texttt{Normal}}}&\multicolumn{1}{|l}{Uniform}     & \multicolumn{1}{|c}{0.5933~\stdvu{±0.0008}} & 43.47~\stdvu{±0.03} & 21.91~\stdvu{±0.03} & - 4.45\%   
            & \multicolumn{1}{|c}{0.7959~\stdvu{±0.0010}} & 51.55~\stdvu{±0.06} & 54.12~\stdvu{±0.04} & - 0.22\% \\
&\multicolumn{1}{|l}{UW~\cite{uw}}    & \multicolumn{1}{|c}{0.5874~\stdvu{±0.0004}} & 43.97~\stdvu{±0.06} & 21.61~\stdvu{±0.03} & - 3.31\%   
                & \multicolumn{1}{|c}{0.8223~\stdvu{±0.0003}} & 52.96~\stdvu{±0.06} & 56.11~\stdvu{±0.04} & + 0.81\% \\
&\multicolumn{1}{|l}{GradNorm~\cite{gradnorm}} & \multicolumn{1}{|c}{0.5951~\stdvu{±0.0011}} & 41.82~\stdvu{±0.09} & \underline{20.66~\stdvu{±0.03}} & - 3.77\%
                        & \multicolumn{1}{|c}{0.8037~\stdvu{±0.0010}} & 52.62~\stdvu{±0.09} & 56.42~\stdvu{±0.07} & + 1.51\% \\ 
&\multicolumn{1}{|l}{DWA~\cite{mtan_dwa}} & \multicolumn{1}{|c}{0.5885~\stdvu{±0.0006}} & 44.13~\stdvu{±0.12} & 21.91~\stdvu{±0.03} & - 3.76\%  
                    & \multicolumn{1}{|c}{0.8011~\stdvu{±0.0008}} & 51.33~\stdvu{±0.08} & 54.82~\stdvu{±0.06} & - 0.15\%\\
&\multicolumn{1}{|l}{MGDA~\cite{mtmo}}    & \multicolumn{1}{|c}{\underline{0.5760~\stdvu{±0.0010}}} & 43.20~\stdvu{±0.08} & 21.42~\stdvu{±0.03} & - 2.93\%
                    & \multicolumn{1}{|c}{0.7966~\stdvu{±0.0015}} & 52.42~\stdvu{±0.09} & 55.11~\stdvu{±0.05} & + 0.89\% \\
&\multicolumn{1}{|l}{PCGrad~\cite{pcgrad}}    & \multicolumn{1}{|c}{0.5860~\stdvu{±0.0009}} & 43.49~\stdvu{±0.09} & 21.74~\stdvu{±0.03} & - 3.82\% 
                        & \multicolumn{1}{|c}{0.7899~\stdvu{±0.0018}} & 53.94~\stdvu{±0.07} & 55.47~\stdvu{±0.03} & + 2.33\% \\
&\multicolumn{1}{|l}{RLW-Normal~\cite{rlw}}    & \multicolumn{1}{|c}{0.5848~\stdvu{±0.0006}} & 44.40~\stdvu{±0.11} & 21.66~\stdvu{±0.03} & - 2.92\% 
                        & \multicolumn{1}{|c}{0.7861~\stdvu{±0.0009}} & 53.82~\stdvu{±0.06} & 55.21~\stdvu{±0.03} & + 2.25\%  \\
&\multicolumn{1}{|l}{IMTL-L~\cite{imtl}}  & \multicolumn{1}{|c}{0.5802~\stdvu{±0.0005}} & 43.31~\stdvu{±0.04} & 21.83~\stdvu{±0.04} & - 3.78\% 
                    & \multicolumn{1}{|c}{0.7849~\stdvu{±0.0010}} & 54.36~\stdvu{±0.06} & 54.97~\stdvu{±0.03} & + 2.50\%\\
&\multicolumn{1}{|l}{IMTL-G~\cite{imtl}}  & \multicolumn{1}{|c}{0.5789~\stdvu{±0.0017}} & 43.88~\stdvu{±0.10} & 20.99~\stdvu{±0.07} & - 1.84\%  
                    & \multicolumn{1}{|c}{0.7702~\stdvu{±0.0027}} & 57.47~\stdvu{±0.11} & 55.99~\stdvu{±0.05} & + 5.65\%\\
&\multicolumn{1}{|l}{IMTL~\cite{imtl}}    & \multicolumn{1}{|c}{0.5791~\stdvu{±0.0010}} & 43.72~\stdvu{±0.06} & 20.90~\stdvu{±0.06} & - 1.82\%  
                    & \multicolumn{1}{|c}{\underline{0.7674~\stdvu{±0.0023}}} & \underline{58.07~\stdvu{±0.07}} & 56.53~\stdvu{±0.04} & + 6.47\%\\
&\multicolumn{1}{|l}{Auto-$\lambda$ (MTL)~\cite{autolambda}}    & \multicolumn{1}{|c}{0.5802~\stdvu{±0.0019}} & 44.01~\stdvu{±0.08} 
                                          & 21.24~\stdvu{±0.03} & - 1.95\%  
                    & \multicolumn{1}{|c}{0.7783~\stdvu{±0.0024}} & 54.67~\stdvu{±0.09} & 55.87~\stdvu{±0.04} & + 3.49\%\\

&\multicolumn{1}{|l}{Miraliev \etal~\cite{RTM2024}}    & \multicolumn{1}{|c}{0.5876~\stdvu{±0.0004}} & 44.00~\stdvu{±0.06} & 21.59~\stdvu{±0.03} & - 3.26\%   
                & \multicolumn{1}{|c}{0.8210~\stdvu{±0.0006}} & 53.54~\stdvu{±0.07} & 56.41~\stdvu{±0.03} & + 1.40\%\\

&\multicolumn{1}{|l}{ATW~\cite{UMTNet2024}}    & \multicolumn{1}{|c}{0.5810~\stdvu{±0.0025}} & 43.95~\stdvu{±0.04} 
                                          & 21.49~\stdvu{±0.03} & - 2.76\%  
                    & \multicolumn{1}{|c}{0.8228~\stdvu{±0.0019}} & 55.73~\stdvu{±0.09} & 51.52~\stdvu{±0.04} & - 0.02\%\\

&\multicolumn{1}{|l}{PUW~\cite{PUM2023}}    & \multicolumn{1}{|c}{0.5800~\stdvu{±0.0015}} & 44.12~\stdvu{±0.10} 
                                          & 21.12~\stdvu{±0.04} & - 1.94\%  
                    & \multicolumn{1}{|c}{0.7858~\stdvu{±0.0010}} & 53.66~\stdvu{±0.09} & 57.10~\stdvu{±0.05} & + 3.28\%\\

\midrule
\multirow{4}{*}{\rotatebox{90}{\texttt{Auxiliary}}}&\multicolumn{1}{|l}{GCS~\cite{gcs}}      & \multicolumn{1}{|c}{0.5862~\stdvu{±0.0012}} & 43.77~\stdvu{±0.13} & 21.67~\stdvu{±0.03} & - 3.49\%  
                    & \multicolumn{1}{|c}{0.7902~\stdvu{±0.0008}} & 53.21~\stdvu{±0.12} & 55.76~\stdvu{±0.05} & + 2.03\%\\
&\multicolumn{1}{|l}{OL-AUX~\cite{ol_aux}}    & \multicolumn{1}{|c}{0.5853~\stdvu{±0.0010}} & 43.73~\stdvu{±0.15} & 21.59~\stdvu{±0.03} & - 3.34\% 
                        & \multicolumn{1}{|c}{0.7963~\stdvu{±0.0008}} & 52.90~\stdvu{±0.11} & 56.56~\stdvu{±0.05} & + 2.07\%\\
&\multicolumn{1}{|l}{Auto-$\lambda$~\cite{autolambda}}    & \multicolumn{1}{|c}{0.5801~\stdvu{±0.0016}} & \underline{44.20~\stdvu{±0.12}} 
                                    & 21.11~\stdvu{±0.03} & \underline{- 1.87\%}  
                                    & \multicolumn{1}{|c}{0.7721~\stdvu{±0.0018}} & 53.61~\stdvu{±0.07} 
                                    & \underline{57.82~\stdvu{±0.04}} & + 4.42\%\\
&\multicolumn{1}{|l}{{\cellcolor{mypink}\texttt{\textbf{Ours}}}}  & \multicolumn{1}{|c}{{\cellcolor{mypink}\textbf{0.5751~\stdvu{±0.0011}}}} & {\cellcolor{mypink}\textbf{44.60~\stdvu{±0.11}}} 
      & {\cellcolor{mypink}\textbf{20.62~\stdvu{±0.03}}} & {\cellcolor{mypink}\textbf{- 0.44\%}} 
      & \multicolumn{1}{|c}{{\cellcolor{mypink}\textbf{0.7612~\stdvu{±0.0010}}}} & {\cellcolor{mypink}\textbf{58.48~\stdvu{±0.07}}} 
      & {\cellcolor{mypink}\textbf{58.63~\stdvu{±0.04}}} & {\cellcolor{mypink}\textbf{+ 8.22}\%}\\

\bottomrule
\end{tabular}}
\end{table*}

\begin{table}[!htbp]
\caption{Comparison of the training time on \textbf{NYUv2}, \textbf{Cityscapes} and \textbf{PASCAL-CONTEXT}. Note that the inference time is the same due to the same model architecture used in our experiments. All results are reimplemented.\label{table:computation_time}}
\vspace{-2mm}
\centering
\renewcommand\arraystretch{1.2}
\resizebox{\linewidth}{!}{
\begin{tabular}{c l c c c}
\toprule
&\textbf{Method}& \textbf{\texttt{NYUv2[hours]}} & \textbf{\texttt{Cityscapes[hours]}} & \textbf{\texttt{PASCAL[hours]}}\\
\cmidrule(r){1-2} \cmidrule(lr){3-3}\cmidrule(lr){4-4}\cmidrule(lr){5-5}
 & Single task    & \multicolumn{1}{|c}{6.0}  & 6.8  & 12.2 \\

\midrule
\multirow{11}{*}{\rotatebox{90}{\texttt{Normal}}}&\multicolumn{1}{|l}{Uniform} & \multicolumn{1}{|c}{\textbf{12.2}} & \textbf{14.7} & \textbf{18.4}\\
&\multicolumn{1}{|l}{UW~\cite{uw}}    & \multicolumn{1}{|c}{\textbf{12.2}} & 14.8 & 18.6\\
&\multicolumn{1}{|l}{GradNorm~\cite{gradnorm}} & \multicolumn{1}{|c}{13.7} & 16.2 & 19.5\\ 
&\multicolumn{1}{|l}{DWA~\cite{mtan_dwa}} & \multicolumn{1}{|c}{12.5} & 15.1 & 18.5\\ 
&\multicolumn{1}{|l}{MGDA~\cite{mtmo}} &\multicolumn{1}{|c}{15.4}  & 17.2 & 20.0\\ 
&\multicolumn{1}{|l}{PCGrad~\cite{pcgrad}} & \multicolumn{1}{|c}{15.2} & 17.1 & 19.3\\ 
&\multicolumn{1}{|l}{RLW-Normal~\cite{rlw}} & \multicolumn{1}{|c}{12.2} & 14.7 & 18.5\\ 
&\multicolumn{1}{|l}{IMTL-L~\cite{imtl}} & \multicolumn{1}{|c}{12.3} & 14.7 & 18.5\\ 
&\multicolumn{1}{|l}{IMTL-G~\cite{imtl}} & \multicolumn{1}{|c}{15.1} & 17.2 & 19.9\\ 
&\multicolumn{1}{|l}{IMTL~\cite{imtl}} & \multicolumn{1}{|c}{15.2} & 17.2 & 20.0\\ 
&\multicolumn{1}{|l}{Auto-$\lambda$ (MTL)~\cite{autolambda}} & \multicolumn{1}{|c}{14.2} & 16.7 & 20.5\\ 
&\multicolumn{1}{|l}{Miraliev \etal~\cite{RTM2024}} & \multicolumn{1}{|c}{\textbf{12.2}} & 14.8 & 18.6\\ 
&\multicolumn{1}{|l}{ATW~\cite{UMTNet2024}} & \multicolumn{1}{|c}{12.3} & 14.9 & 18.8\\ 
&\multicolumn{1}{|l}{PUW~\cite{PUM2023}} & \multicolumn{1}{|c}{14.5} & 17.9 & 22.5\\ 

\midrule
\multirow{4}{*}{\rotatebox{90}{\texttt{Auxiliary}}}&\multicolumn{1}{|l}{GCS~\cite{gcs}} & \multicolumn{1}{|c}{\textbf{13.9}} & \textbf{16.3} & \textbf{19.0}\\ 
&\multicolumn{1}{|l}{OL-AUX~\cite{ol_aux}} & \multicolumn{1}{|c}{14.0} & 16.5 & \textbf{19.0}\\ 
&\multicolumn{1}{|l}{Auto-$\lambda$~\cite{autolambda}} & \multicolumn{1}{|c}{14.2} & 16.7 & 20.5\\ 
&\multicolumn{1}{|l}{{\cellcolor{mypink}\texttt{\textbf{Ours}}}}  & \multicolumn{1}{|c}{{\cellcolor{mypink}14.1}} & {\cellcolor{mypink}16.5}&  {\cellcolor{mypink} 19.4} \\

\bottomrule
\end{tabular}}
\end{table}

\begin{figure*}[!htbp]
\centering
\subfigure[Image]{
    \begin{minipage}{0.19\linewidth}
        \centering
        \includegraphics[width=0.993\textwidth,height=0.7in]{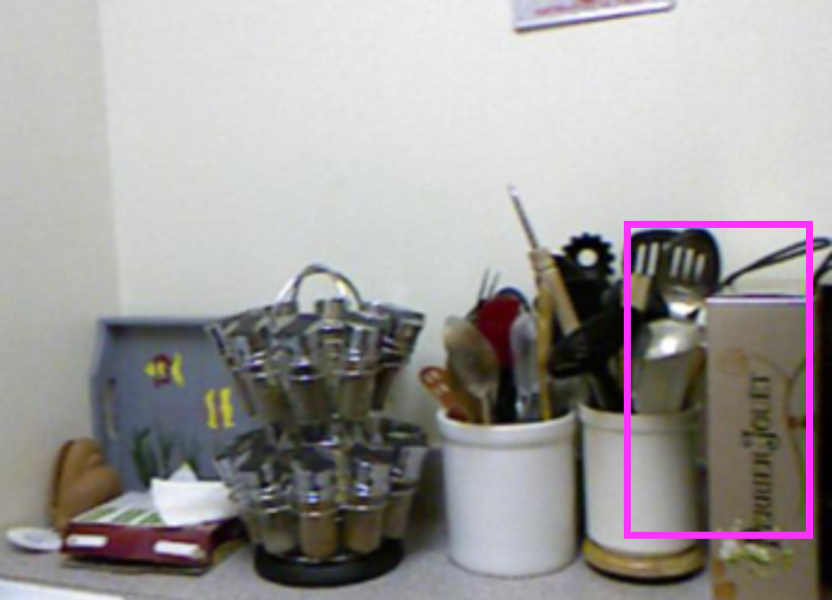}\\
        %\vspace{1mm}
        %\includegraphics[width=0.993\textwidth,height=0.7in]{images/nyu_seg/56_origin_box.png}\\
        \vspace{1mm}
        \includegraphics[width=0.993\textwidth,height=0.7in]{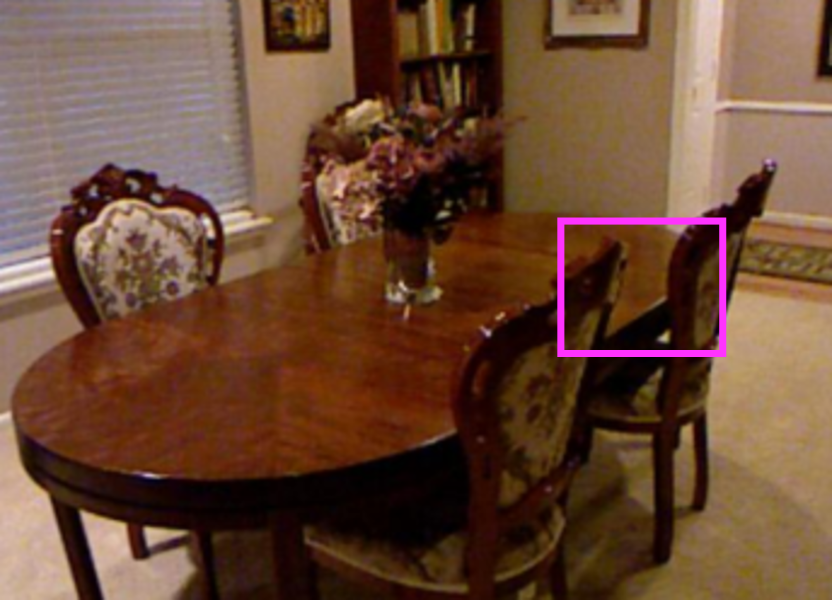}\\    
        \vspace{1mm}
        % \caption{}
    \end{minipage}%
}%
\subfigure[OL$\text{-} $AUX~\cite{ol_aux}]{
    \begin{minipage}{0.19\linewidth}
        \centering
        \includegraphics[width=0.993\textwidth,height=0.7in]{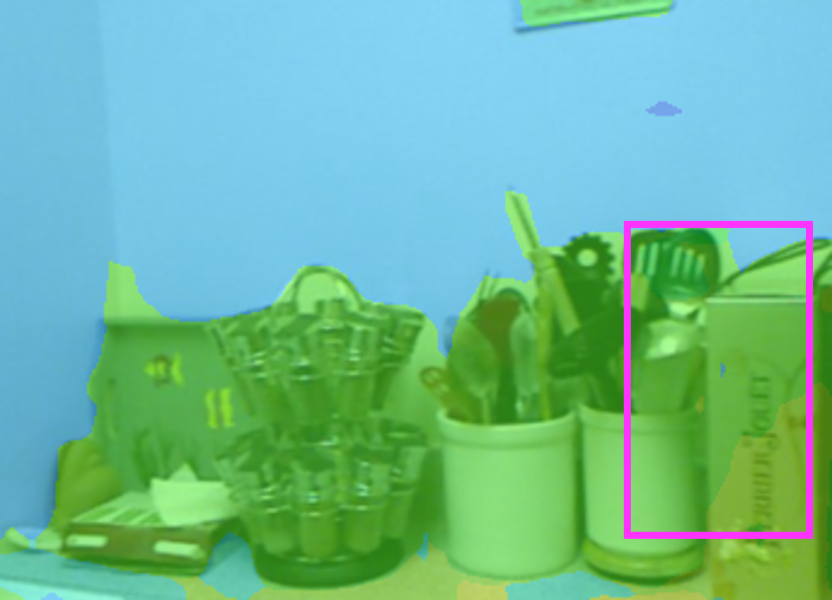}\\
        %\vspace{1mm}
        %\includegraphics[width=0.993\textwidth,height=0.7in]{images/nyu_seg/56_OL-AUX_box.png}\\
        \vspace{1mm}
        \includegraphics[width=0.993\textwidth,height=0.7in]{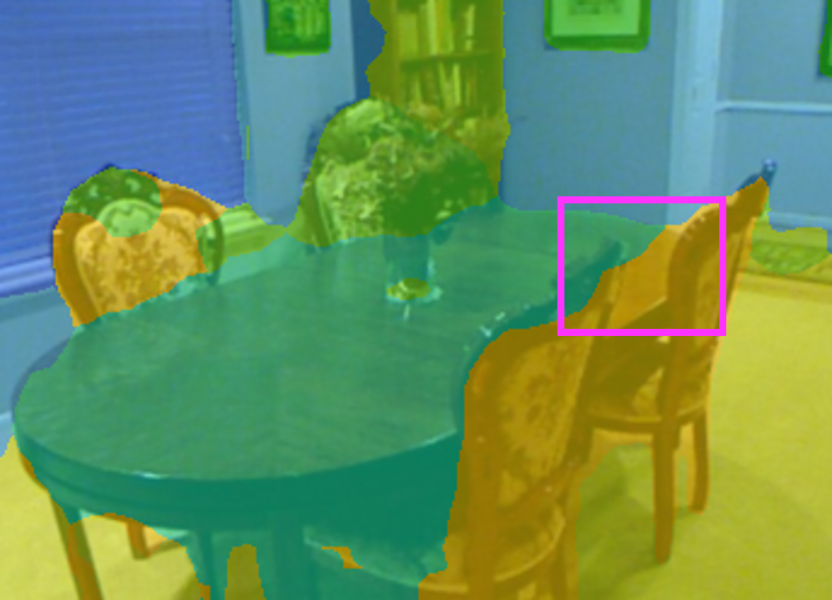}\\  
        \vspace{1mm}
    \end{minipage}%
}%
\subfigure[Auto$\text{-}\lambda $~\cite{autolambda}]{
    \begin{minipage}{0.19\linewidth}
        \centering
        \includegraphics[width=0.993\textwidth,height=0.7in]{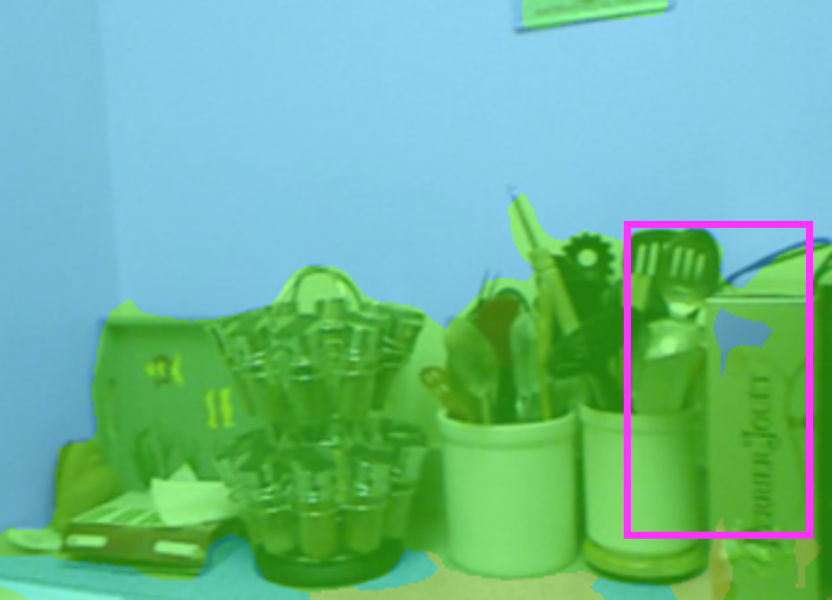}\\
        %\vspace{1mm}
        %\includegraphics[width=0.993\textwidth,height=0.7in]{images/nyu_seg/56_auto-lambda_box.png}\\
        \vspace{1mm}
        \includegraphics[width=0.993\textwidth,height=0.7in]{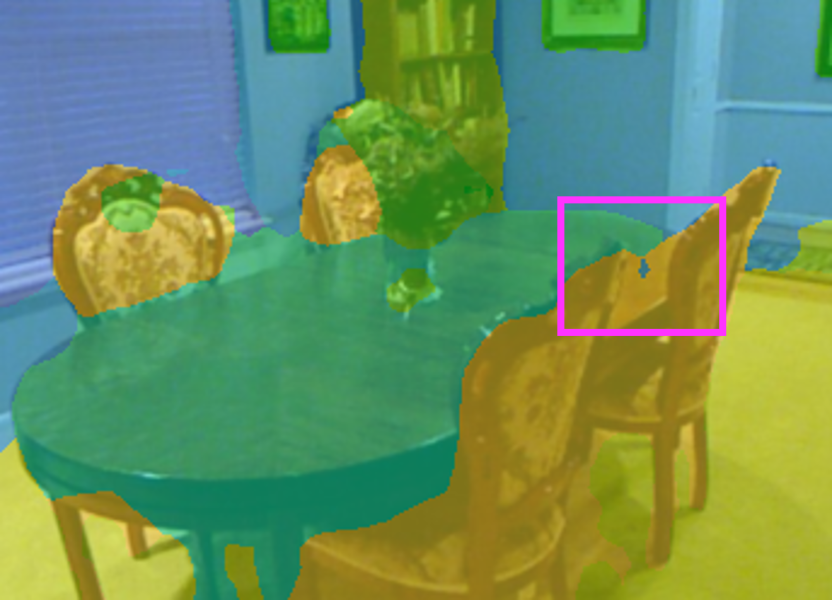}\\  
        \vspace{1mm}
    \end{minipage}%
}%
\subfigure[Ours]{
    \begin{minipage}{0.19\linewidth}
        \centering
        \includegraphics[width=0.993\textwidth,height=0.7in]{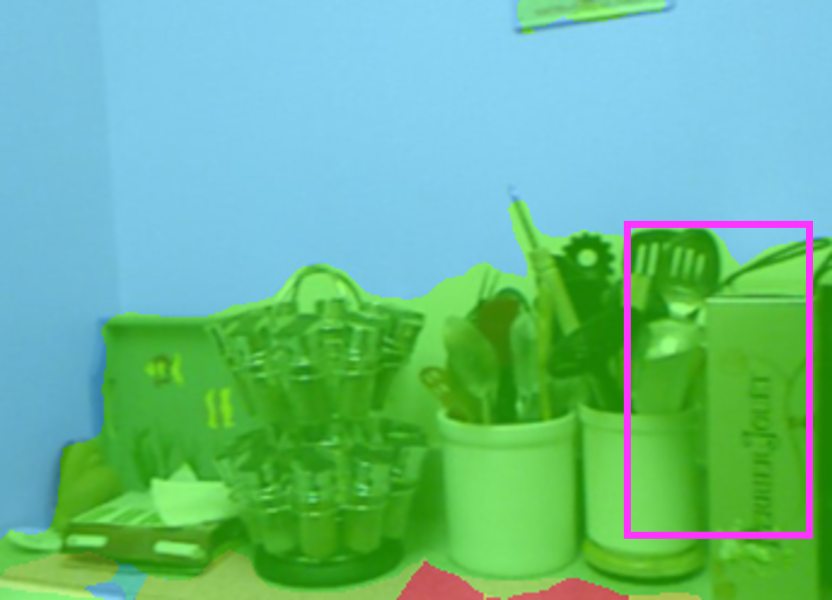}\\
        %\vspace{1mm}
        %\includegraphics[width=0.993\textwidth,height=0.7in]{images/nyu_seg/56_ours_box.png}\\
        \vspace{1mm}
        \includegraphics[width=0.993\textwidth,height=0.7in]{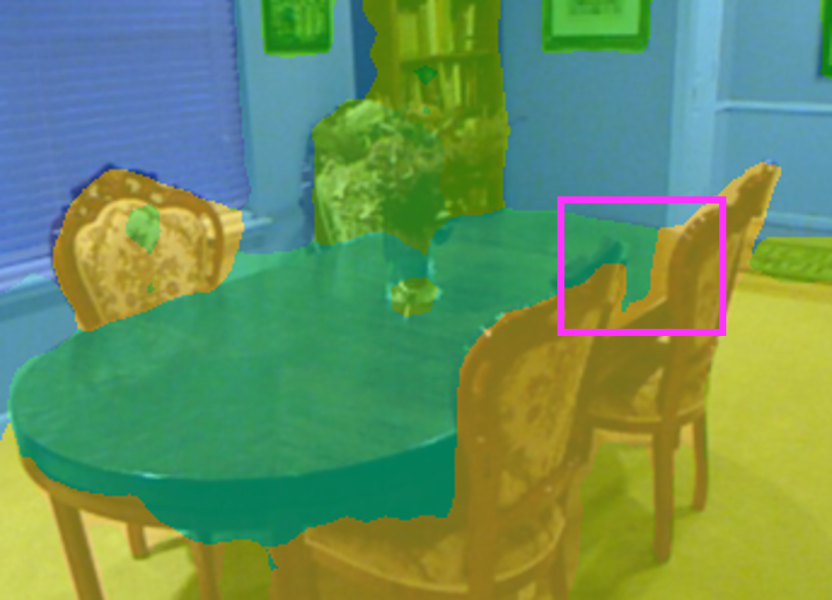}\\
        \vspace{1mm}
    \end{minipage}%
}
\subfigure[GT]{
    \begin{minipage}{0.19\linewidth}
        \centering
        \includegraphics[width=0.993\textwidth,height=0.7in]{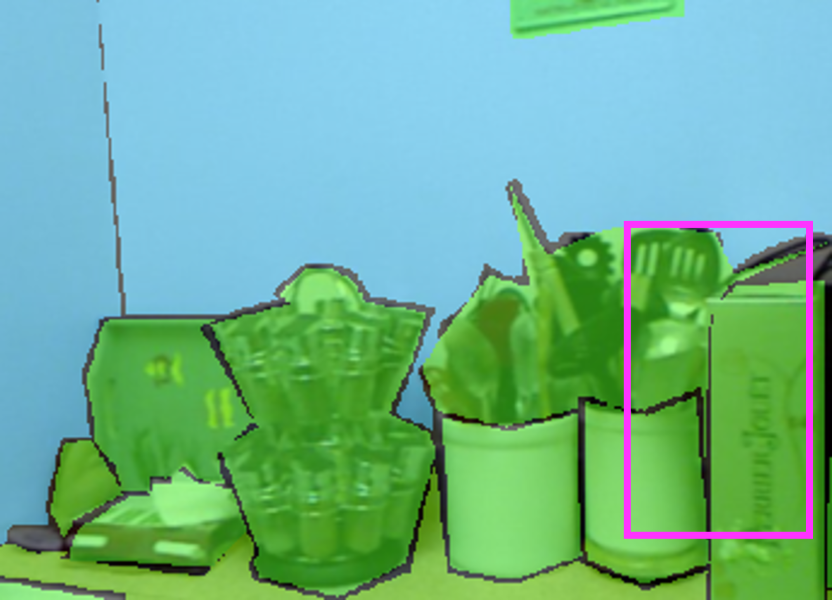}\\
        %\vspace{1mm}
        %\includegraphics[width=0.993\textwidth,height=0.7in]{images/nyu_seg/56_gt_box.png}\\
        \vspace{1mm}
        \includegraphics[width=0.993\textwidth,height=0.7in]{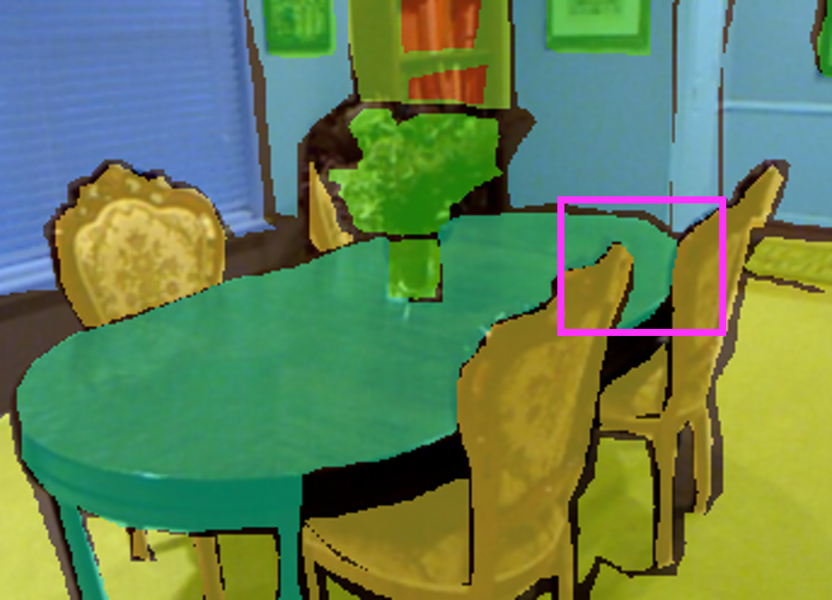}\\ 
        \vspace{1mm}
    \end{minipage}%
}%
\centering
\vspace{-3mm}
\caption{Visualization on NYUv2~\cite{nyuv2} with semantic segmentation as the primary task and other two tasks~(depth estimation and normal prediction) as auxiliary. The impressive improvements are marked with a purple box. }
\label{fig:nyu_seg}
\vspace{-6mm}
\end{figure*}

\subsection{Datasets and Baselines}

To substantiate the generalization and superiority of the proposed method, an evaluation is conducted using the standard multi-task learning benchmark on two datasets, 
\ie,  \textbf{NYUv2}~\cite{nyuv2}, \textbf{Cityscapes}~\cite{cityscapes} and \textbf{PASCAL-Context}~\cite{pascal_context}.
In these experiments, ResNet50 with dilated convolution is employed as the feature encoder, while a deeplab-like head~\cite{deeplabv3} is used as the decoders for dense prediction tasks.
Furthermore, the evaluation setting is enhanced by integrating two large-scale pre-trained models from ImageNet~\cite{imagenet} and COCO, thereby constructing two pseudo-tasks as additional auxiliary tasks on both \textbf{NYUv2}~\cite{nyuv2} and \textbf{Cityscapes}~\cite{cityscapes}.
Unlike other methods, the proposed approach capitalizes on noisy tasks to amplify primary task generalization.
Finally, in accordance with~\cite{autolambda}, the method is assessed on a more complex multi-domain benchmark, namely, Multi-CIFAR100~\cite{autolambda}. This benchmark consists of 20 tasks with distinct domain gaps between each other. The proposed method demonstrates its adaptability by effectively handling up to 20 tasks and fitting seamlessly within the multi-domain environment.

The method is evaluated alongside various state-of-the-art methods in two multi-task learning settings: i) all tasks are treated as primary tasks (referred to as the \texttt{Normal setting}), ii) only one tasks are designated as primary tasks (referred to as the \texttt{Auxiliary setting}). Further comparisons with IAL(Impartial Auxliary learning) under the auxiliary settings includes i) Gradient Cosine Similarity (GCS)~\cite{gcs},
ii) Online Learning with AUXiliary tasks (OL-AUX)~\cite{ol_aux}, and iii) Auto-lambda~\cite{autolambda}.
Additionally, the results of cutting-edge approaches in multi-task settings are provided, encompassing both loss- and gradient-based methods.

\subsection{Training Details}
By default, each task is considered the primary task, and the remaining tasks are treated as auxiliary tasks, unless labeled otherwise. 
In the experiments conducted, the task weightings were initialized to $0.1$, a small weighting that assumes all tasks are equally unrelated. 
The learning rate to update these weightings is hand-selected for each dataset. To ensure a fair comparison, the optimization strategies utilized in all baselines and the proposed method are consistent with respect to each dataset and data domain.
We adapt Auto-$\lambda$ and IMTL code into MTI-Net repository~\cite{mti, vanden_survey}, which has different model implement and data pre-process.

Concretely, two task categories are defined, \ie, classification and pixel-wise prediction, to evaluate the method. The pixel-wise prediction task includes semantic segmentation, depth/disparity estimation, surface normal prediction and part segmentation. Each category of task shares the same feature representation in the backbone model. ResNet~\cite{resnet} and Feature Pyramid Networks (FPN) are adopted as the backbone models.

Common practice is followed to design the feature representations for both the classification and segmentation tasks. Specifically, the C$5$ feature map is utilized for classification. If the multi-task model learns from multiple tasks in the same task category (\eg, semantic segmentation and depth prediction), each task possesses its own task-specific decoder without sharing parameters.

In our experiments, except for the baseline methods, all methods are able to adapt the weights for each tasks. We only tune the training strategy and some hyper-parameters as in their papers. The details about tuning hyper-parameters are shown below: 
For uniform, Uncertainty Weight, DWA, MGDA, PCGrad, RLW-Normal, ATW, Miraliev \etal, PUW, OL-AUX, GCS and IMTL, there is no hyper-parameter to tune. 
Specifically, PUW requires several epochs of warm-up to obtain a better starting point. The first 3\% of epochs are selected for warm-up.
For GradNorm, we follow the IMTL settings: tune the $\eta$ near $1.5$ for NYUv2 and $0.0$ for Cityscapes. 
For Auto-$\lambda$, the hyper-parameters is the initial weights of the tasks. Due to the long training time, we only tune the initial weights in $ 0.005, 0.01, 0.05, 0.1, 1.0$ and find the best setting is $0.01$ as reported in the original paper. 
We also tune the training strategy including optimizer, batch size and learning rate for NYUv2, Cityscapes, and PASCAL-Context. 
For NYUv2 and PASCAL-Context, the optimizer, learning rate and batch size are optimized with a grid search following ~\cite{vanden_survey}. The optimizer is tested between Adam and SGD with momentum 0.9. The learning rate and batch size are adjusted in 12 hyper-parameter settings in total. The batch size is selected from size 6 and 12. For SGD, the learning rate is searched in \{ 1e-3, 5e-4, 1e-2, 5e-2 \}. For Adam, the learning rate is searched in \{ 1e-4, 5e-4 \}. In practice, Adam with the learning rate 1e-4 and batch size 12 is the best combination in most experiments.
For cityscapes, the learning rate is simply set to $3e-5$  and the optimizer is SGD as Auto-$\lambda$.

\subsection{Dense Prediction Benchmarks}

The proposed methods were evaluated for dense prediction tasks on three standard multi-task datasets, \ie, \textbf{NYUv2}~\cite{nyuv2},\textbf{Cityscapes}~\cite{cityscapes} and \textbf{PASCAL-Context}~\cite{pascal_context}, in a single-domain setting. 
For \textbf{NYUv2}, the evaluation encompassed three tasks, \ie, $13$-class semantic segmentation, depth prediction, and surface normal prediction, following the same experimental setting as in \cite{uw}.
For \textbf{Cityscapes}, the method was evaluated on three tasks, \ie, $19$-class semantic segmentation, disparity (inverse depth) estimation, and a recently proposed $10$-class part segmentation, aligned with the experimental setting in \cite{autolambda}.
\textbf{PASCAL} is a comprehensive dataset providing images of both indoor and outdoor scenes. There are 4,998 training images and 5,105 testing images, with labels of semantic segmentation, human parsing, and object boundary detection. 

\subsubsection{Evaluation Metrics}
We evaluated our method on segmentation~(including semantic segmentation and part segmentation), depth, and normals via three metrics, \ie,  mean intersection over union (mIoU), root MSE (rMSE), and mean angle distances (mDist), respectively.
Following \cite{vanden_survey}, the overall relative improvement $\Delta MTL$ on all task-specific metrics $\{ M_{m, i} \}_{i \in T}$ of the final multi-task model $m$ is also reported with respect to the metric  $M_{s_i}$ of the single-task counterpart $\{ s_i \}_{i\in T}$:

\begin{equation}
\Delta MTL = \frac{1}{T} \sum\limits_{i=1}^{T}(-1)^{l_i}(M_{m,i}-M_{s_i})/M_{s_i}.
\label{eq:theta_mtl}
\end{equation}
For the task $l_i = 1$, the lower values of $M_{\cdot,i}$ (classification accuracies), the better performance. For the task $l_i = 0$, the higher values of $M_{\cdot,i}$(L1 errors for depth prediction), the better performance.

We report the mean and standard deviation of the experimental results, which are calculated from five runs using different random seeds.
\begin{figure*}[!htbp]
\centering
\subfigure[Image]{
    \begin{minipage}{0.19\linewidth}
        \centering
        \includegraphics[width=0.993\textwidth,height=0.7in]{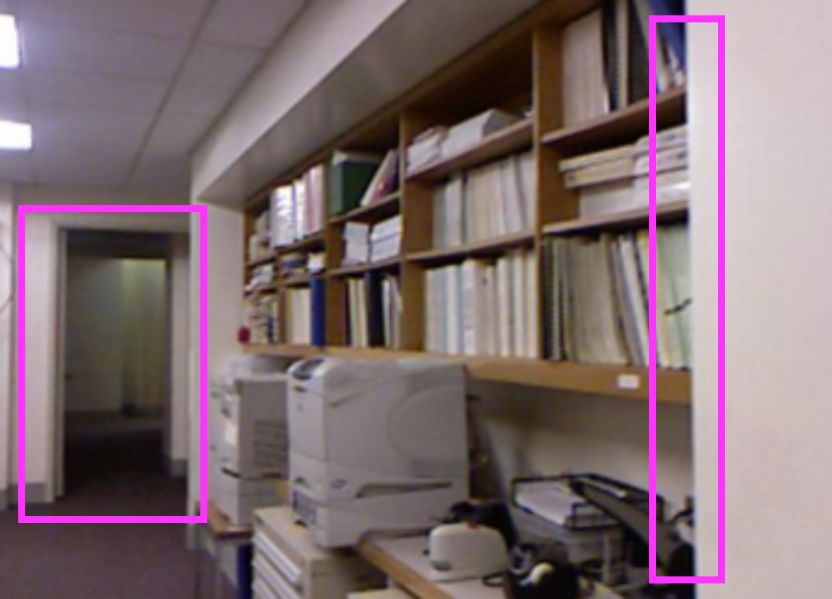}\\
        %\vspace{1mm}
        %\includegraphics[width=0.993\textwidth,height=0.7in]{images/nyu_depth/38_origin_box.png}\\
        \vspace{1mm}
        \includegraphics[width=0.993\textwidth,height=0.7in]{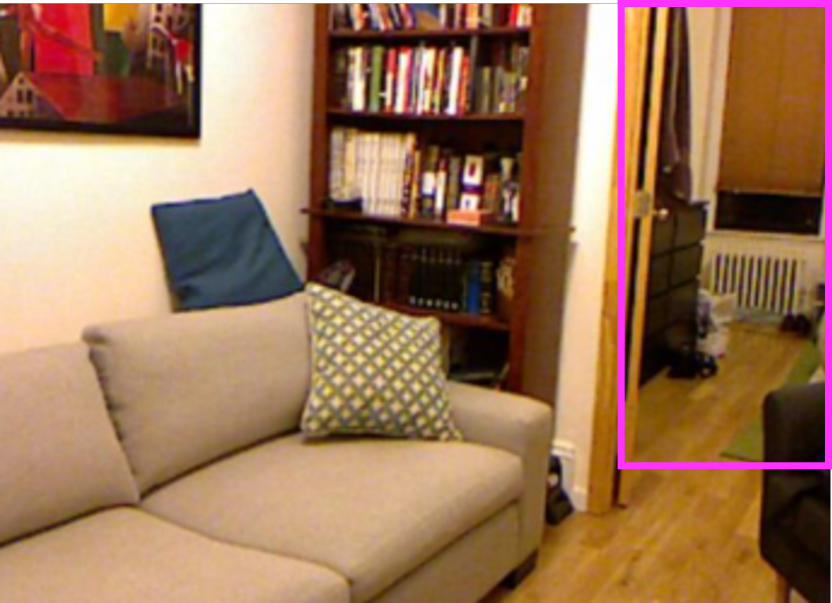}\\     
        \vspace{1mm}
        % \caption{}
    \end{minipage}%
}%
\subfigure[OL$\text{-} $AUX~\cite{ol_aux}]{
    \begin{minipage}{0.19\linewidth}
        \centering
        \includegraphics[width=0.993\textwidth,height=0.7in]{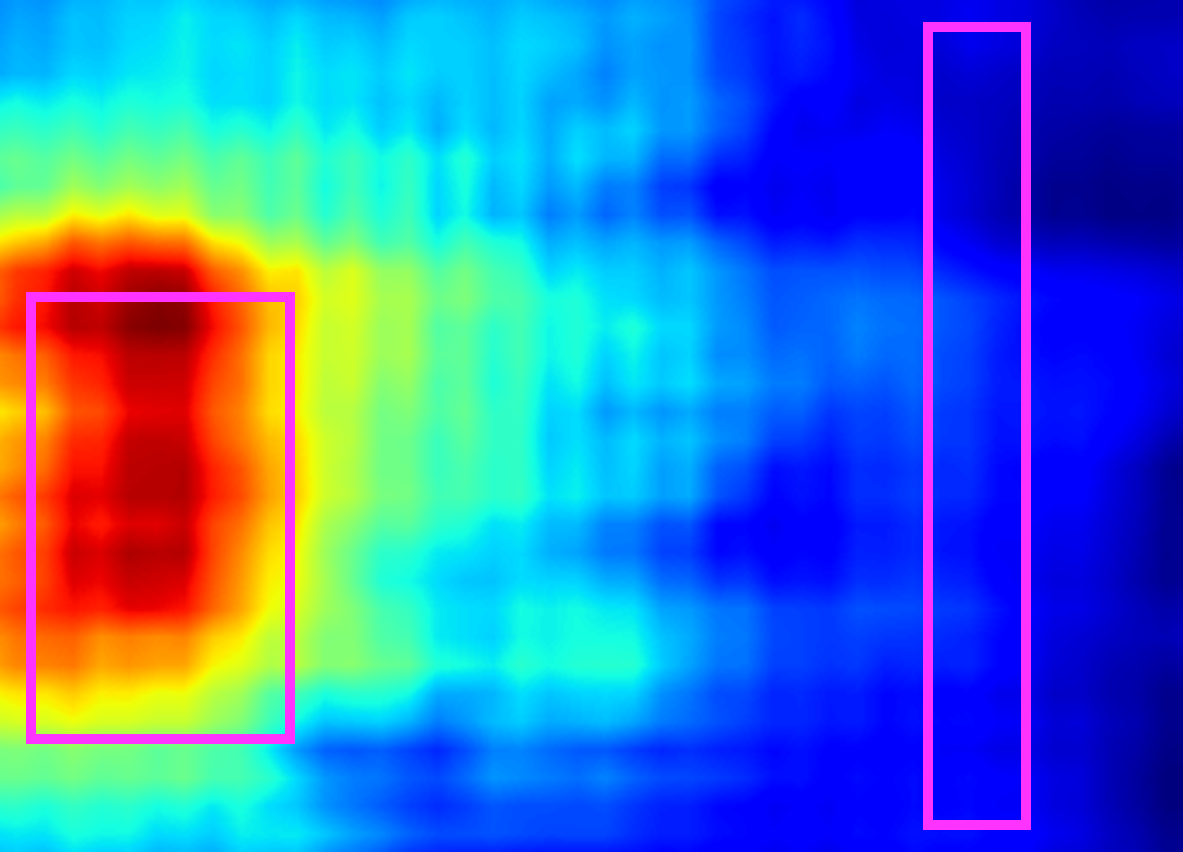}\\
        %\vspace{1mm}
        %\includegraphics[width=0.993\textwidth,height=0.7in]{images/nyu_depth/38_OL-AUX_box.png}\\
        \vspace{1mm}
        \includegraphics[width=0.993\textwidth,height=0.7in]{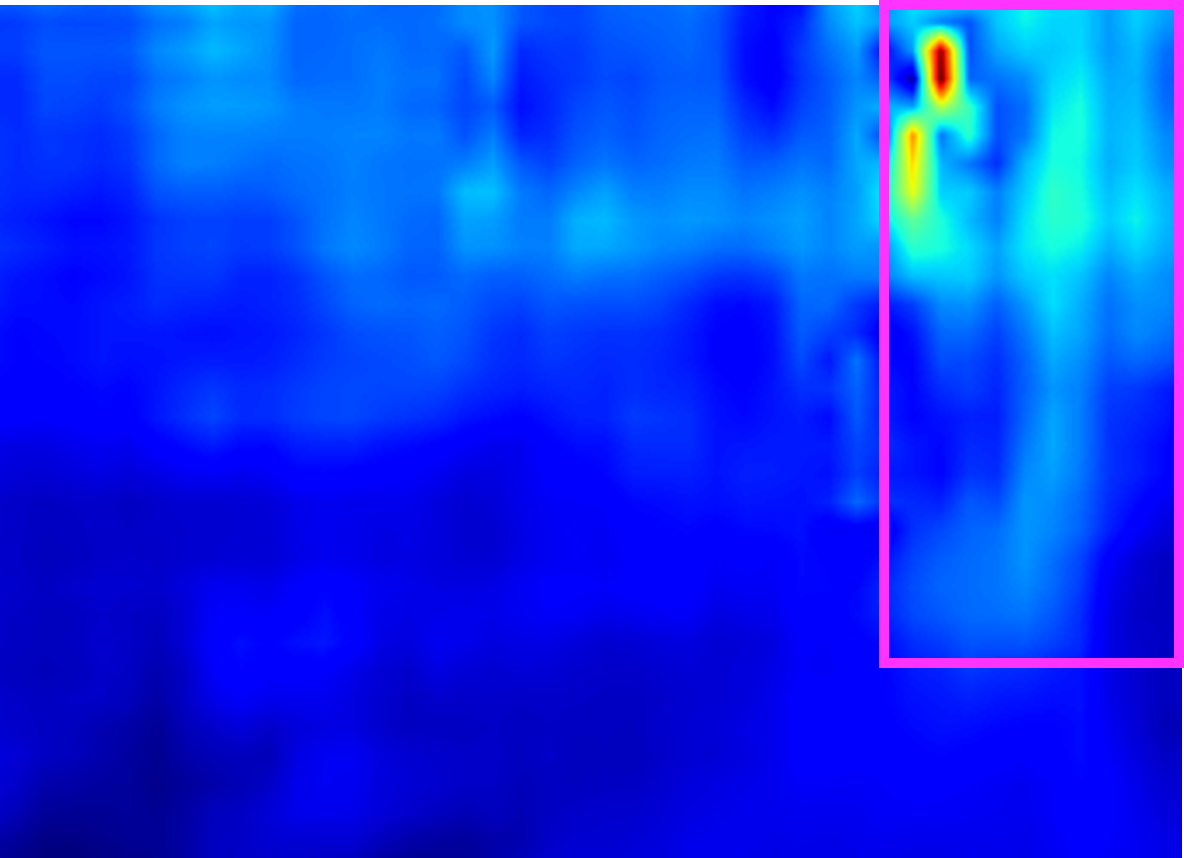}\\   
        \vspace{1mm}
    \end{minipage}%
}%
\subfigure[Auto$\text{-}\lambda $~\cite{autolambda}]{
    \begin{minipage}{0.19\linewidth}
        \centering
        \includegraphics[width=0.993\textwidth,height=0.7in]{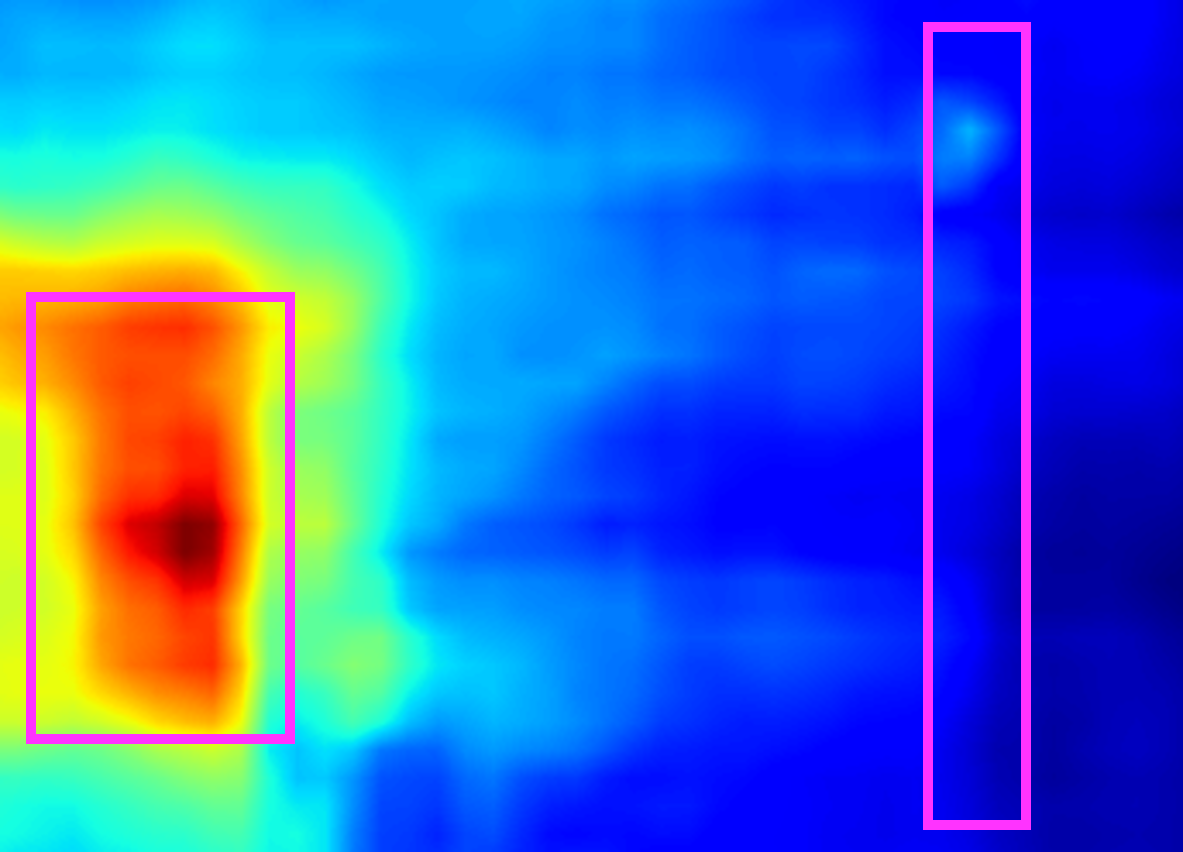}\\
        %\vspace{1mm}
        %\includegraphics[width=0.993\textwidth,height=0.7in]{images/nyu_depth/38_auto-lambda_box.png}\\
        \vspace{1mm}
        \includegraphics[width=0.993\textwidth,height=0.7in]{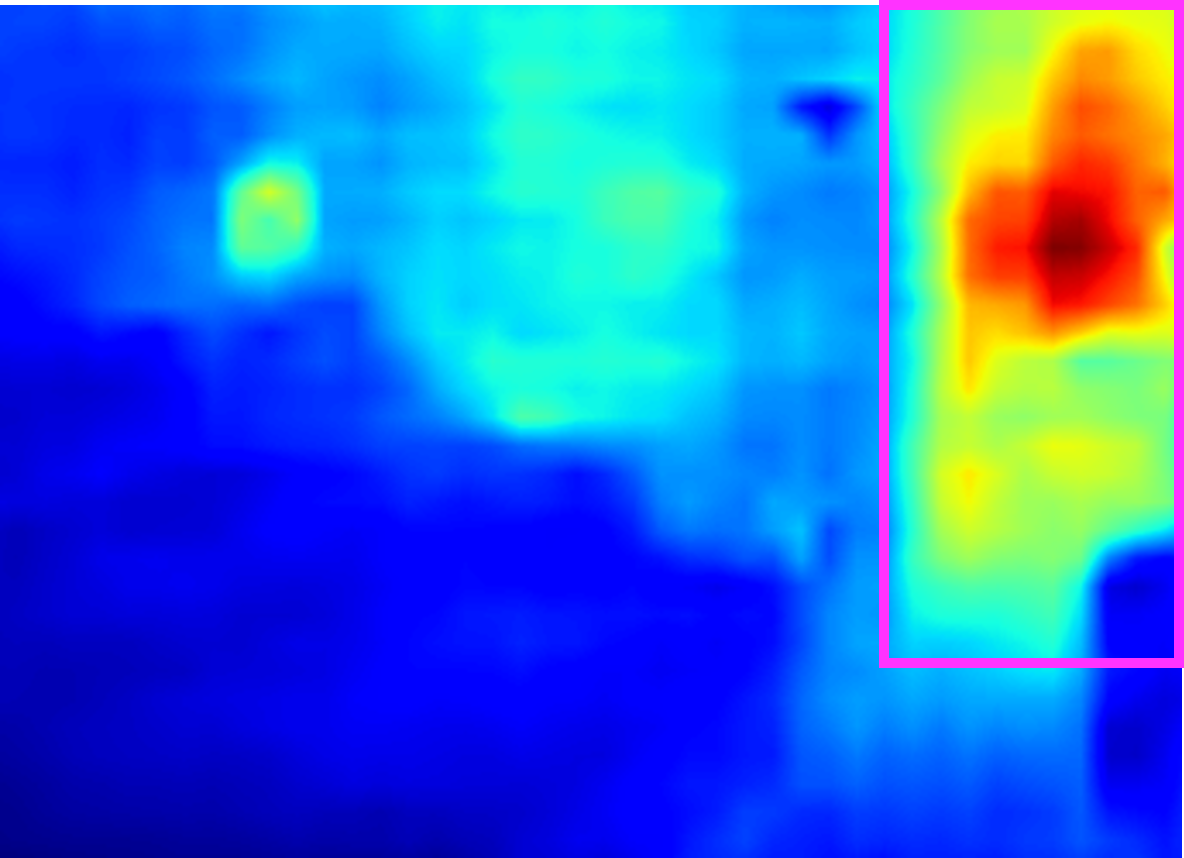}\\  
        \vspace{1mm}
    \end{minipage}%
}%
\subfigure[Ours]{
    \begin{minipage}{0.19\linewidth}
        \centering
        \includegraphics[width=0.993\textwidth,height=0.7in]{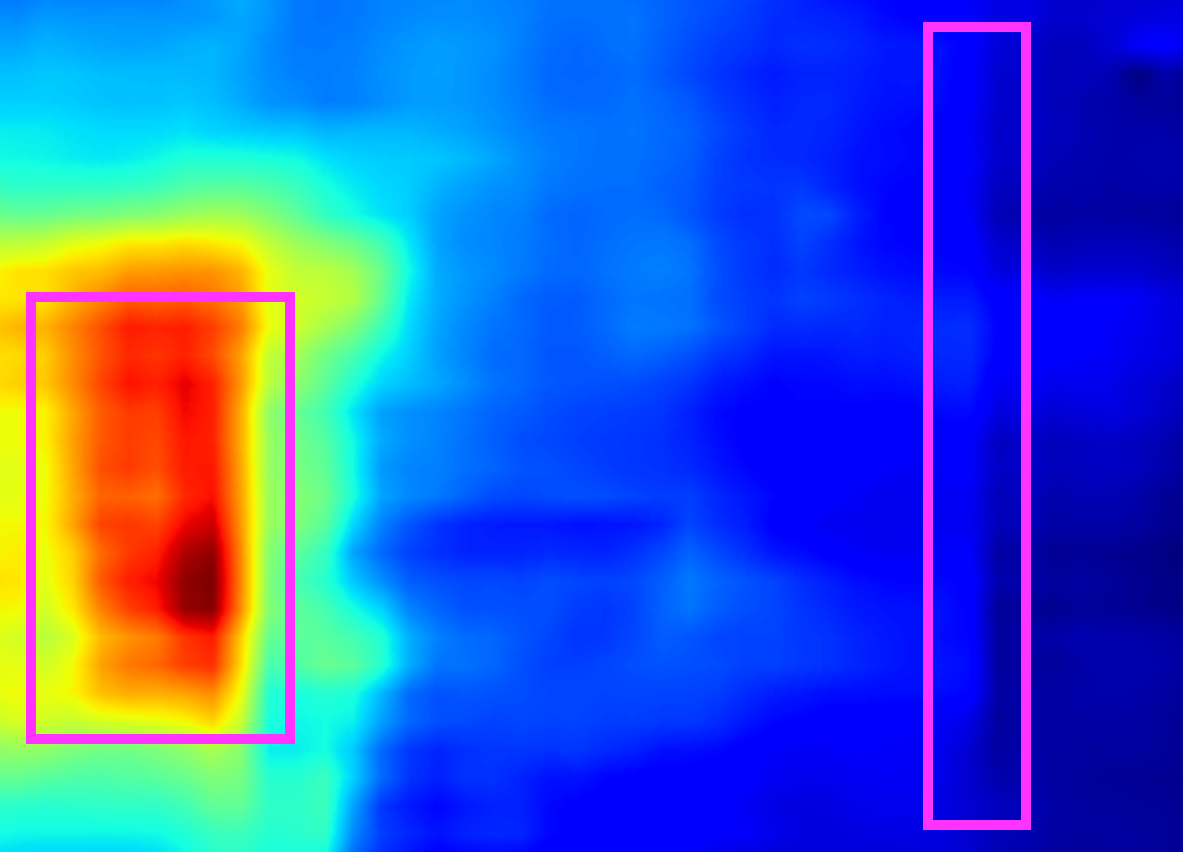}\\
        %\vspace{1mm}
        %\includegraphics[width=0.993\textwidth,height=0.7in]{images/nyu_depth/38_ours_box.png}\\
        \vspace{1mm}
        \includegraphics[width=0.993\textwidth,height=0.7in]{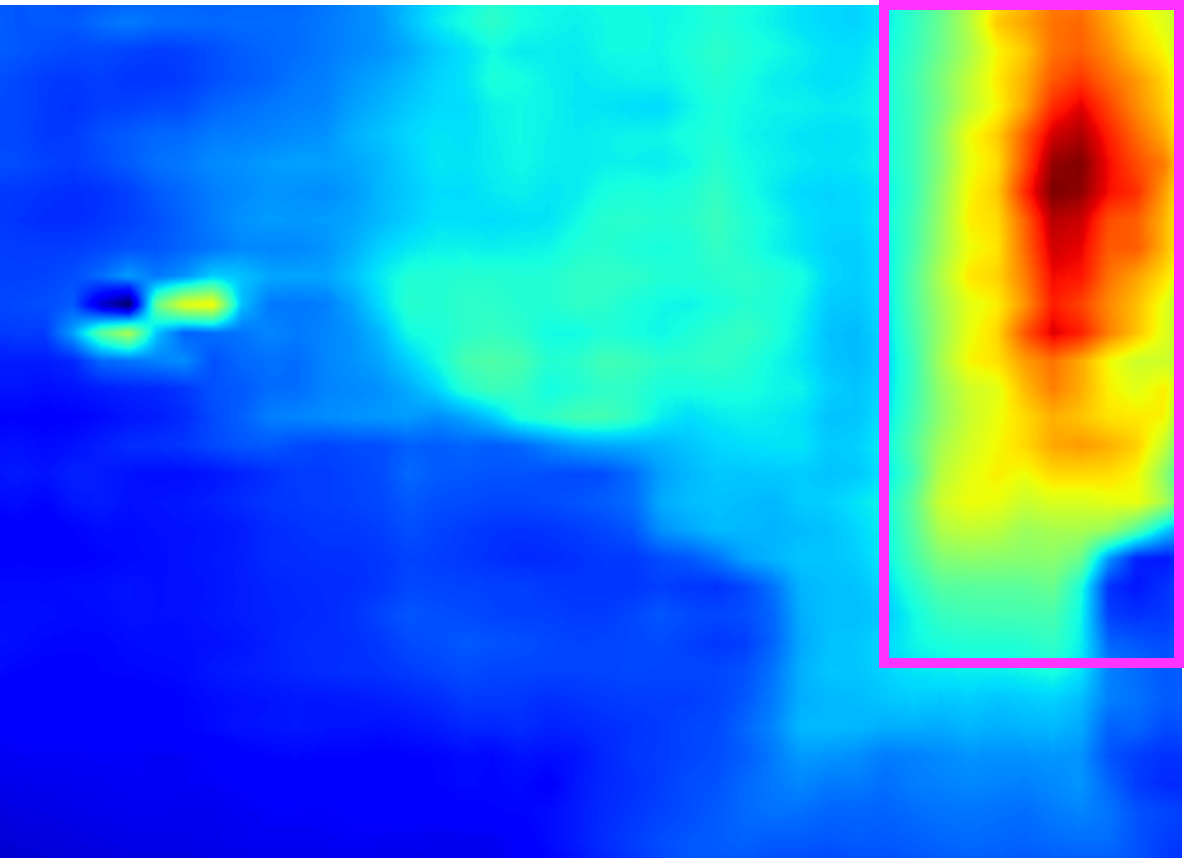}\\ 
        \vspace{1mm}
    \end{minipage}%
}%
\subfigure[GT]{
    \begin{minipage}{0.19\linewidth}
        \centering
        \includegraphics[width=0.993\textwidth,height=0.7in]{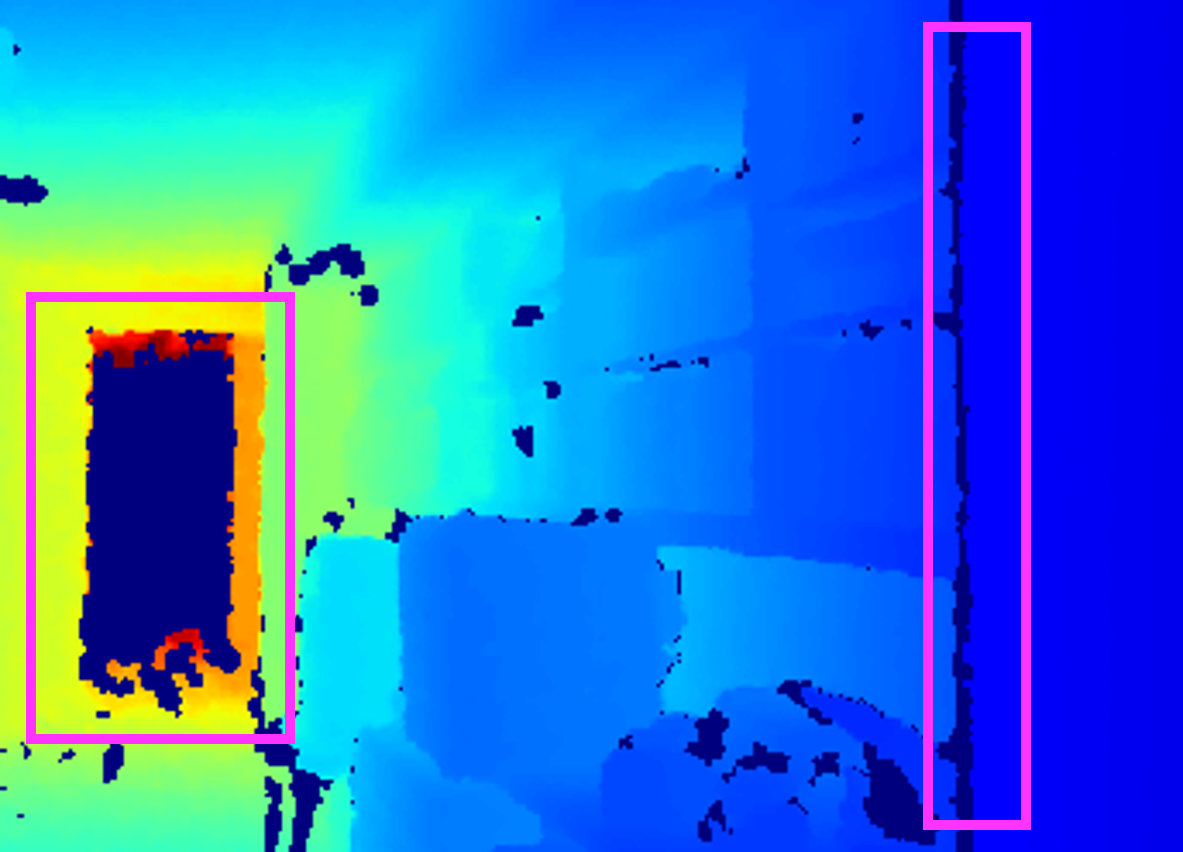}\\
        %\vspace{1mm}
        %\includegraphics[width=0.993\textwidth,height=0.7in]{images/nyu_depth/38_gt_box.png}\\
        \vspace{1mm}
        \includegraphics[width=0.993\textwidth,height=0.7in]{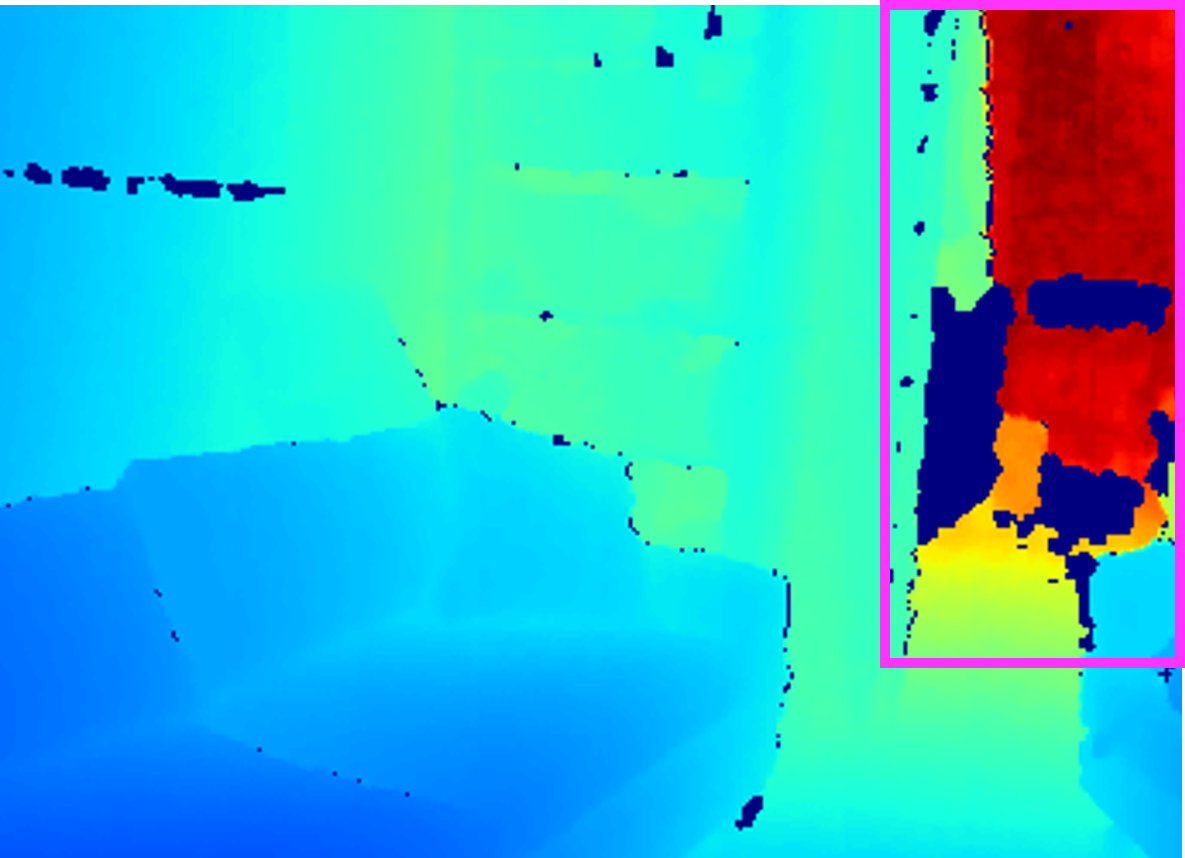}\\ 
        \vspace{1mm}
    \end{minipage}%
}%
\centering
    \vspace{-3mm}
\caption{Visualization on NYUv2~\cite{nyuv2} with depth estimation as the primary task and other two tasks~(semantic segmentation and normal prediction) as auxiliary. The impressive improvements are marked with a purple box.  }
\label{fig:nyu_depth}
\vspace{-6mm}
\end{figure*}
\begin{figure*}[!htbp]
\centering
\subfigure[Image]{
    \begin{minipage}{0.19\linewidth}
        \centering
        \includegraphics[width=0.993\textwidth,height=0.7in]{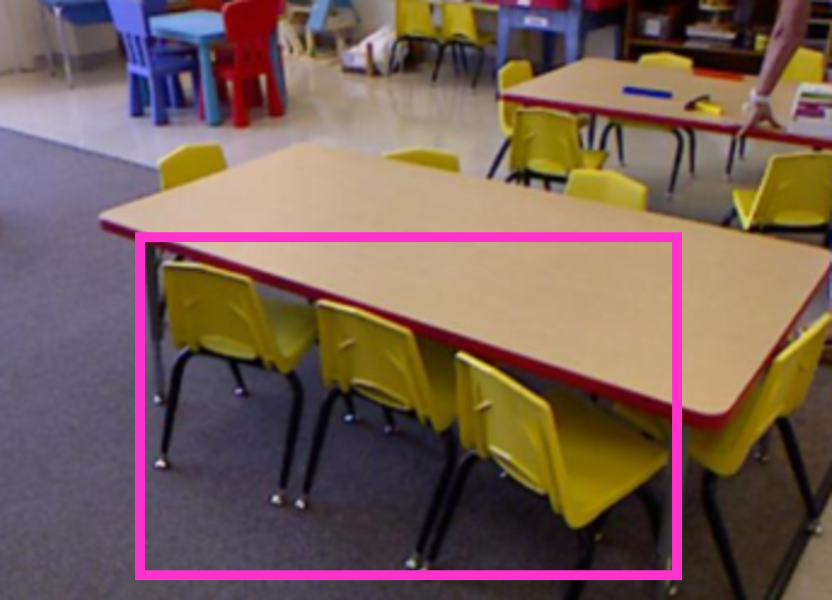}\\
        %\vspace{1mm}
        %\includegraphics[width=0.993\textwidth,height=0.7in]{images/nyu_normal/80_origin_box.png}\\
        \vspace{1mm}
        \includegraphics[width=0.993\textwidth,height=0.7in]{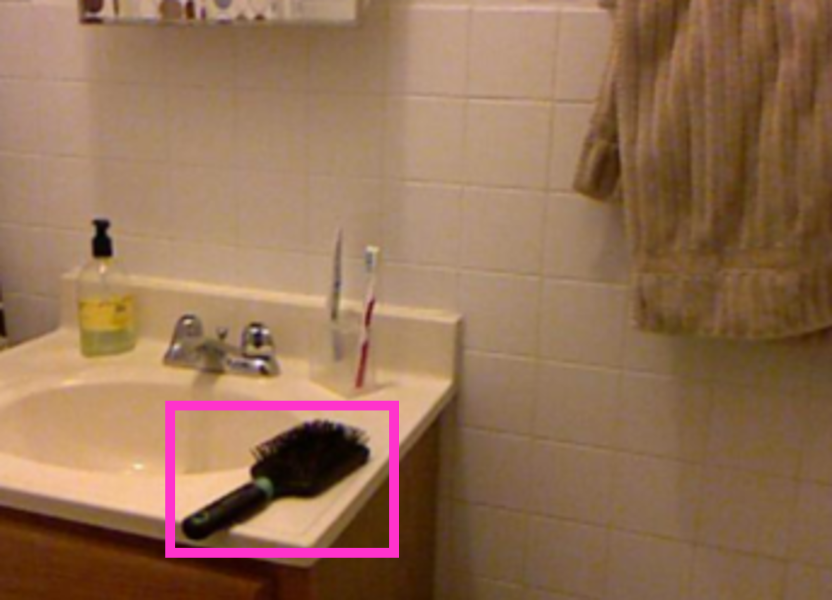}\\     
        \vspace{1mm}
        % \caption{}
    \end{minipage}%
}%
\subfigure[OL$\text{-} $AUX~\cite{ol_aux}]{
    \begin{minipage}{0.19\linewidth}
        \centering
        \includegraphics[width=0.993\textwidth,height=0.7in]{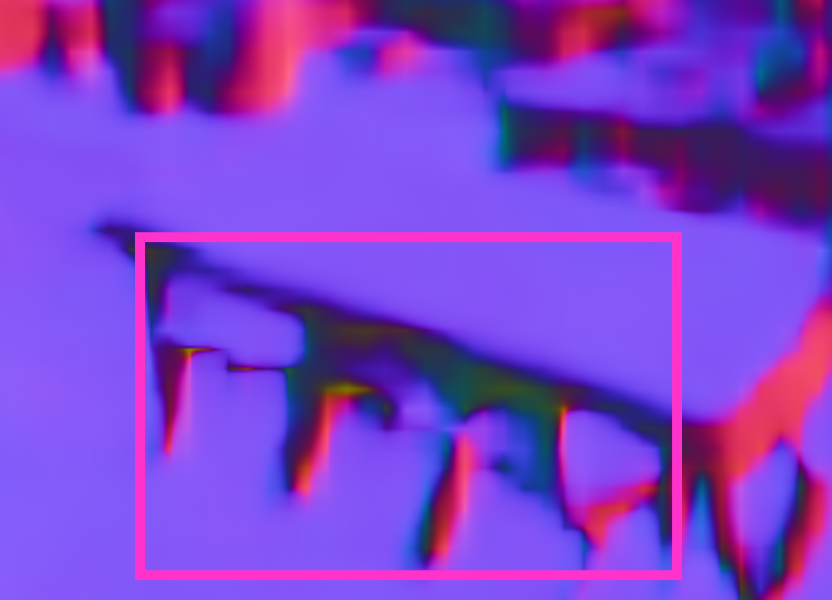}\\
        %\vspace{1mm}
        %\includegraphics[width=0.993\textwidth,height=0.7in]{images/nyu_normal/80_OL-AUX_box.png}\\
        \vspace{1mm}
        \includegraphics[width=0.993\textwidth,height=0.7in]{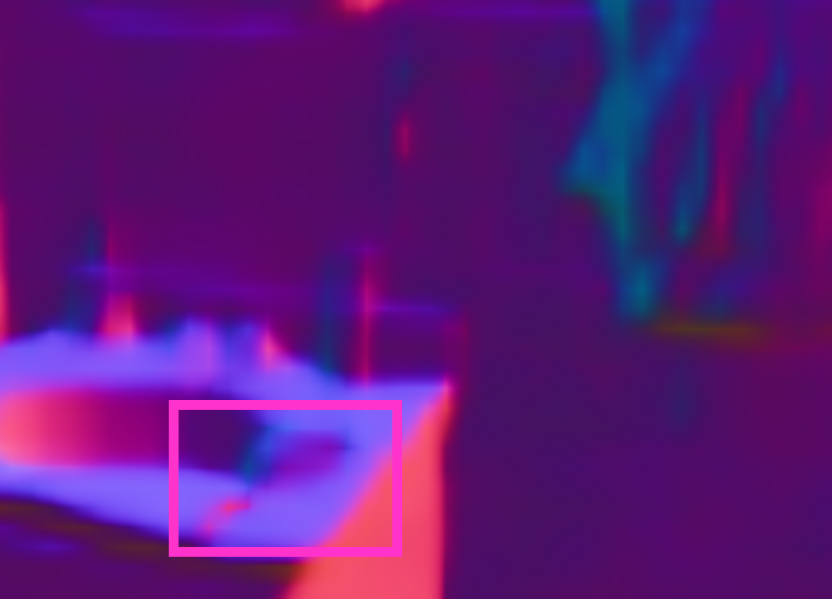}\\   
        \vspace{1mm}
    \end{minipage}%
}%
\subfigure[Auto$\text{-}\lambda $~\cite{autolambda}]{
    \begin{minipage}{0.19\linewidth}
        \centering
        \includegraphics[width=0.993\textwidth,height=0.7in]{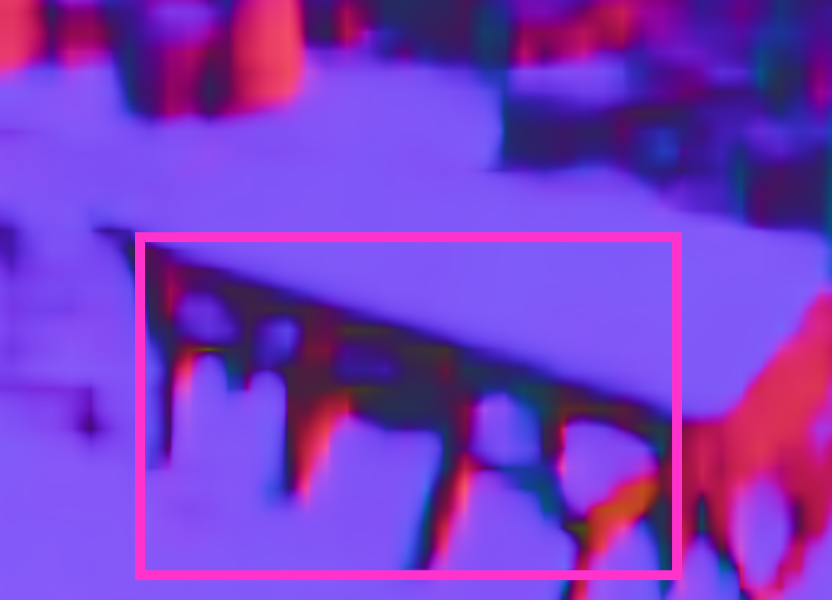}\\
        %\vspace{1mm}
        %\includegraphics[width=0.993\textwidth,height=0.7in]{images/nyu_normal/80_auto-lambda_box.png}\\
        \vspace{1mm}
        \includegraphics[width=0.993\textwidth,height=0.7in]{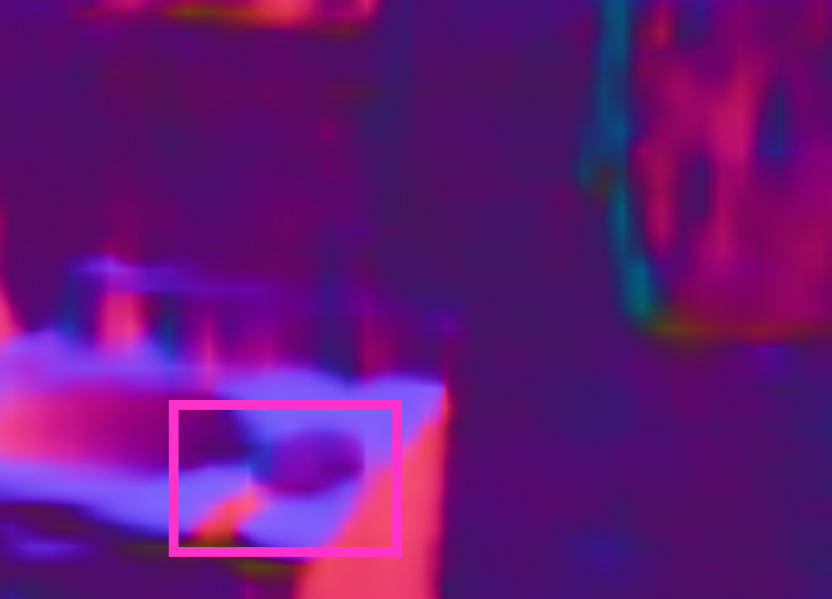}\\  
        \vspace{1mm}
    \end{minipage}%
}%
\subfigure[Ours]{
    \begin{minipage}{0.19\linewidth}
        \centering
        \includegraphics[width=0.993\textwidth,height=0.7in]{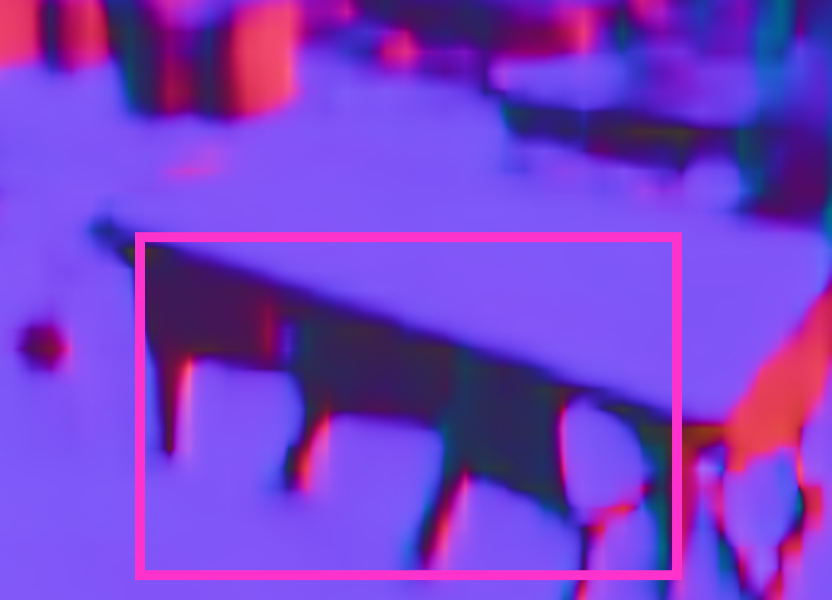}\\
        %\vspace{1mm}
        %\includegraphics[width=0.993\textwidth,height=0.7in]{images/nyu_normal/80_ours_box.png}\\
        \vspace{1mm}
        \includegraphics[width=0.993\textwidth,height=0.7in]{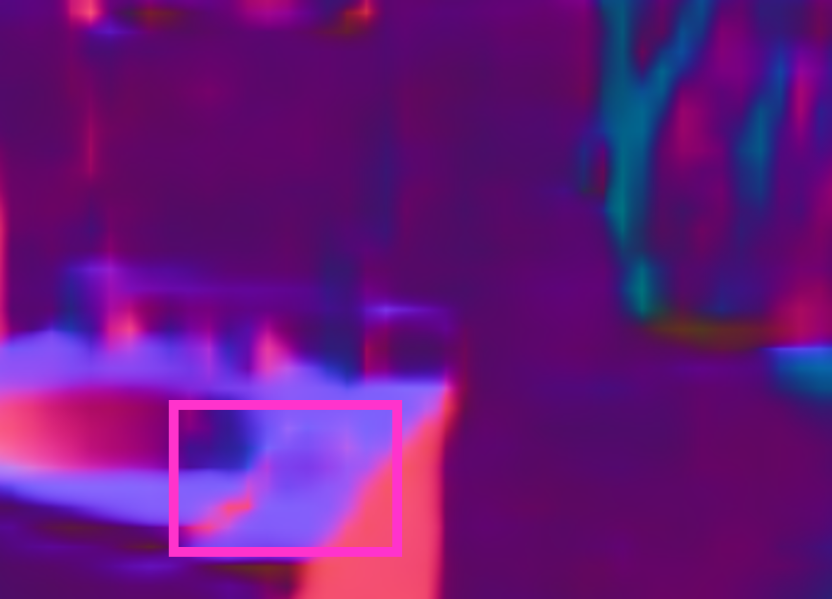}\\ 
        \vspace{1mm}
    \end{minipage}%
}%
\subfigure[GT]{
    \begin{minipage}{0.19\linewidth}
        \centering
        \includegraphics[width=0.993\textwidth,height=0.7in]{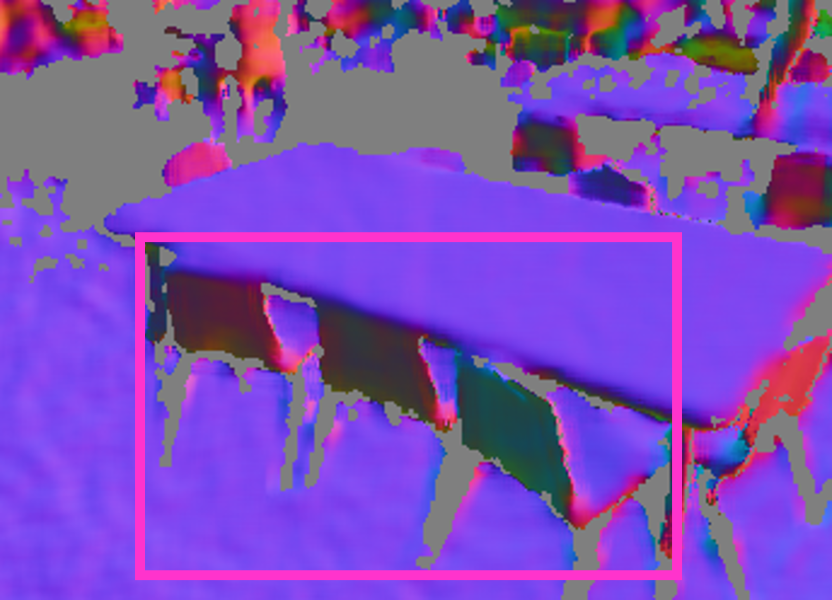}\\
        %\vspace{1mm}
        %\includegraphics[width=0.993\textwidth,height=0.7in]{images/nyu_normal/80_gt_box.png}\\
        \vspace{1mm}
        \includegraphics[width=0.993\textwidth,height=0.7in]{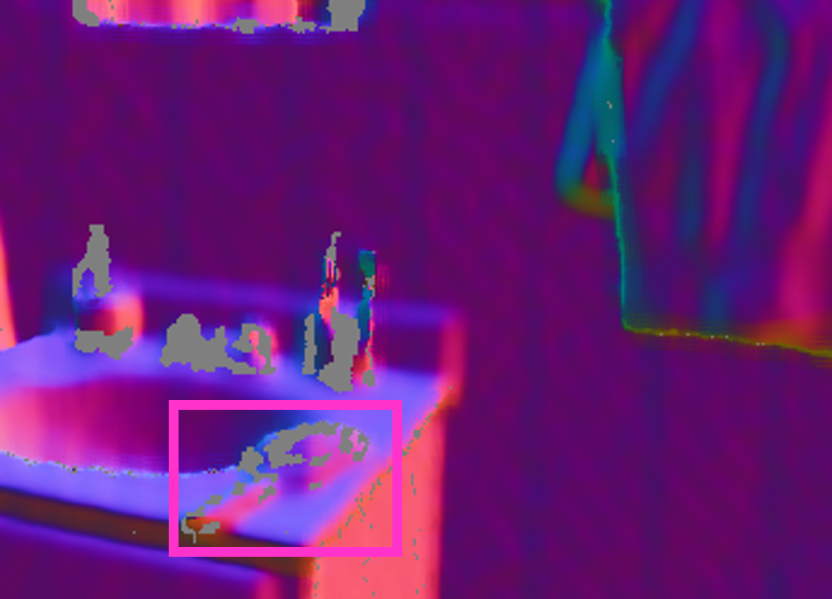}\\ 
        \vspace{1mm}
    \end{minipage}%
}%
\centering
\vspace{-3mm}
\caption{Visualization on NYUv2~\cite{nyuv2} with surface normal prediction as the primary task and other two tasks~(semantic segmentation and depth estimation) as auxiliary. The impressive improvements are marked with a purple box.  }
\vspace{-6mm}
\label{fig:nyu_normal}
\end{figure*}

\subsubsection{Results on NYUv2}
Table~\ref{table:dense_prediction} summarizes the results on the \textbf{NYUv2}. We set one task as the primary task and the other two tasks as auxiliary tasks. It can be seen from this table that our method outperforms all baselines in multi-task settings. Concretely, IMTL~\cite{imtl}, Auto-$\lambda$~\cite{autolambda} and PUW~\cite{PUM2023} reach the best performance among all the baselines in the multi-task learning setting. However, IMTL outperforms auto-$\lambda$ with regard to surface normal prediction and semantic segmentation. This shows that standard multi-task optimization is susceptible to negative transfer, whereas our method can avoid negative transfer due to its ability to minimize the negative effect brought by tasks that do not assist with the primary task. As such, our proposed method not only reaches the highest relative improvement among all tasks but also shows the best performance on each task under the \texttt{Auxiliary setting}.

In Fig.~\ref{fig:nyu_seg}-\ref{fig:nyu_normal}, we provide some visualization on NYUv2~\cite{nyuv2}. In Figure~\ref{fig:nyu_seg}, we visualize a comparison between our approach and existing methods for auxiliary learning.
The impressive outperforming aspects are highlighted with pink boxes. By leveraging auxiliary tasks which are fully trained such as depth estimation and normal prediction, our method achieves accurate dense predictions for objects with similar depths without introducing contamination from other categories. Moreover, our approach excels at distinguishing objects with different depths, effectively separating them. 
In Fig.~\ref{fig:nyu_depth}, we present a visualization of the depth estimation task as the primary task on the NYUv2~\cite{nyuv2} dataset. By incorporating a well-trained semantic segmentation as an auxiliary task, we observe that our method is capable of predicting more accurate depth values along the classification surfaces of objects at different depths.  
Figure~\ref{fig:nyu_normal} presents the visual outcomes of employing surface normal prediction as the primary task. Our results conspicuously reveal a better illustration of objects that have been overlooked or inaccurately predicted by alternative methods.

\subsubsection{Results on Cityscapes}
Here, we aim to evaluate our method using a more complex outdoor scenario, \ie,  \textbf{Cityscapes}~\cite{cityscapes}, which may have a more complex imbalance problem and a different relationship between primary and auxiliary tasks.
Specifically, \textbf{Cityscapes}~\cite{cityscapes} contains nearly $5,000$ fine-annotated video frames shot in over $50$ urban cities of different countries. This dataset has various annotations, including disparity maps (used for depth regression), panoptic segmentation, and detection. 
We follow the standard setting, which splits $5,000$ images into $2,975$, $500$, and $1,525$ for training, validation, and testing, respectively. Our experiments perform on three tasks with \textbf{Cityscapes}~\cite{cityscapes}: depth regression, $19$-class semantic segmentation, and $10$-class part segmentation, including parts of humans and cars following the benchmarks arising in \cite{autolambda}. 

\begin{figure*}[!htbp]
\centering
\subfigure[Image]{
    \begin{minipage}{0.19\linewidth}
        \centering
        \includegraphics[width=0.993\textwidth,height=0.7in]{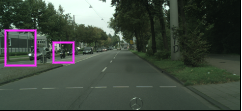}\\
        %\vspace{1mm}
        %\includegraphics[width=0.993\textwidth,height=0.7in]{images/seg/323_origin_box.png}\\
        \vspace{1mm}
        \includegraphics[width=0.993\textwidth,height=0.7in]{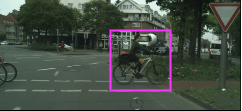}\\     
        \vspace{1mm}
        % \caption{}
    \end{minipage}%
}%
\subfigure[OL$\text{-} $AUX~\cite{ol_aux}]{
    \begin{minipage}{0.19\linewidth}
        \centering
        \includegraphics[width=0.993\textwidth,height=0.7in]{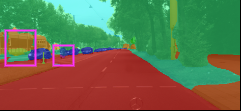}\\
        %\vspace{1mm}
        %\includegraphics[width=0.993\textwidth,height=0.7in]{images/seg/323_OL-AUX_box.png}\\
        \vspace{1mm}
        \includegraphics[width=0.993\textwidth,height=0.7in]{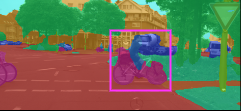}\\  
        \vspace{1mm}
    \end{minipage}%
}%
\subfigure[Auto$\text{-}\lambda $~\cite{autolambda}]{
    \begin{minipage}{0.19\linewidth}
        \centering
        \includegraphics[width=0.993\textwidth,height=0.7in]{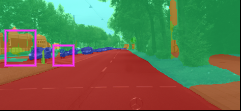}\\
        %\vspace{1mm}
        %\includegraphics[width=0.993\textwidth,height=0.7in]{images/seg/323_auto-lambda_box.png}\\
        \vspace{1mm}
        \includegraphics[width=0.993\textwidth,height=0.7in]{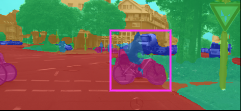}\\  
        \vspace{1mm}
    \end{minipage}%
}%
\subfigure[Ours]{
    \begin{minipage}{0.19\linewidth}
        \centering
        \includegraphics[width=0.993\textwidth,height=0.7in]{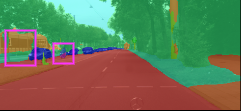}\\
        %\vspace{1mm}
        %\includegraphics[width=0.993\textwidth,height=0.7in]{images/seg/323_ours_box.png}\\
        \vspace{1mm}
        \includegraphics[width=0.993\textwidth,height=0.7in]{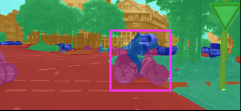}\\ 
        \vspace{1mm}
    \end{minipage}%
}%
\subfigure[GT]{
    \begin{minipage}{0.19\linewidth}
        \centering
        \includegraphics[width=0.993\textwidth,height=0.7in]{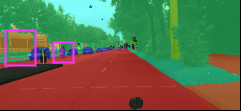}\\
        %\vspace{1mm}
        %\includegraphics[width=0.993\textwidth,height=0.7in]{images/seg/323_gt_box.png}\\
        \vspace{1mm}
        \includegraphics[width=0.993\textwidth,height=0.7in]{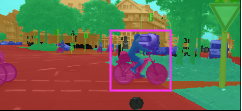}\\ 
        \vspace{1mm}
    \end{minipage}%
}%
\centering
    \vspace{-3mm}
\caption{Visualization on Cityscapes~\cite{cityscapes} with semantic segmentation as the primary task and other two tasks~(disparity estimation and part segmentation) as auxiliary. The impressive improvements are marked with a purple box.  }
\vspace{-6mm}
\label{fig:cityscapes_seg}
\end{figure*}

\begin{figure*}[!htbp]
\centering
\subfigure[Image]{
    \begin{minipage}{0.19\linewidth}
        \centering
        \includegraphics[width=0.993\textwidth,height=0.7in]{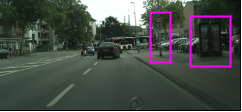}\\
        %\vspace{1mm}
        %\includegraphics[width=0.993\textwidth,height=0.7in]{images/disp/294_origin_box.png}\\
        \vspace{1mm}
        \includegraphics[width=0.993\textwidth,height=0.7in]{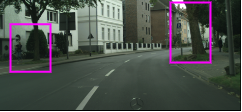}\\ 
        \vspace{1mm}
        % \caption{}
    \end{minipage}%
}%
\subfigure[OL$\text{-} $AUX~\cite{ol_aux}]{
    \begin{minipage}{0.19\linewidth}
        \centering
        \includegraphics[width=0.993\textwidth,height=0.7in]{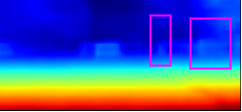}\\
        %\vspace{1mm}
        %\includegraphics[width=0.993\textwidth,height=0.7in]{images/disp/294_OL-AUX_box.png}\\
        \vspace{1mm}
        \includegraphics[width=0.993\textwidth,height=0.7in]{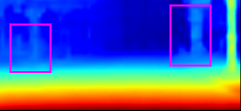}\\   
        \vspace{1mm}
    \end{minipage}%
}%
\subfigure[Auto$\text{-}\lambda $~\cite{autolambda}]{
    \begin{minipage}{0.19\linewidth}
        \centering
        \includegraphics[width=0.993\textwidth,height=0.7in]{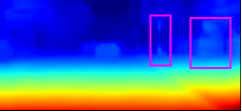}\\
        %\vspace{1mm}
        %\includegraphics[width=0.993\textwidth,height=0.7in]{images/disp/294_auto-lambda_box.png}\\
        \vspace{1mm}
        \includegraphics[width=0.993\textwidth,height=0.7in]{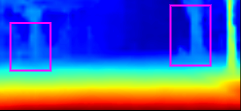}\\  
        \vspace{1mm}
    \end{minipage}%
}%
\subfigure[Ours]{
    \begin{minipage}{0.19\linewidth}
        \centering
        \includegraphics[width=0.993\textwidth,height=0.7in]{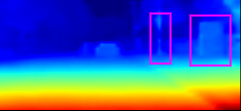}\\
        %\vspace{1mm}
        %\includegraphics[width=0.993\textwidth,height=0.7in]{images/disp/294_ours_box.png}\\
        \vspace{1mm}
        \includegraphics[width=0.993\textwidth,height=0.7in]{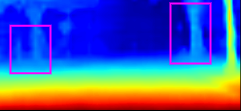}\\ 
        \vspace{1mm}
    \end{minipage}%
}%
\subfigure[GT]{
    \begin{minipage}{0.19\linewidth}
        \centering
        \includegraphics[width=0.993\textwidth,height=0.7in]{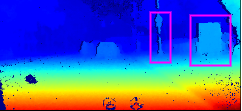}\\
        %\vspace{1mm}
        %\includegraphics[width=0.993\textwidth,height=0.7in]{images/disp/294_gt_box.png}\\
        \vspace{1mm}
        \includegraphics[width=0.993\textwidth,height=0.7in]{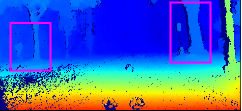}\\ 
        \vspace{1mm}
    \end{minipage}%
}%
\centering
\vspace{-3mm}
\caption{Visualization on Cityscapes~\cite{cityscapes} with disparity estimation as the primary task and other two tasks~(semantic segmentation and part segmentation) as auxiliary. The impressive improvements are marked with a purple box.  }
\label{fig:cityscapes_disp}
\vspace{-6mm}
\end{figure*}

\begin{figure*}[!htbp]
\centering
\subfigure[Image]{
    \begin{minipage}{0.19\linewidth}
        \centering
        \includegraphics[width=0.993\textwidth,height=0.7in]{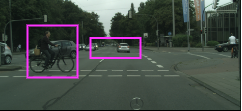}\\
        %\vspace{1mm}
        %\includegraphics[width=0.993\textwidth,height=0.7in]{images/partseg/159_origin_box.png}\\
        \vspace{1mm}
        \includegraphics[width=0.993\textwidth,height=0.7in]{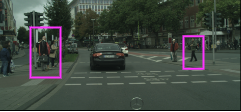}\\     
        \vspace{1mm}
        % \caption{}
    \end{minipage}%
}%
\subfigure[OL$\text{-} $AUX~\cite{ol_aux}]{
    \begin{minipage}{0.19\linewidth}
        \centering
        \includegraphics[width=0.993\textwidth,height=0.7in]{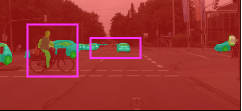}\\
        %\vspace{1mm}
        %\includegraphics[width=0.993\textwidth,height=0.7in]{images/partseg/159_OL-AUX_box.png}\\
        \vspace{1mm}
        \includegraphics[width=0.993\textwidth,height=0.7in]{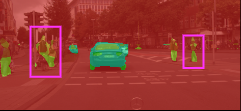}\\   
        \vspace{1mm}
    \end{minipage}%
}%
\subfigure[Auto$\text{-}\lambda $~\cite{autolambda}]{
    \begin{minipage}{0.19\linewidth}
        \centering
        \includegraphics[width=0.993\textwidth,height=0.7in]{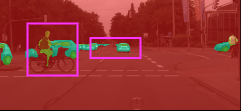}\\
        %\vspace{1mm}
        %\includegraphics[width=0.993\textwidth,height=0.7in]{images/partseg/159_auto-lambda_box.png}\\
        \vspace{1mm}
        \includegraphics[width=0.993\textwidth,height=0.7in]{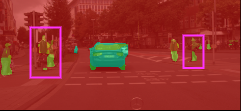}\\  
        \vspace{1mm}
    \end{minipage}%
}%
\subfigure[Ours]{
    \begin{minipage}{0.19\linewidth}
        \centering
        \includegraphics[width=0.993\textwidth,height=0.7in]{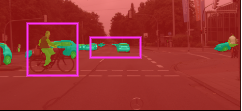}\\
        %\vspace{1mm}
        %\includegraphics[width=0.993\textwidth,height=0.7in]{images/partseg/159_ours_box.png}\\
        \vspace{1mm}
        \includegraphics[width=0.993\textwidth,height=0.7in]{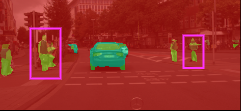}\\ 
        \vspace{1mm}
    \end{minipage}%
}%
\subfigure[GT]{
    \begin{minipage}{0.19\linewidth}
        \centering
        \includegraphics[width=0.993\textwidth,height=0.7in]{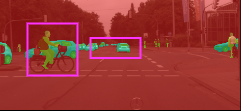}\\
        %\vspace{1mm}
        %\includegraphics[width=0.993\textwidth,height=0.7in]{images/partseg/159_gt_box.png}\\
        \vspace{1mm}
        \includegraphics[width=0.993\textwidth,height=0.7in]{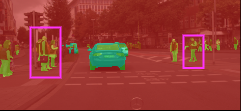}\\ 
        \vspace{1mm}
    \end{minipage}%
}%
\centering
\vspace{-3mm}
\caption{Visualization on Cityscapes~\cite{cityscapes} with part segmentation as the primary task and other two tasks~(semantic segmentation and disparity estimation) as auxiliary.The impressive improvements are marked with a purple box.  }
\vspace{-6mm}
\label{fig:cityscapes_partseg}
\end{figure*}

In Fig.~\ref{fig:cityscapes_seg}-\ref{fig:cityscapes_partseg}, we provide visualize comparison on Cityscapes~\cite{cityscapes}. 
In Fig.~\ref{fig:cityscapes_seg}, we illustrate the results of the semantic segmentation as the primary task on the Cityscapes~\cite{cityscapes} dataset. By utilizing well-trained part segmentation and disparity estimation tasks, our method demonstrates improved segmentation performance on challenging categories such as differentiating between humans (pedestrians and rider) and bicycles, which are prone to confusion.
In Fig.~\ref{fig:cityscapes_disp}, the disparity estimation task is enhanced by the assistance of two segmentation tasks, namely semantic segmentation and part segmentation. This assistance allows for more accurate prediction of the disparity differences between objects and their backgrounds, enabling clear distinction of objects that are challenging to identify in other methods within our predicted disparity maps. As a result, this effectively enhances the overall accuracy of the disparity estimation model.
In Fig.~\ref{fig:cityscapes_partseg}, the part segmentation task, aided by the supplementary information provided by semantic segmentation and disparity estimation tasks, achieves more precise segmentation, thereby reducing confusion.

As shown in Table~\ref{table:dense_prediction}, our method achieves the best performance in both \texttt{Normal} and \texttt{Auxiliary setting}s across all the state-of-the-art baselines, \textit{e}.\textit{g}.  outperform IMTL~\cite{imtl} by about $1.77$ \% for the average relative improvement across all tasks.  
On the contrary, auto-$\lambda$ only obtains $4.37$ \% improvement compared to its single-task counterparts, even under the \texttt{Auxiliary setting}. 
Notably, all multi-task learning methods, except for DWA~\cite{mtan_dwa} and ATW~\cite{UMTNet2024}, have achieved multitask performance surpassing the single-task baseline on cityscapes.
However, in \texttt{Auxiliary setting}, GCS and OL-AUX perform worse than their multi-task counterparts, except for UW, GradNorm, DWA and MGDA. 
It suggests that current \texttt{Auxiliary setting} need more accurate and sufficient relationship between primary tasks and auxiliary tasks. 
Nonetheless, our proposed method enables the impartial training of the auxiliary tasks, which in turn promotes the performance of primary task. These results illustrate that our method reaches the highest performance on all tasks in \textbf{Cityscapes}~\cite{cityscapes} including segmentation and regression problems. 

\subsubsection{Results on PASCAL Context}
\label{sec:pascal}
Further, we conduct our experiments on PASCAL Dataset. The dataset contains both indoor and outdoor senarios, which is a challenge in multi-task learning. We use the split from PASCAL-Context~\cite{pascal_context} which has additional annotations for 21-class semantic segmentation, 7-class human part segmentation and edge prediction. 
Specifically, it contains 10,103 images, which are divided into two parts:4998 for training and 5105 for validation. Following the same data pre-process, we evaluate 5 tasks in PASCAL-Context: semantic segmentation~(\textbf{Seg.}), human part segmentation~(\textbf{H.~Part}), surface normal prediction~(\textbf{Norm.}), saliency estimation~(\textbf{Sal.}) and edge prediction~(\textbf{Edge}). 

We use the same training strategy with previous works~\cite{mtan_dwa, vanden_survey}, which use an Adam optimizer with initial 0.0001 learning rate to train a Resnet-18 with deeplab~\cite{deeplabv3} heads. Note that the single task is only trained for 60 epochs and all multi-task methods are trained for 100 epochs with early-stop mechanism.

As shown in Table~\ref{table:pascal}, our method achieves the best performance on semantic segmentation, human part segmentation, normal estimation and saliency estimation, \eg outperform Auto-$\lambda$~\cite{autolambda} by 0.7\% on semantic segmentation, surpass IMTL~\cite{imtl} by relative 2.1\% improvement on normal estimation, and reach the comparable SOTA performance~(59.5\% and 65.4\% on human part segmentation and saliency estimation respectively). On edge prediction, uncertainty weight~\cite{uw}, DWA~\cite{mtan_dwa} and IMTL-L~\cite{imtl} surpass our method, while our method achieves the best multi-task performance across the five tasks and surpass the 2nd best performance IMTL by 1.2\%. 

\begin{table}[ht]
\caption{Comparison of multi-task learning performance on \textbf{PASCAL-Context}, where the mean and standard deviation over $5$ random seeds for each measurement are reported. The best and $2$nd best performances are marked in \textbf{bold} and \underline{underlined}, respectively. All results are reimplemented.\label{table:pascal}}
\vspace{-2mm}
\centering
\renewcommand\arraystretch{1.2}
\resizebox{\linewidth}{!}{
\begin{tabular}{l cccccc}
\toprule
  \multirow{2}{*}{Method}& Seg. & H.~Parts & Norm. & Sal. & Edge & $ \Delta MTL $  \\
  & mIoU[\%]($\uparrow$) & mIoU[\%]($\uparrow$) & mErr($\downarrow$) & mIoU[\%]($\uparrow$) & odsF($\uparrow$) & [\%]($\uparrow$) \\
\cmidrule(r){1-1} \cmidrule(lr){2-2} \cmidrule(lr){3-3} \cmidrule(lr){4-4} \cmidrule(r){5-5} \cmidrule(r){6-6} \cmidrule(r){7-7}

Single task    & \multicolumn{1}{|c}{66.2~\stdvu{±0.1}}  & 59.4~\stdvu{±0.3} & 13.9~\stdvu{±0.1} & 66.3~\stdvu{±0.1}    & 68.8~\stdvu{±0.2} & - \\

\hline
Uniform               & \multicolumn{1}{|c}{64.0~\stdvu{±0.2}} & 58.5~\stdvu{±0.1} & 15.0~\stdvu{±0.2} & 64.8~\stdvu{±0.2} & 67.9~\stdvu{±0.2} & -3.26\%   \\
UW~\cite{uw}                    & \multicolumn{1}{|c}{65.1~\stdvu{±0.2}} & \textbf{59.5~\stdvu{±0.2}} & 16.0~\stdvu{±0.3} & 65.1~\stdvu{±0.1} & \underline{68.5~\stdvu{±0.3}} & - 3.77\% \\
GradNorm~\cite{gradnorm}              & \multicolumn{1}{|c}{64.8~\stdvu{±0.1}} & 59.0~\stdvu{±0.3} & 14.8~\stdvu{±0.1} & 65.0~\stdvu{±0.1}& 67.8~\stdvu{±0.4} & - 2.65\% \\
DWA~\cite{mtan_dwa}              & \multicolumn{1}{|c}{63.9~\stdvu{±0.1}} & 58.7~\stdvu{±0.2} & 14.7~\stdvu{±0.1} & 65.1~\stdvu{±0.2}& \textbf{69.0~\stdvu{±0.2}} & - 2.36\% \\
MGDA~\cite{mtmo}              & \multicolumn{1}{|c}{65.0~\stdvu{±0.3}} & 58.1~\stdvu{±0.3} & 15.2~\stdvu{±0.4} & 62.8~\stdvu{±0.2}& 62.1~\stdvu{±0.4} & - 5.67\% \\
PCGrad~\cite{pcgrad}              & \multicolumn{1}{|c}{64.9~\stdvu{±0.3}} & 58.8~\stdvu{±0.2} & 15.4~\stdvu{±0.3} & \textbf{65.4~\stdvu{±0.3}}& 68.0~\stdvu{±0.3} & - 3.26\% \\
RLW-Normal~\cite{rlw}              & \multicolumn{1}{|c}{64.3~\stdvu{±0.2}} & 58.4~\stdvu{±0.1} & 15.0~\stdvu{±0.1} & 64.6~\stdvu{±0.1}& 67.8~\stdvu{±0.2} & - 3.30\% \\
IMTL-L~\cite{imtl}              & \multicolumn{1}{|c}{65.0~\stdvu{±0.2}} & 58.6~\stdvu{±0.2} & 15.5~\stdvu{±0.2} & 65.0~\stdvu{±0.1}& 68.2~\stdvu{±0.3} & - 3.50\% \\
IMTL-G~\cite{imtl}              & \multicolumn{1}{|c}{64.6~\stdvu{±0.3}} & 58.9~\stdvu{±0.3} & 14.6~\stdvu{±0.3} & 64.9~\stdvu{±0.2}& 67.9~\stdvu{±0.4} & - 2.34\% \\
IMTL~\cite{imtl}              & \multicolumn{1}{|c}{64.7~\stdvu{±0.2}} & 58.9~\stdvu{±0.2} & \underline{14.5~\stdvu{±0.2}} & 64.9~\stdvu{±0.2} & 68.1~\stdvu{±0.3} & \underline{- 2.11\%} \\
Auto-$\lambda$~(MTL)~\cite{autolambda}    & \multicolumn{1}{|c}{64.9~\stdvu{±0.3}} & 58.8~\stdvu{±0.3} & 14.9~\stdvu{±0.2} & 65.2~\stdvu{±0.3} & 67.5~\stdvu{±0.3} & - 2.74\% \\

Miraliev \etal~\cite{RTM2024}    & \multicolumn{1}{|c}{65.0~\stdvu{±0.2}} & 59.3~\stdvu{±0.2} & 15.8~\stdvu{±0.3} & 65.1~\stdvu{±0.1} & 68.2~\stdvu{±0.3} & - 3.67\% \\

ATW~\cite{UMTNet2024}    & \multicolumn{1}{|c}{64.8~\stdvu{±0.2}} & 59.0~\stdvu{±0.2} & 15.1~\stdvu{±0.2} & 65.0~\stdvu{±0.2} & 68.0~\stdvu{±0.2} & - 2.91\% \\

PUW~\cite{PUM2023}    & \multicolumn{1}{|c}{64.7~\stdvu{±0.3}} & 59.2~\stdvu{±0.3} & 15.0~\stdvu{±0.3} & 64.8~\stdvu{±0.3} & 67.9~\stdvu{±0.4} & - 2.82\% \\
\hline
GCS~\cite{gcs}                   & \multicolumn{1}{|c}{64.6~\stdvu{±0.3}} & 58.5~\stdvu{±0.3} & 14.8~\stdvu{±0.2} & 64.6~\stdvu{±0.3} & 66.3~\stdvu{±0.4} & - 3.32\%  \\
OL-AUX~\cite{ol_aux}                & \multicolumn{1}{|c}{64.2~\stdvu{±0.2}} & 58.3~\stdvu{±0.2} & 14.9~\stdvu{±0.1} & 64.9~\stdvu{±0.3} & 67.7~\stdvu{±0.2} & - 3.16\%  \\
Auto-$\lambda$~\cite{autolambda}        & \multicolumn{1}{|c}{\underline{65.1~\stdvu{±0.3}}} & 59.0~\stdvu{±0.2} & 14.6~\stdvu{±0.2} & 65.3~\stdvu{±0.3} & 67.1~\stdvu{±0.4} & - 2.27\%  \\
{\cellcolor{mypink}\texttt{\textbf{Ours}}}                  & \multicolumn{1}{|c}{{\cellcolor{mypink}\textbf{65.8~\stdvu{±0.3}}}} & {\cellcolor{mypink}\textbf{59.5~\stdvu{±0.2}}} & {\cellcolor{mypink}\textbf{14.2~\stdvu{±0.1}}} &
{\cellcolor{mypink}\textbf{65.4~\stdvu{±0.3}}} &{\cellcolor{mypink}68.1~\stdvu{±0.3}} & {\cellcolor{mypink}\textbf{- 0.91\%}} \\
\bottomrule
\end{tabular}}
\end{table}

\subsubsection{training time}
\label{sec:training_time}
We also report the training time of our method with that of comparative methods in Table~\ref{table:computation_time}. Since our method utilizes both gradients and adaptive weights, it tends to have a longer training time compared to methods like Uncertainty Weight and DWA, which only involve weighting. However, as our method only uses the magnitude of gradients and does not require gradient surgery, it is slightly faster than methods like MGDA, PCGrad, and IMTL-G. Additionally, in comparison to auto-lambda, which employs meta-learning for bi-level optimization, our method has a comparable training duration.

\subsection{Pseudo Tasks Evaluation.}
\label{sec:noisy_exp}

\begin{table}[ht]
\caption{Comparison across existing multi-task methods on the noisy \textbf{NYUv2} benchmark, where two pseudo-tasks that are constructed by large models pre-trained on COCO and ImageNet-21K are added. All results are reimplemented.\label{table:nyu_noise}}
\vspace{-2mm}
\centering
\renewcommand\arraystretch{1.2}
\resizebox{\linewidth}{!}{
\begin{tabular}{l cccc}
\toprule
  \multirow{2}{*}{Method}& Depth        & Segmentation & Normals & $ \Delta MTL $  \\
  & RMSE[m]($\downarrow$) & mIoU[\%]($\uparrow$) & Mean Error($\downarrow$) & [\%]($\uparrow$) \\
\cmidrule(r){1-1} \cmidrule(lr){2-2} \cmidrule(lr){3-3} \cmidrule(lr){4-4} \cmidrule(r){5-5}

\multirow{3}{*}{Single task}    & \multicolumn{1}{|c}{0.5877~\stdvu{±0.0006}}  & -       & -       & - \\
                                & \multicolumn{1}{|c}{-}       & 43.58~\stdvu{±0.05}   & -       & - \\
                                & \multicolumn{1}{|c}{-}       & -       & 19.49~\stdvu{±0.03}   & - \\

\hline
Uniform               & \multicolumn{1}{|c}{0.6105~\stdvu{±0.0008}} & 40.47~\stdvu{±0.02} & 22.13~\stdvu{±0.03} & - 8.17\%   \\
UW~\cite{uw}                    & \multicolumn{1}{|c}{0.6122~\stdvu{±0.0007}} & 41.38~\stdvu{±0.09} & 22.76~\stdvu{±0.03} & - 8.59\% \\
GradNorm~\cite{gradnorm}              & \multicolumn{1}{|c}{0.6076~\stdvu{±0.0010}} & 38.95~\stdvu{±0.13} & 23.99~\stdvu{±0.04} & - 12.37\% \\
IMTL~\cite{imtl}                  & \multicolumn{1}{|c}{0.6241~\stdvu{±0.0006}} & 36.47~\stdvu{±0.10} & 24.01~\stdvu{±0.03} & - 15.23\% \\
\hline
GCS~\cite{gcs}                   & \multicolumn{1}{|c}{0.5834~\stdvu{±0.0017}} & 43.88~\stdvu{±0.14} & 21.75~\stdvu{±0.05} & - 3.39\%  \\
OL-AUX~\cite{ol_aux}                & \multicolumn{1}{|c}{0.5873~\stdvu{±0.0011}} & 43.47~\stdvu{±0.16} & 21.90~\stdvu{±0.05} & - 4.18\%  \\
Auto-$\lambda$~\cite{autolambda}        & \multicolumn{1}{|c}{0.5803~\stdvu{±0.0012}} & 43.86~\stdvu{±0.09} & 21.57~\stdvu{±0.04} & - 2.92\%  \\
{\cellcolor{mypink}\texttt{\textbf{Ours}}}                  & \multicolumn{1}{|c}{{\cellcolor{mypink}\textbf{0.5732~\stdvu{±0.0009}}}} & {\cellcolor{mypink}\textbf{44.82~\stdvu{±0.09}}} & {\cellcolor{mypink}\textbf{19.36~\stdvu{±0.03}}} & {\cellcolor{mypink}\textbf{+ 1.99\%}} \\

\bottomrule
\end{tabular}}
\end{table}

\subsubsection{Settings}
Pseudo tasks evaluation employs the prediction labels of pre-trained large-scale models as the annotations in order to construct auxiliary tasks~(known as pseudo tasks). Pseudo tasks generate more interference on the primary tasks because their annotation quality is lower compared to that of humans.In such setting, it allows for better testing of the robustness of the method we propose.

To achieve this, transferring the knowledge of the large-scale models to standard multi-task datasets by pseudo-labeling provides more available auxiliary tasks. 
Follow~\cite{must}, we are pre-processing various types of pseudo-labels.
For the segmentation task, we use a hard score threshold of $0.5$ to generate segmentation masks where low-score pixels are ignored.
For the classification task, we use soft labels, which contain the probability distribution of all classes, 
because experiments show that their performance appears to be better than that of hard labels. Moreover, we add an additional supervised signal following the token label~\cite{tokenlabel}. This annotation is generated via a pre-trained classification model and is beneficial for knowledge transfer. 

\begin{figure*}[t]
\centering
\subfigure[Image]{
    \begin{minipage}{0.19\linewidth}
        \centering
        \includegraphics[width=0.993\textwidth,height=0.7in]{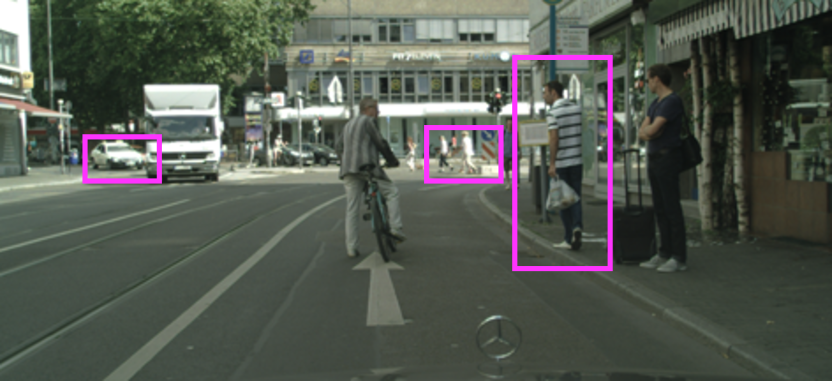}\\
        %\vspace{1mm}
        %\includegraphics[width=0.993\textwidth,height=0.7in]{images/seg_noise/493_origin_box.png}\\
        \vspace{1mm}
        \includegraphics[width=0.993\textwidth,height=0.7in]{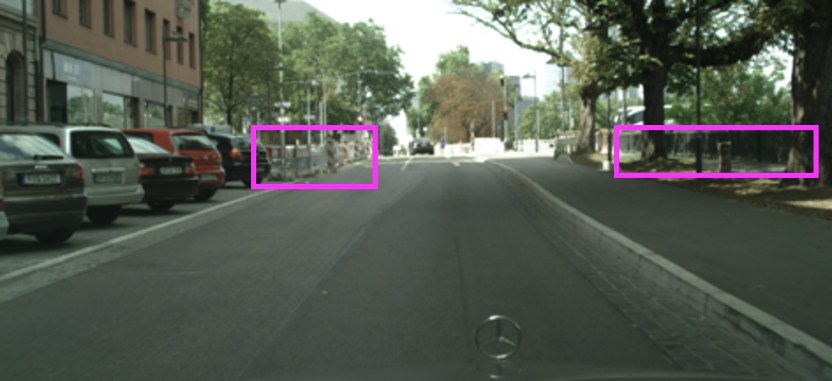}\\ 
        \vspace{1mm}
        % \caption{}
    \end{minipage}%
}%
\subfigure[OL$\text{-} $AUX~\cite{ol_aux}]{
    \begin{minipage}{0.19\linewidth}
        \centering
        \includegraphics[width=0.993\textwidth,height=0.7in]{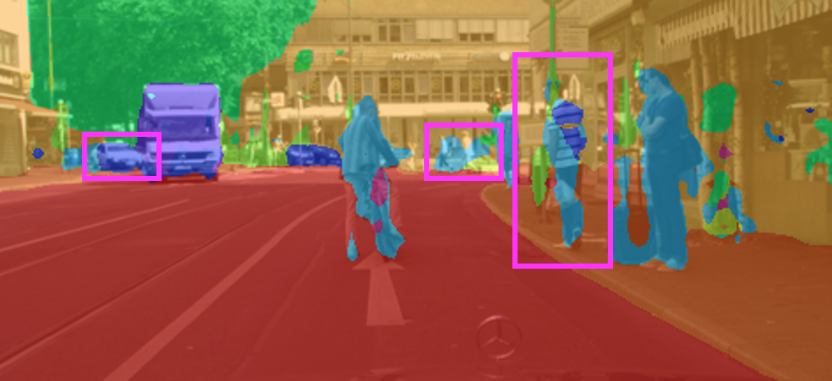}\\
        %\vspace{1mm}
        %\includegraphics[width=0.993\textwidth,height=0.7in]{images/seg_noise/493_OL-AUX_box.png}\\
        \vspace{1mm}
        \includegraphics[width=0.993\textwidth,height=0.7in]{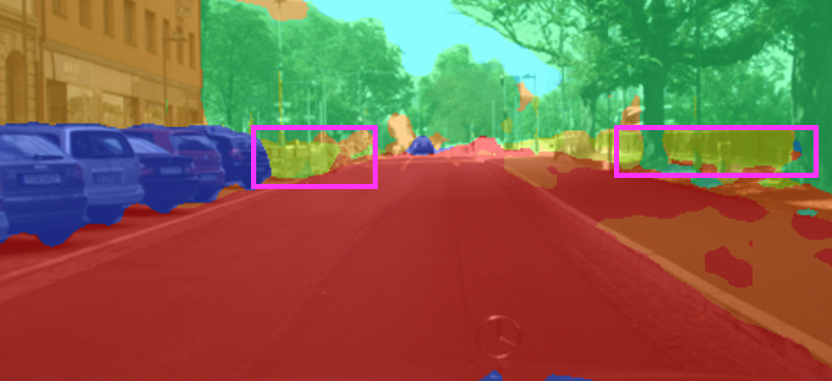}\\    
        \vspace{1mm}
    \end{minipage}%
}%
\subfigure[Auto$\text{-}\lambda $~\cite{autolambda}]{
    \begin{minipage}{0.19\linewidth}
        \centering
        \includegraphics[width=0.993\textwidth,height=0.7in]{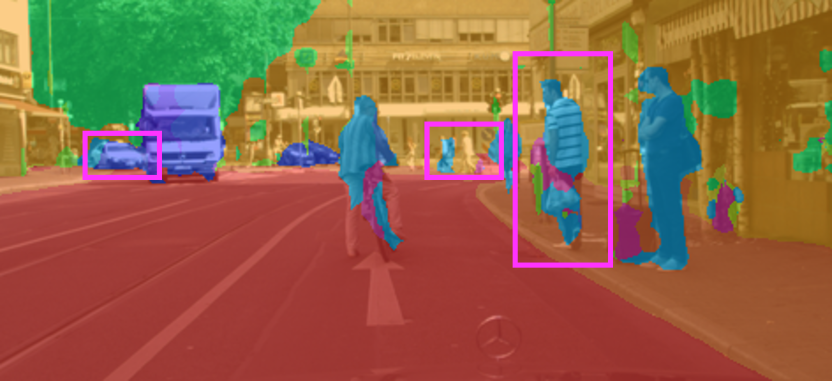}\\
        %\vspace{1mm}
        %\includegraphics[width=0.993\textwidth,height=0.7in]{images/seg_noise/493_auto-lambda_box.png}\\
        \vspace{1mm}
        \includegraphics[width=0.993\textwidth,height=0.7in]{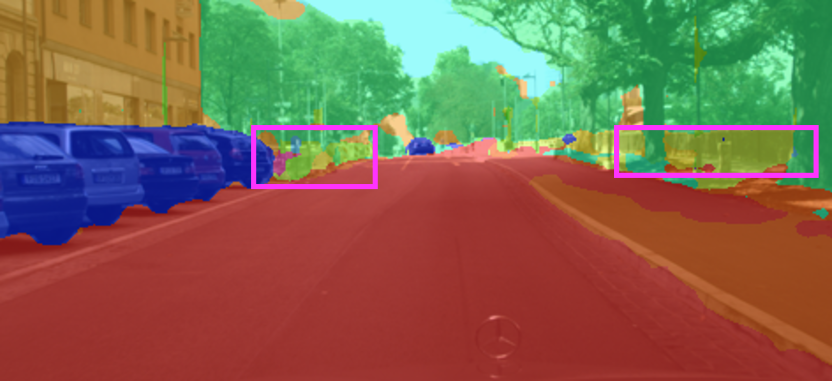}\\
        \vspace{1mm}
    \end{minipage}%
}%
\subfigure[Ours]{
    \begin{minipage}{0.19\linewidth}
        \centering
        \includegraphics[width=0.993\textwidth,height=0.7in]{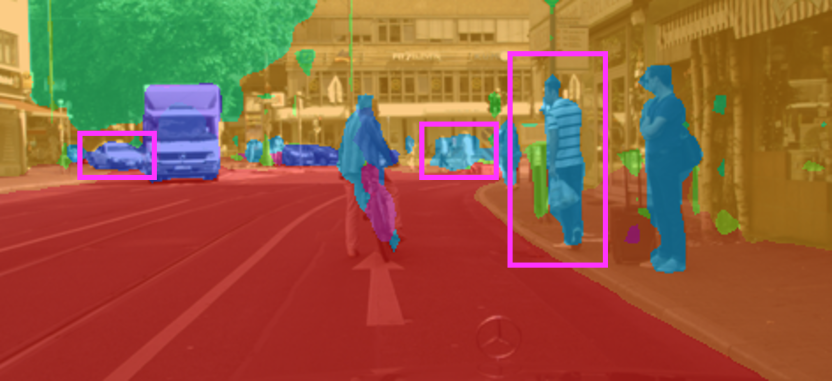}\\
        %\vspace{1mm}
        %\includegraphics[width=0.993\textwidth,height=0.7in]{images/seg_noise/493_ours_box.png}\\
        \vspace{1mm}
        \includegraphics[width=0.993\textwidth,height=0.7in]{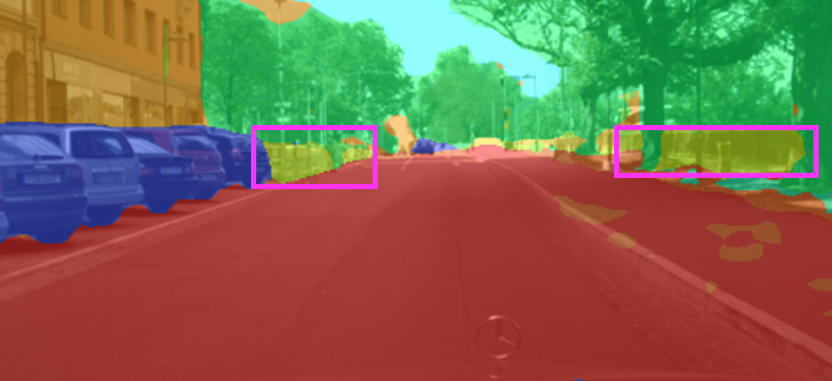}\\
        \vspace{1mm}
    \end{minipage}%
}%
\subfigure[GT]{
    \begin{minipage}{0.19\linewidth}
        \centering
        \includegraphics[width=0.993\textwidth,height=0.7in]{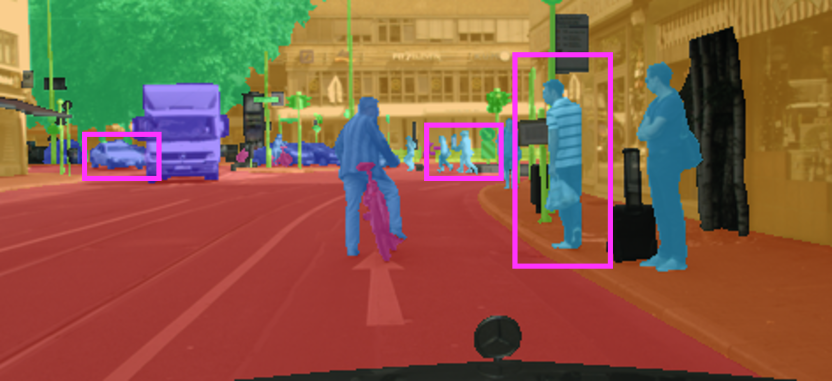}\\
        %\vspace{1mm}
        %\includegraphics[width=0.993\textwidth,height=0.7in]{images/seg_noise/493_gt_box.png}\\
        \vspace{1mm}
        \includegraphics[width=0.993\textwidth,height=0.7in]{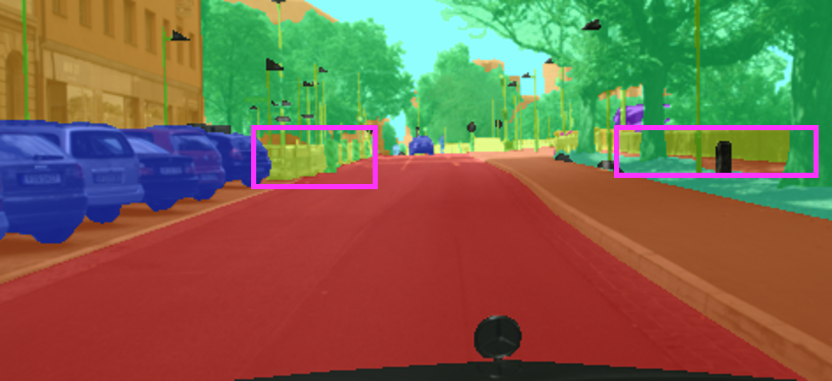}\\
        \vspace{1mm}
    \end{minipage}%
}%
\centering
\vspace{-3mm}
\caption{Visualization on Cityscapes~\cite{cityscapes} with semantic segmentation as the primary task and other two tasks~(disparity estimation and part segmentation) and additional pseudo tasks as auxiliary.The impressive improvements are marked with a purple box.   }
\label{fig:cityscapes_noisy_seg}
\vspace{-6mm}
\end{figure*}

\begin{figure*}[t]
\centering
\subfigure[Image]{
    \begin{minipage}{0.19\linewidth}
        \centering
        \includegraphics[width=0.993\textwidth,height=0.7in]{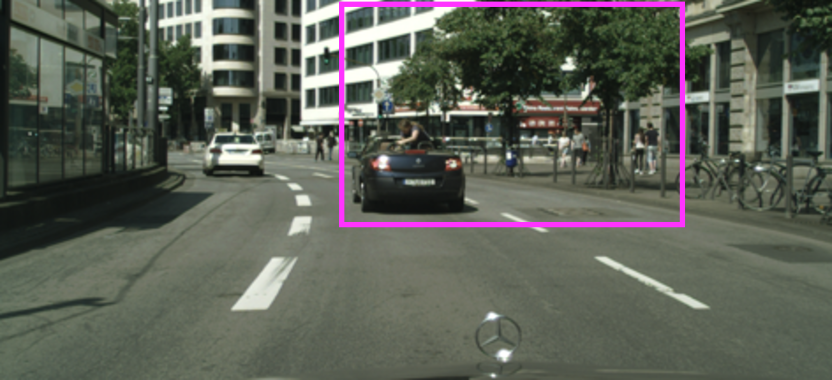}\\
        %\vspace{1mm}
        %\includegraphics[width=0.993\textwidth,height=0.7in]{images/disp_noise/404_origin_box.png}\\
        \vspace{1mm}
        \includegraphics[width=0.993\textwidth,height=0.7in]{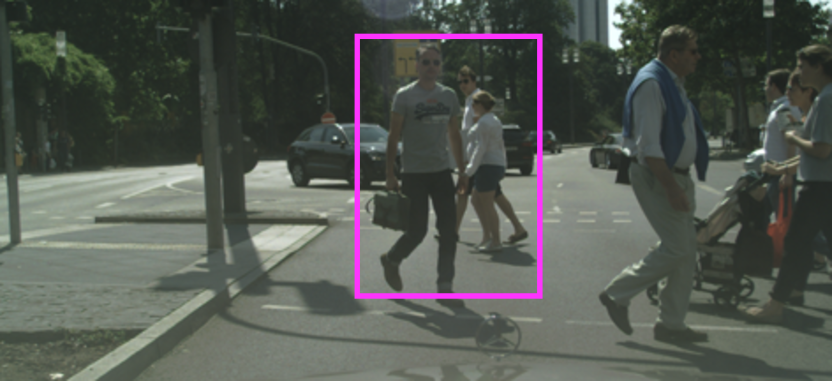}\\     
        \vspace{1mm}
        % \caption{}
    \end{minipage}%
}%
\subfigure[OL$\text{-} $AUX~\cite{ol_aux}]{
    \begin{minipage}{0.19\linewidth}
        \centering
        \includegraphics[width=0.993\textwidth,height=0.7in]{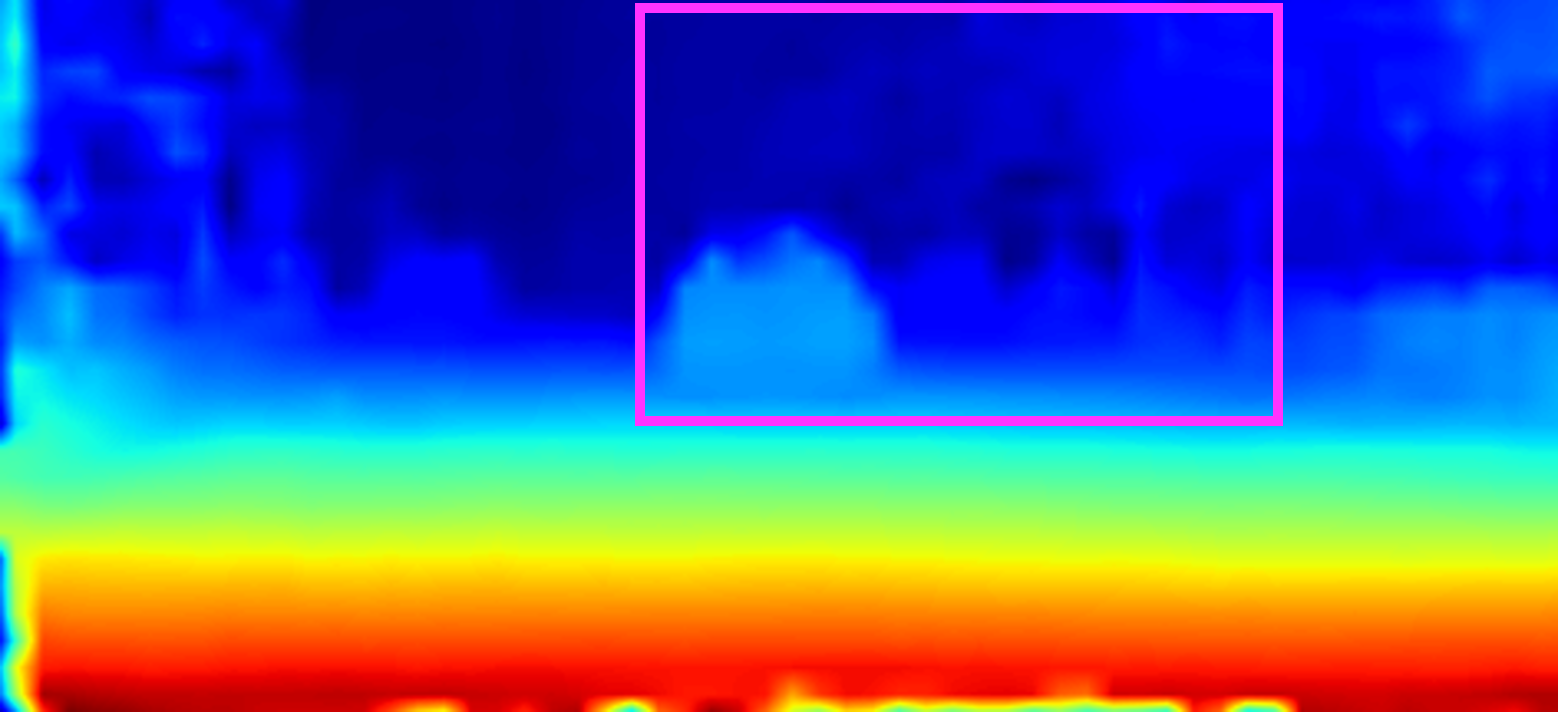}\\
        %\vspace{1mm}
        %\includegraphics[width=0.993\textwidth,height=0.7in]{images/disp_noise/404_OL-AUX_box.png}\\
        \vspace{1mm}
        \includegraphics[width=0.993\textwidth,height=0.7in]{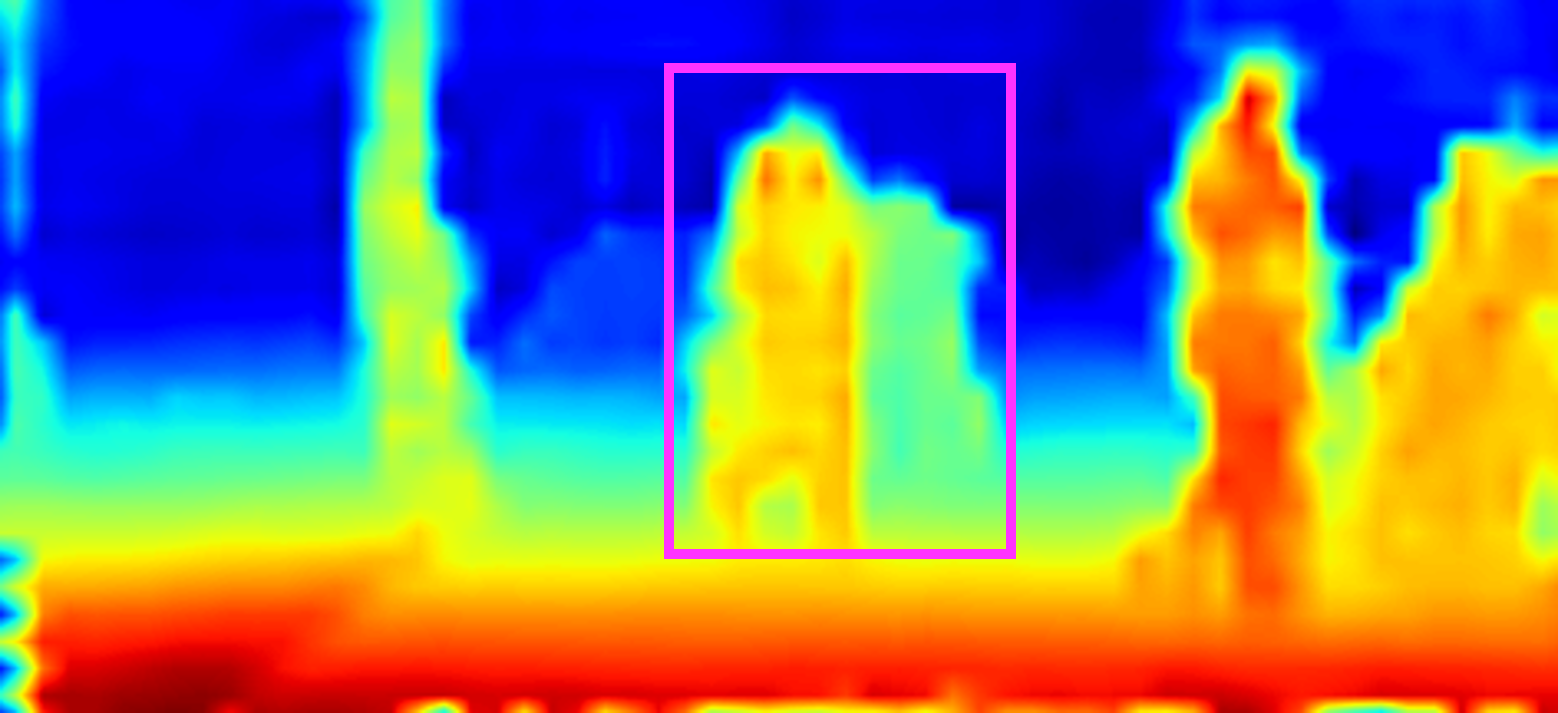}\\    
        \vspace{1mm}
    \end{minipage}%
}%
\subfigure[Auto$\text{-}\lambda $~\cite{autolambda}]{
    \begin{minipage}{0.19\linewidth}
        \centering
        \includegraphics[width=0.993\textwidth,height=0.7in]{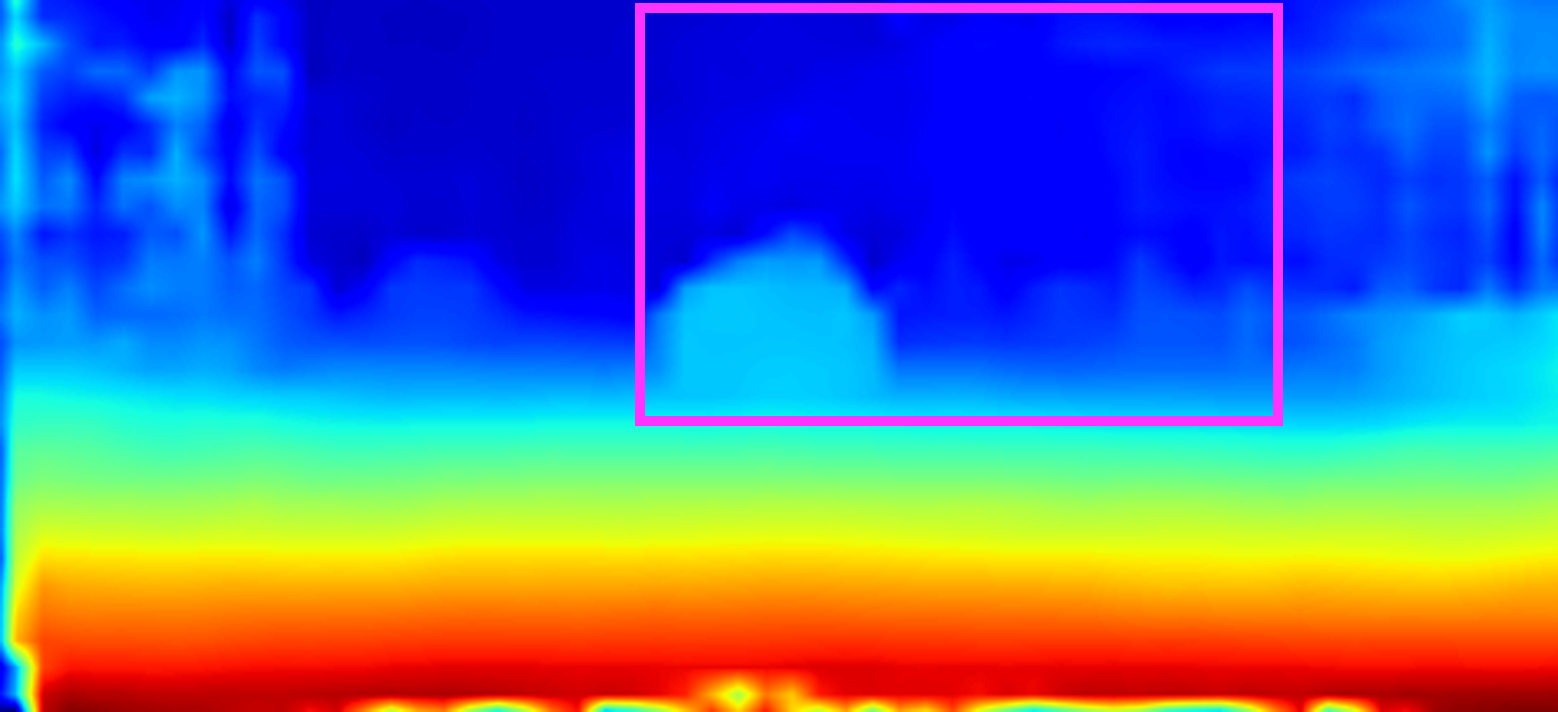}\\
        %\vspace{1mm}
        %\includegraphics[width=0.993\textwidth,height=0.7in]{images/disp_noise/404_auto-lambda_box.png}\\
        \vspace{1mm}
        \includegraphics[width=0.993\textwidth,height=0.7in]{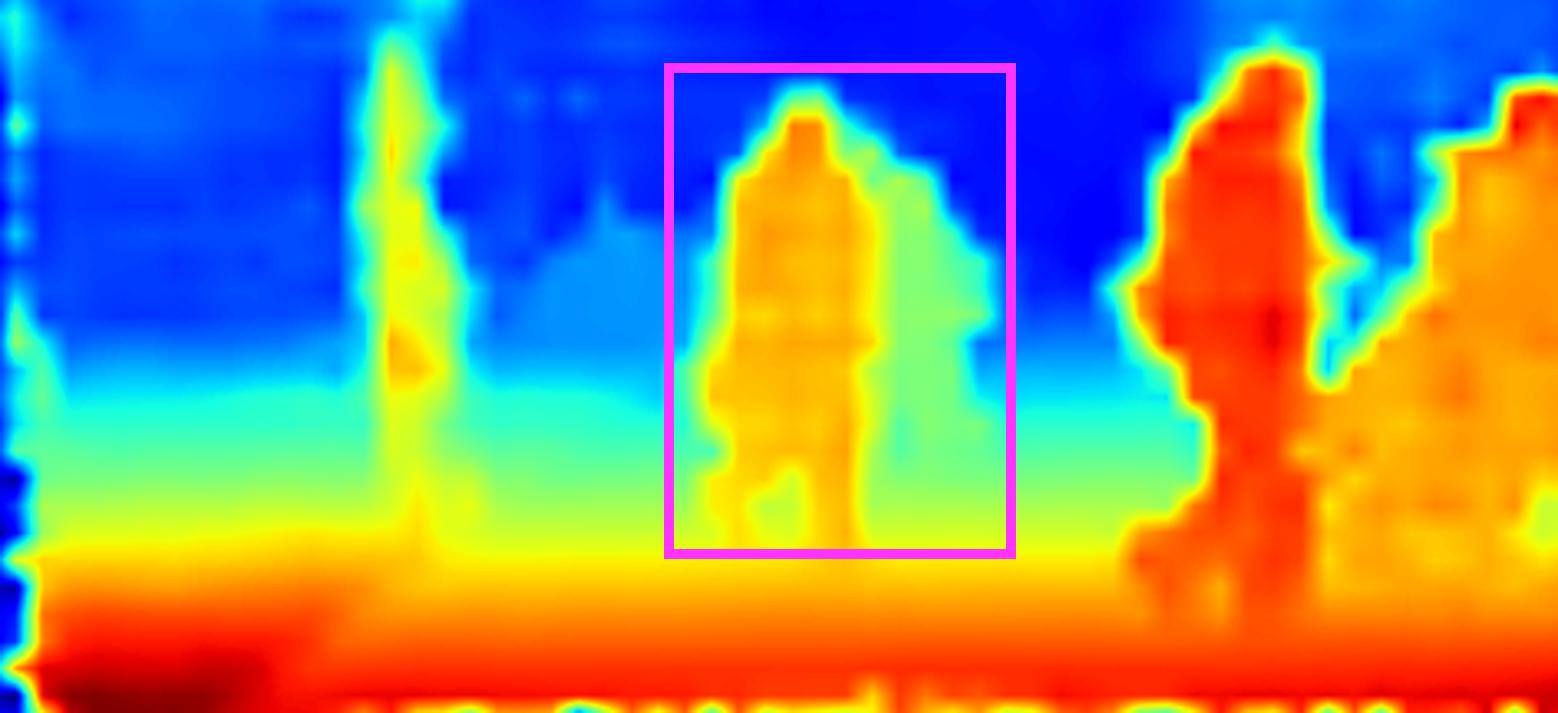}\\  
        \vspace{1mm}
    \end{minipage}%
}%
\subfigure[Ours]{
    \begin{minipage}{0.19\linewidth}
        \centering
        \includegraphics[width=0.993\textwidth,height=0.7in]{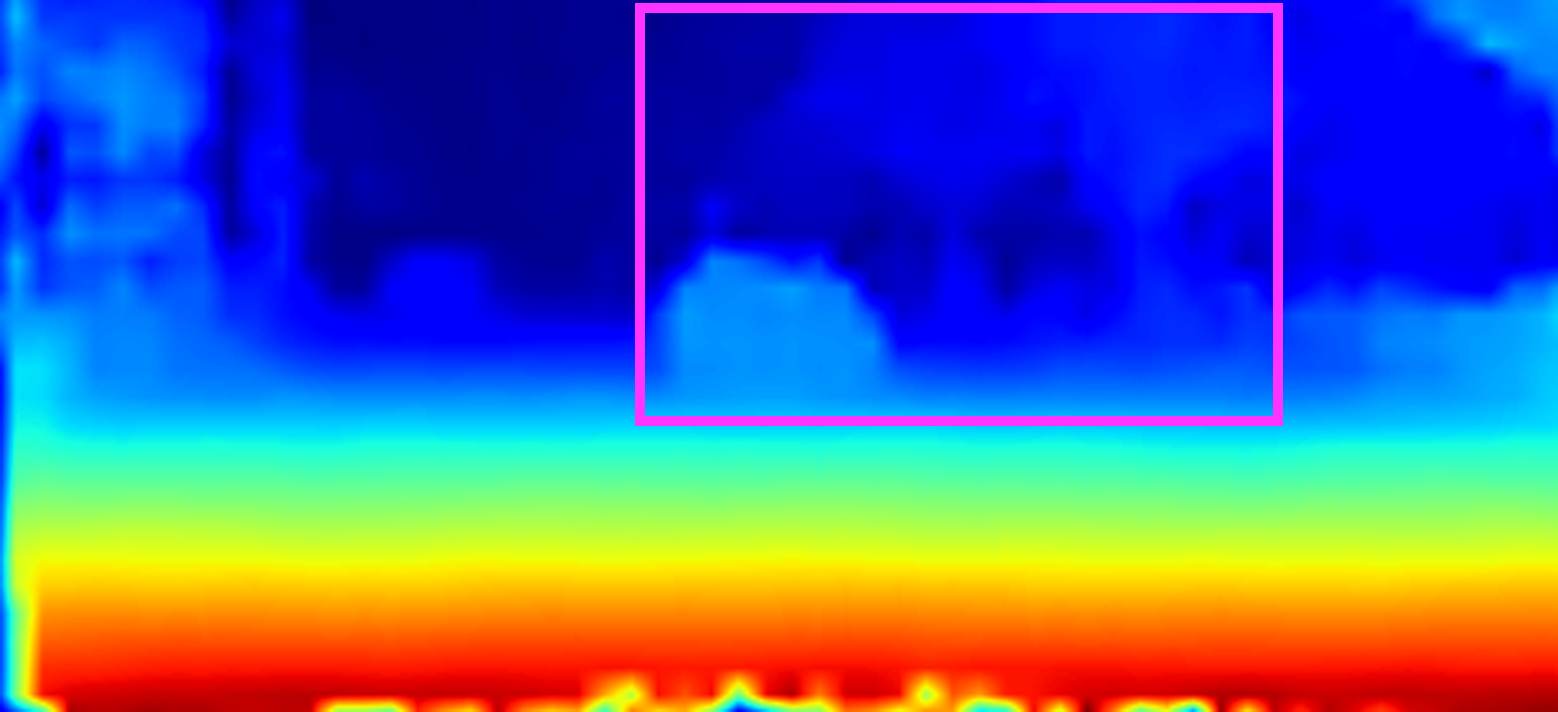}\\
        %\vspace{1mm}
        %\includegraphics[width=0.993\textwidth,height=0.7in]{images/disp_noise/404_ours_box.png}\\
        \vspace{1mm}
        \includegraphics[width=0.993\textwidth,height=0.7in]{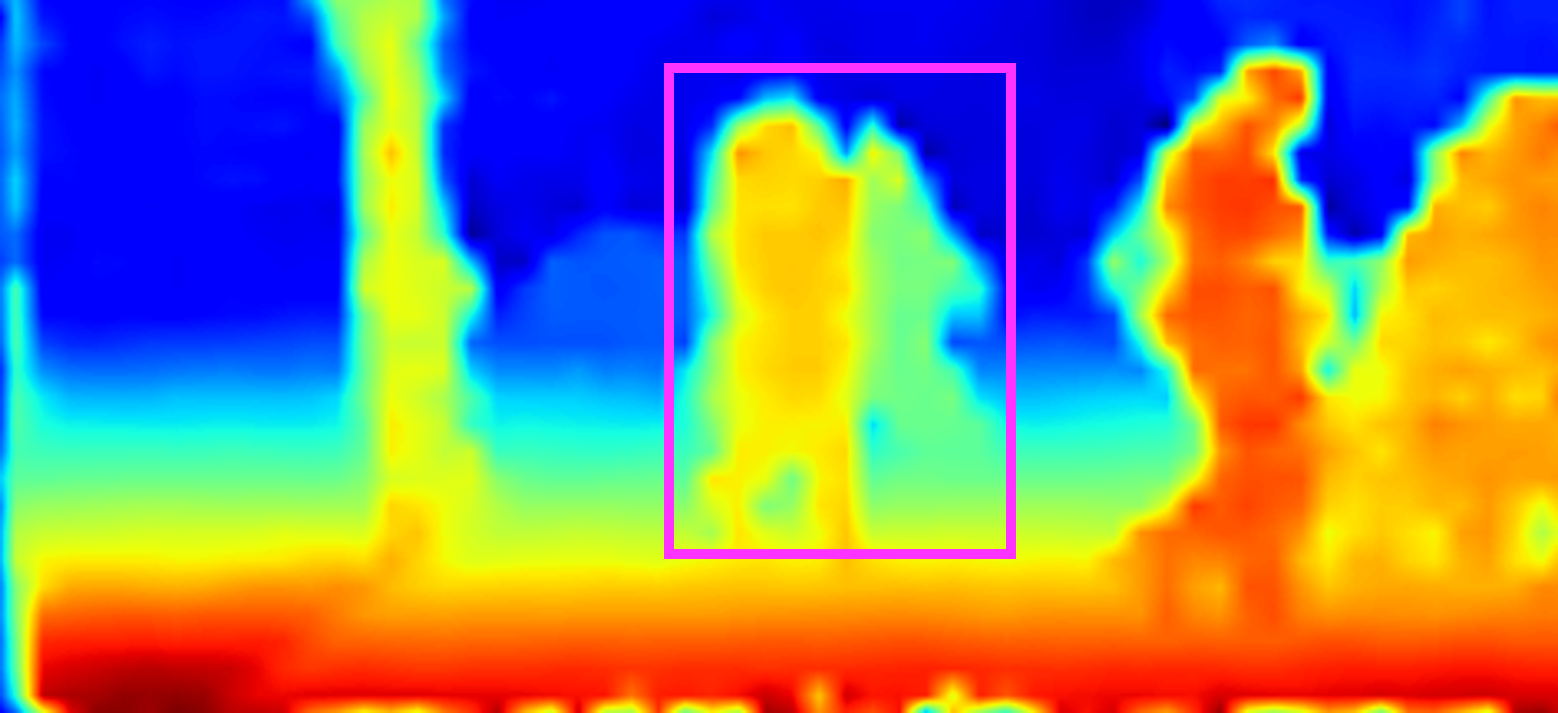}\\  
        \vspace{1mm}
    \end{minipage}%
}%
\subfigure[GT]{
    \begin{minipage}{0.19\linewidth}
        \centering
        \includegraphics[width=0.993\textwidth,height=0.7in]{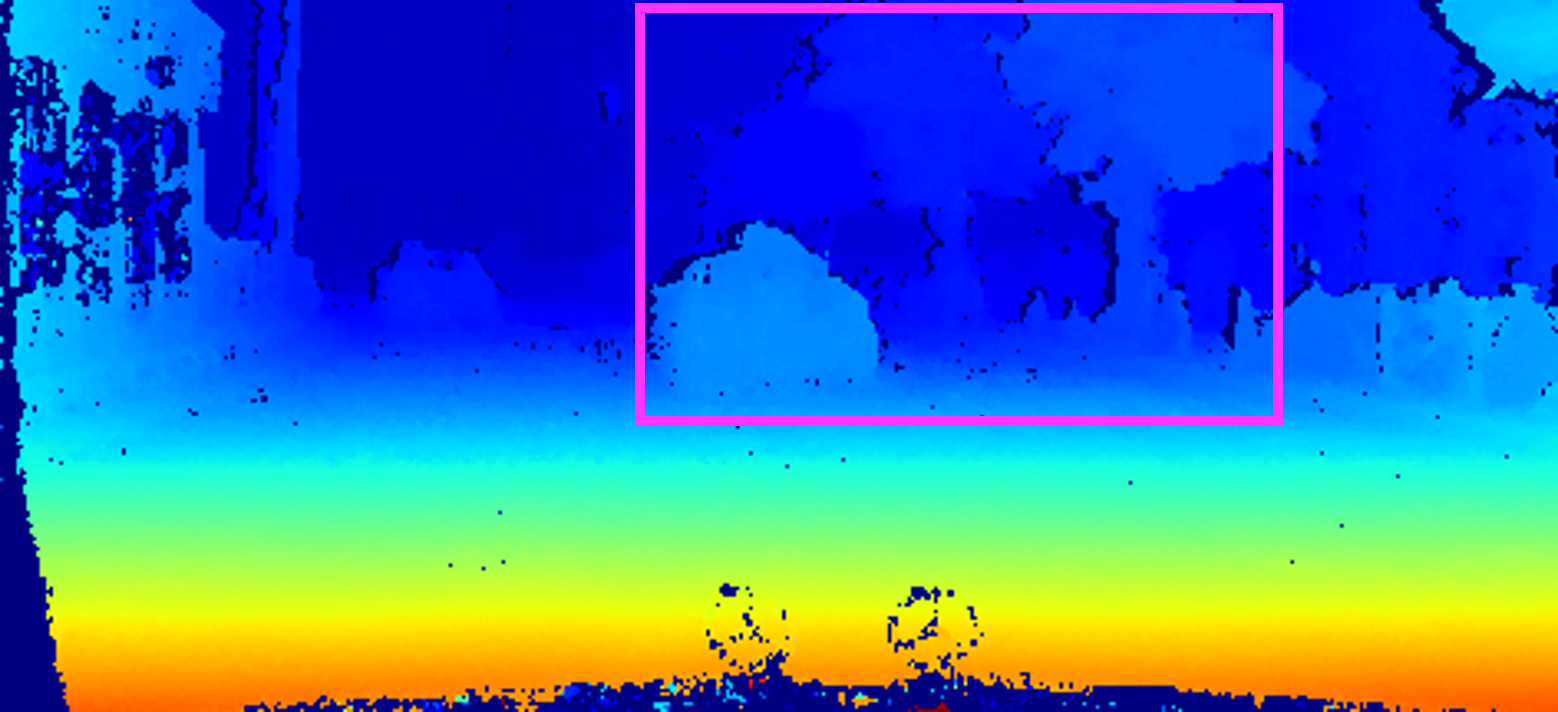}\\
        %\vspace{1mm}
        %\includegraphics[width=0.993\textwidth,height=0.7in]{images/disp_noise/404_gt_box.png}\\
        \vspace{1mm}
        \includegraphics[width=0.993\textwidth,height=0.7in]{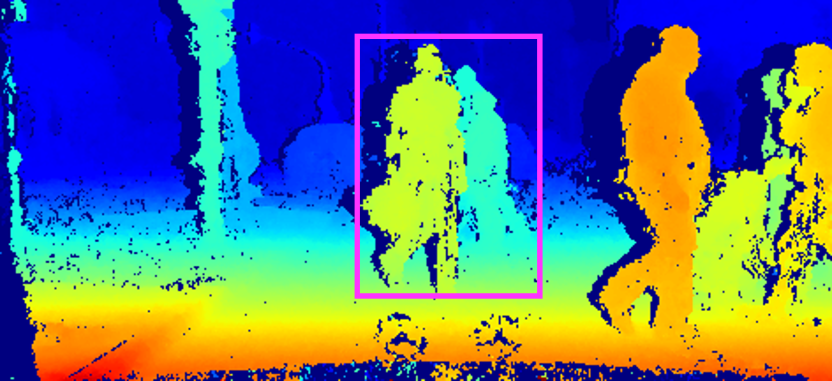}\\  
        \vspace{1mm}
    \end{minipage}%
}%
\centering
\vspace{-3mm}
\caption{Visualization on Cityscapes~\cite{cityscapes} with disparity estimation as the primary task and other two tasks~(semantic segmentation and part segmentation) and additional pseudo tasks as auxiliary.The impressive improvements are marked with a purple box.  }
\vspace{-6mm}
\label{fig:cityscapes_noisy_disp}
\end{figure*}

\begin{table}[ht]
\caption{Comparison of the performance under the \texttt{Auxiliary setting} on the noisy \textbf{Cityscapes} benchmarks, where the results of the original three tasks influenced by the additional two pseudo tasks are reported. All results are reimplemented. \label{table:cityscapes}}
\vspace{-2mm}
\centering
\renewcommand\arraystretch{1.2}
\resizebox{\linewidth}{!}{
\begin{tabular}{l cccc}
\toprule
  \multirow{2}{*}{Method}& Disparity & Part Seg. & Semantic Seg. & $ \Delta MTL $  \\
  & aErr[m]($\downarrow$) & mIoU[\%]($\uparrow$) & mIoU[\%]($\uparrow$) & [\%]($\uparrow$) \\
\cmidrule(r){1-1} \cmidrule(lr){2-2} \cmidrule(lr){3-3} \cmidrule(lr){4-4} \cmidrule(r){5-5}

Single task    & \multicolumn{1}{|c}{0.8403~\stdvu{±0.0009}}  & 52.71~\stdvu{±0.06}       & 56.22~\stdvu{±0.03}       & + 0.00\% \\

\hline
GCS~\cite{gcs}                     & \multicolumn{1}{|c}{0.7999~\stdvu{±0.0012}} & 52.30~\stdvu{±0.06} & 54.71~\stdvu{±0.05} & + 0.45\%  \\
OL-AUX~\cite{ol_aux}                  & \multicolumn{1}{|c}{0.8012~\stdvu{±0.0012}} & 51.55~\stdvu{±0.06} & 55.47~\stdvu{±0.05} & + 0.37\%  \\
Auto-$\lambda$~\cite{autolambda}          & \multicolumn{1}{|c}{0.7944~\stdvu{±0.0012}} & 52.99~\stdvu{±0.06} & 56.85~\stdvu{±0.05} & + 2.37\%  \\
{\cellcolor{mypink}\texttt{\textbf{Ours}}}                    & \multicolumn{1}{|c}{{\cellcolor{mypink}\textbf{0.7588}~\stdvu{±0.0007}}} & {\cellcolor{mypink}\textbf{59.10}~\stdvu{±0.04}} & {\cellcolor{mypink}\textbf{58.97}~\stdvu{±0.04}} & {\cellcolor{mypink}\textbf{+ 8.90\%}} \\
\bottomrule
\end{tabular}}
\end{table}

\begin{figure*}[t]
\centering
\subfigure[Image]{
    \begin{minipage}{0.19\linewidth}
        \centering
        \includegraphics[width=0.993\textwidth,height=0.7in]{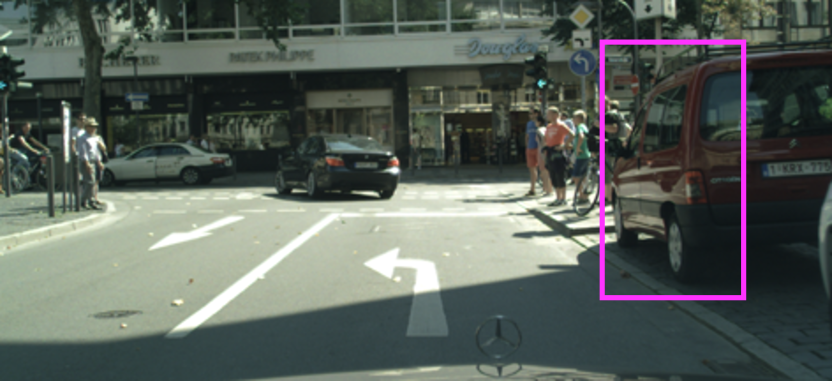}\\
        %\vspace{1mm}
        %\includegraphics[width=0.993\textwidth,height=0.7in]{images/partseg_noise/445_origin_box.png}\\
        \vspace{1mm}
        \includegraphics[width=0.993\textwidth,height=0.7in]{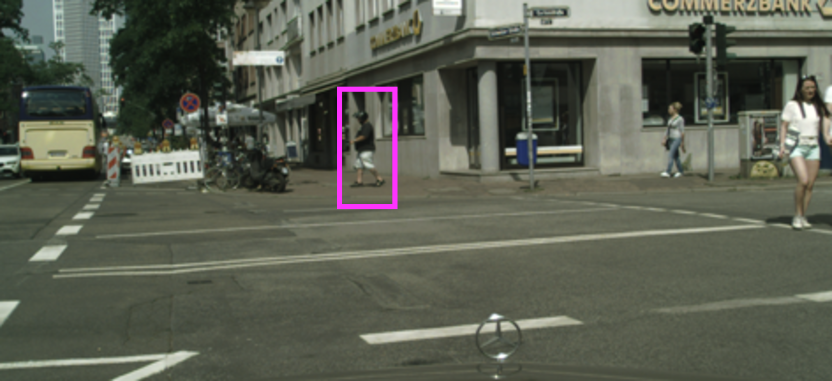}\\   
        \vspace{1mm}
        % \caption{}
    \end{minipage}%
}%
\subfigure[OL$\text{-} $AUX~\cite{ol_aux}]{
    \begin{minipage}{0.19\linewidth}
        \centering
        \includegraphics[width=0.993\textwidth,height=0.7in]{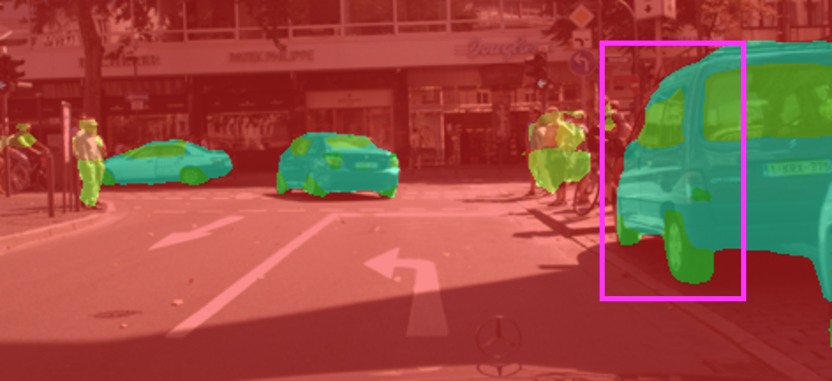}\\
        %\vspace{1mm}
        %\includegraphics[width=0.993\textwidth,height=0.7in]{images/partseg_noise/445_OL-AUX_box.png}\\
        \vspace{1mm}
        \includegraphics[width=0.993\textwidth,height=0.7in]{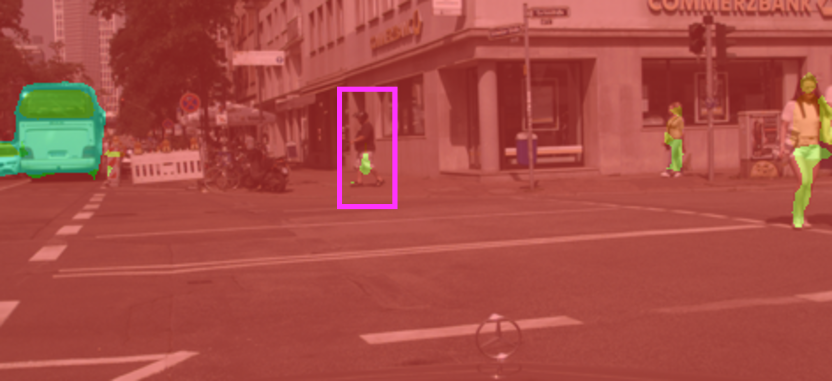}\\    
        \vspace{1mm}
    \end{minipage}%
}%
\subfigure[Auto$\text{-}\lambda $~\cite{autolambda}]{
    \begin{minipage}{0.19\linewidth}
        \centering
        \includegraphics[width=0.993\textwidth,height=0.7in]{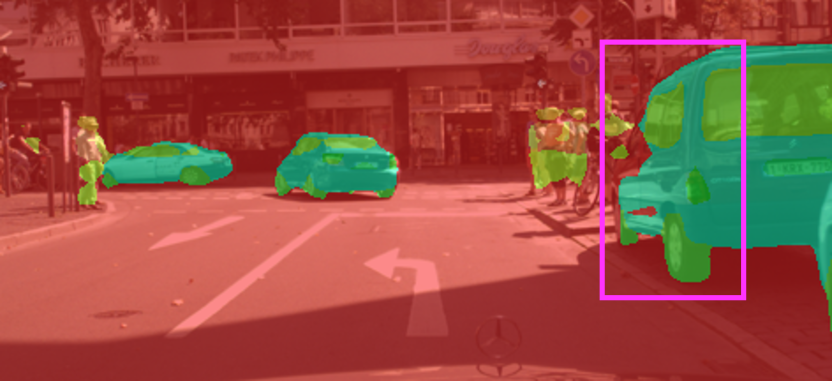}\\
        %\vspace{1mm}
        %\includegraphics[width=0.993\textwidth,height=0.7in]{images/partseg_noise/445_auto-lambda_box.png}\\
        \vspace{1mm}
        \includegraphics[width=0.993\textwidth,height=0.7in]{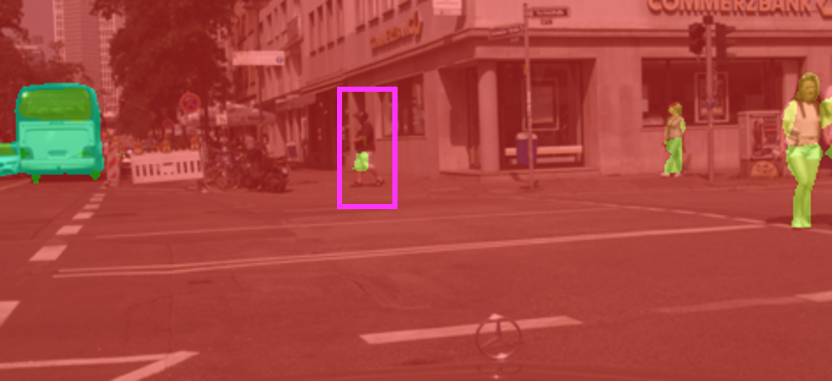}\\  
        \vspace{1mm}
    \end{minipage}%
}%
\subfigure[Ours]{
    \begin{minipage}{0.19\linewidth}
        \centering
        \includegraphics[width=0.993\textwidth,height=0.7in]{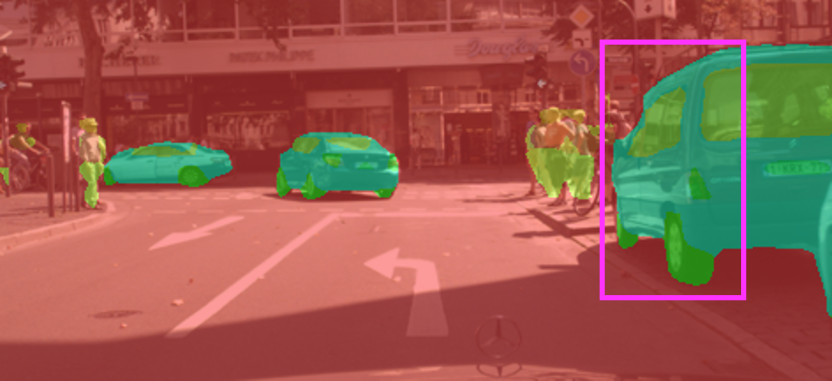}\\
        %\vspace{1mm}
        %\includegraphics[width=0.993\textwidth,height=0.7in]{images/partseg_noise/445_ours_box.png}\\
        \vspace{1mm}
        \includegraphics[width=0.993\textwidth,height=0.7in]{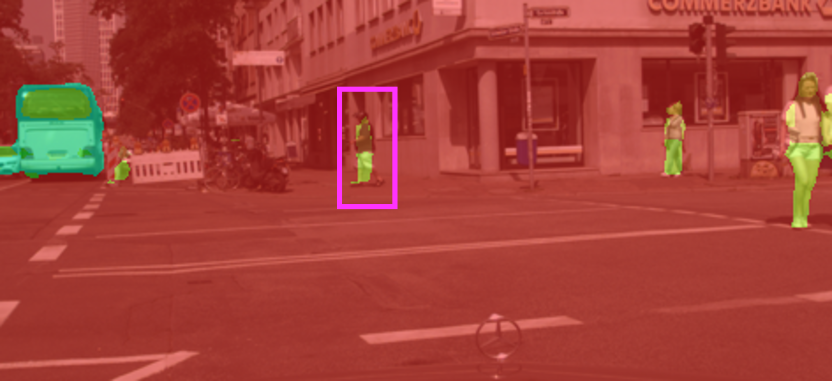}\\  
        \vspace{1mm}
    \end{minipage}%
}%
\subfigure[GT]{
    \begin{minipage}{0.19\linewidth}
        \centering
        \includegraphics[width=0.993\textwidth,height=0.7in]{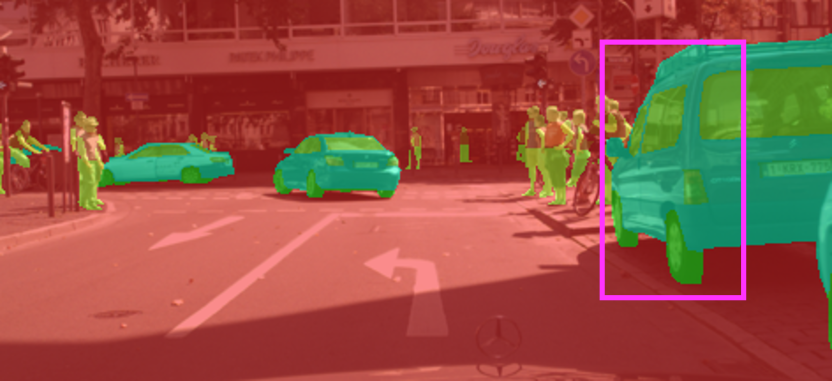}\\
        %\vspace{1mm}
        %\includegraphics[width=0.993\textwidth,height=0.7in]{images/partseg_noise/445_gt_box.png}\\
        \vspace{1mm}
        \includegraphics[width=0.993\textwidth,height=0.7in]{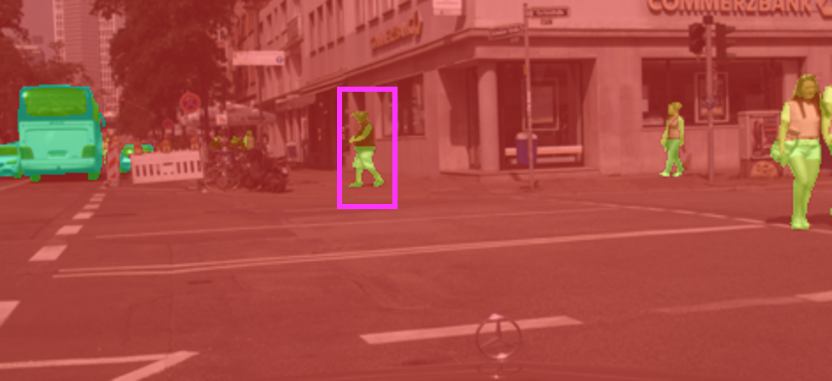}\\ 
        \vspace{1mm}
    \end{minipage}%
}%
\centering
\vspace{-3mm}
\caption{Visualization on Cityscapes~\cite{cityscapes} with part segmentation as the primary task and other two tasks~(semantic segmentation and disparity estimation) and additional pseudo tasks as auxiliary.The impressive improvements are marked with a purple box.  }
\vspace{-6mm}
\label{fig:cityscapes_noisy_part}
\end{figure*}

For all datasets, including \textbf{NYUv2} and \textbf{Cityscapes}, we add two large-scale pre-trained models to construct pseudo auxiliary tasks: $80$-class semantic segmentation on COCO and $1,000$-class classification on ImageNet~\cite{imagenet}. Following standard multi-task learning~\cite{compositelearning}, we do not augment images additionally, since the inference of augmentation in supervised multi-task learning is not fully explored in this work. We leave it for future work. 

\subsubsection{Results on NYUv2}
We report the performance of all the state-of-the-art and multi-task learning baselines on \textbf{NYUv2} with two pseudo auxiliary tasks in Table~\ref{table:nyu_noise}. 
For multi-task learning baselines that weight all tasks equally (noted as “uniform"), performance drops obviously due to the noisy labels of new auxiliary tasks. 
Those methods under \texttt{Normal setting} show interesting trends, \ie,  uncertainty weights improve performance compared to the uniform baseline, while other methods such as IMTL~\cite{imtl} and GradNorm~\cite{gradnorm} perform poorly and are even worse than the baseline. 
Additionally, those methods under \texttt{Auxiliary setting} show good resistance to the noisy tasks, \textit{e}.\textit{g}.  they always set the weight of noisy tasks to $0$ after a few steps and successfully avoid interruption by these auxiliary tasks.
Especially, the auto-$\lambda$ can still improve performance by $0.8$\%. This result indicates that the performance of multi-task learning can be promoted via additional auxiliary tasks and a good task weighting method. 

We presented the results without using token labels in Table~\ref{table:nyu_noise}. Although our approach which rejects noisy tasks do not result in large performance degradation, it also did not provide any additional performance improvement~(in terms of $\Delta MTL$, with token labels -0.45\% v.s. -0.44\% without auxiliary tasks.)

Our proposed method is different from former methods. Concretely, we explicitly build a relationship between the primary and auxiliary tasks via both uncertainty information and gradient norms, rather than setting the weights to $0$. The results show that the performance can be improved up to the baseline without noise and also promoted against the corresponding version without the additional pseudo tasks by $2$\%. 

\subsubsection{Results on Cityscapes}
Similarly, we also evaluate our method on the \textbf{Cityscapes}. 
As in \textbf{NYUv2}, we observe that the baseline methods in \texttt{Auxiliary setting}s are interferenced by the noisy pseudo tasks and the accuracy drops on all tasks. 
For Auto-$\lambda$, different from the results in \textbf{NYUv2}, it lost $2$\% performance according to the original \textbf{Cityscapes} experiment reported in Table~\ref{table:cityscapes}. 
Nonetheless, our method still gets promotion on part segmentation and depth regression and reaches comparable performance on semantic segmentation.
The relative multi-task learning improvement achieves $0.7$\%. 

As in sec.~\ref{sec:noisy_exp}, we report results without token labels in Table~\ref{table:cityscapes}. The same trend is consistent as in NYUv2: the performance with additional pseudo auxiliary tasks do not promote but drop in some tasks~(\textbf{0.0025} in disparity and \textbf{0.45} in semantic segmentation.) 

Although the proposed auxiliary tasks only promote the primary tasks by a smaller margin compared to their \textbf{NYUv2} counterpart, the experiment results still prove that our method can build a relationship between the primary and auxiliary tasks, leading to knowledge transfer from the noisy pseudo tasks which is the main target of multi-task learning.

In Fig.~\ref{fig:cityscapes_noisy_seg}-\ref{fig:cityscapes_noisy_part}, the inclusion of the 1000-class classification task on Imagenet (with token labels) and the 80-class segmentation task on COCO enhances the original Cityscapes semantic segmentation and part segmentation tasks, enabling more comprehensive segmentation. When used as the primary task, the disparity estimation demonstrates improved performance in capturing fine-grained details of distant backgrounds. This improvement might be attributed to the additional semantic information provided by the auxiliary tasks.

\subsubsection{Results on PASCAL Context}
\label{sec:pascal_noise}
We also evaluate the noise setting on PASCAL-Context~\cite{pascal_context}. The results are reported on Table~\ref{table:pascal_noise}. 
Different from NYUv2 and Cityscapes, the pseudo tasks generalize well on PASCAL dataset, resulting in smaller performance drop on all methods. Specifically, uncertainty weight~\cite{uw} and all auxiliary methods shows a small performance drop under 0.3\%. Our proposed method achieves comparable performance with the noisy tasks and improve the multi-task performance by about 0.3\% against its normal counterpart. 

\begin{table}[ht]
\caption{Comparison of multi-task learning performance on the noisy \textbf{PASCAL-Context}, where the mean and standard deviation over $5$ random seeds for each measurement are reported. The best and $2$nd best performances are marked in \textbf{bold} and \underline{underlined}, respectively. All results are reimplemented. \label{table:pascal_noise}}
\vspace{-2mm}
\centering
\renewcommand\arraystretch{1.2}
\resizebox{\linewidth}{!}{
\begin{tabular}{l cccccc}
\toprule
  \multirow{2}{*}{Method}& Seg. & H.~Parts & Norm. & Sal. & Edge & $ \Delta MTL $  \\
  & mIoU[\%]($\uparrow$) & mIoU[\%]($\uparrow$) & mErr($\downarrow$) & mIoU[\%]($\uparrow$) & odsF($\uparrow$) & [\%]($\uparrow$) \\
\cmidrule(r){1-1} \cmidrule(lr){2-2} \cmidrule(lr){3-3} \cmidrule(lr){4-4} \cmidrule(r){5-5} \cmidrule(r){6-6} \cmidrule(r){7-7}

Single task    & \multicolumn{1}{|c}{66.2~\stdvu{±0.1}}  & 59.4~\stdvu{±0.3} & 13.9~\stdvu{±0.1} & 66.3~\stdvu{±0.1}    & 68.8~\stdvu{±0.2} & - \\

\hline
Uniform               & \multicolumn{1}{|c}{63.8~\stdvu{±0.4}} & 58.0~\stdvu{±0.3} & 15.4~\stdvu{±0.4} & 65.0~\stdvu{±0.3} & 67.5~\stdvu{±0.4} & -4.12\%   \\
UW~\cite{uw}                    & \multicolumn{1}{|c}{64.8~\stdvu{±0.3}} & \underline{59.0~\stdvu{±0.3}} & 15.8~\stdvu{±0.3} & 64.8~\stdvu{±0.2} & \textbf{68.0\stdvu{±0.4}} & - 3.98\% \\
GradNorm~\cite{gradnorm}              & \multicolumn{1}{|c}{64.4~\stdvu{±0.2}} & 58.6~\stdvu{±0.4} & 15.2~\stdvu{±0.2} & 64.7~\stdvu{±0.3}& 66.9~\stdvu{±0.4} & - 3.72\% \\
IMTL~\cite{imtl}              & \multicolumn{1}{|c}{64.0~\stdvu{±0.4}} & 58.5~\stdvu{±0.3} & 14.9~\stdvu{±0.3} & 65.0~\stdvu{±0.4} & 67.5~\stdvu{±0.3} & - 3.18\% \\
\hline
GCS~\cite{gcs}                   & \multicolumn{1}{|c}{64.7~\stdvu{±0.3}} & 57.9~\stdvu{±0.4} & 14.8~\stdvu{±0.2} & 64.6~\stdvu{±0.3} & 66.0~\stdvu{±0.4} & - 3.58\%  \\
OL-AUX~\cite{ol_aux}                & \multicolumn{1}{|c}{64.1~\stdvu{±0.2}} & 58.0~\stdvu{±0.2} & 15.0~\stdvu{±0.1} & 64.7~\stdvu{±0.3} & 67.4~\stdvu{±0.2} & - 3.58\%  \\
Auto-$\lambda$~\cite{autolambda}        & \multicolumn{1}{|c}{\underline{65.4~\stdvu{±0.4}}} & 58.9~\stdvu{±0.3} & \underline{14.6~\stdvu{±0.3}} & \underline{65.4~\stdvu{±0.5}} & 66.9~\stdvu{±0.4} & \underline{- 2.24\%}  \\
{\cellcolor{mypink}\texttt{\textbf{Ours}}}                  & \multicolumn{1}{|c}{{\cellcolor{mypink}\textbf{66.4~\stdvu{±0.3}}}} & {\cellcolor{mypink}\textbf{59.6~\stdvu{±0.3}}} & {\cellcolor{mypink}\textbf{14.2~\stdvu{±0.2}}} &
{\cellcolor{mypink}\textbf{66.0~\stdvu{±0.3}}} &{\cellcolor{mypink}\textbf{68.0~\stdvu{±0.3}}} & {\cellcolor{mypink}\textbf{- 0.63\%}} \\
\bottomrule
\end{tabular}}
\end{table}

\subsection{Failure Case}
\label{sec:fail_case}
\begin{figure*}
    \centering
    \subfigure{
        \begin{minipage}{0.18\linewidth}
            \centering
            \includegraphics[width=0.993\textwidth,height=0.7in]{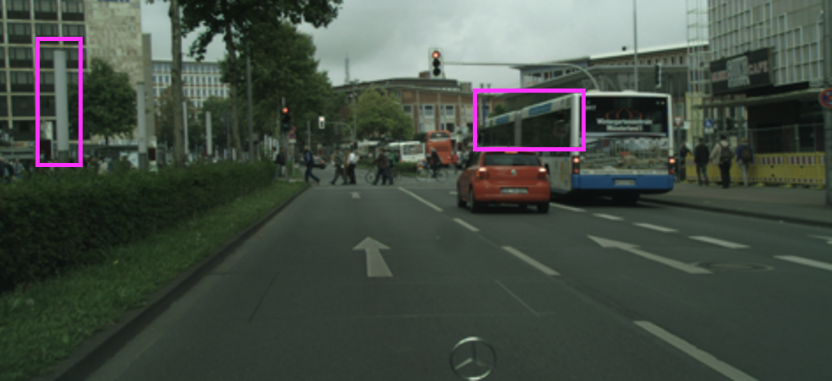}\\
        \end{minipage}%
    }%
    \subfigure{
        \begin{minipage}{0.18\linewidth}
            \centering
            \includegraphics[width=0.993\textwidth,height=0.7in]{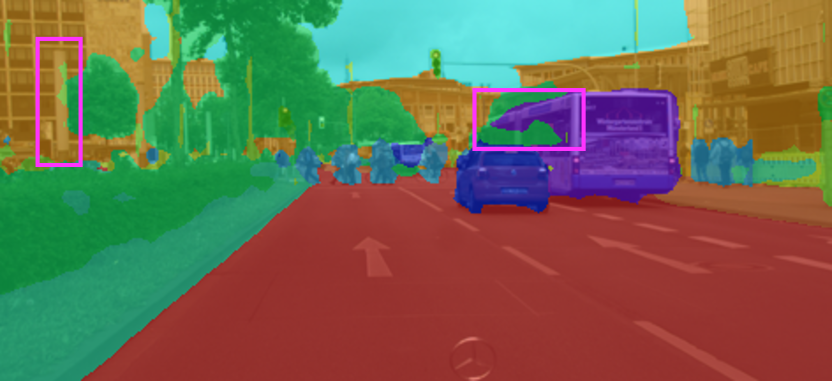}\\
        \end{minipage}%
    }%
    \subfigure{
        \begin{minipage}{0.18\linewidth}
            \centering
            \includegraphics[width=0.993\textwidth,height=0.7in]{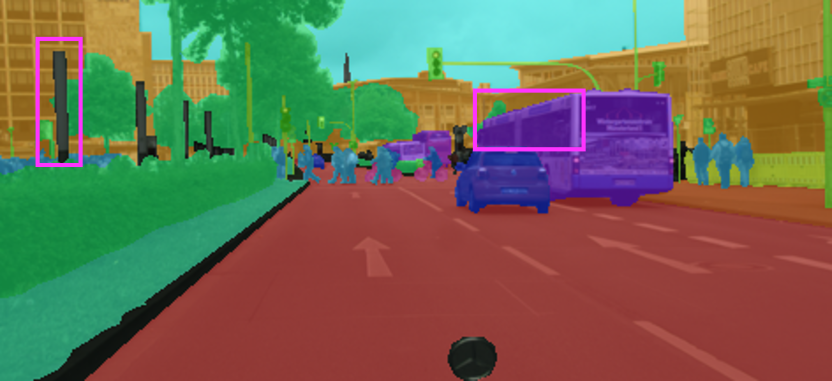}\\
        \end{minipage}%
    }%
    \subfigure{
        \begin{minipage}{0.18\linewidth}
            \centering
            \includegraphics[width=0.993\textwidth,height=0.7in]{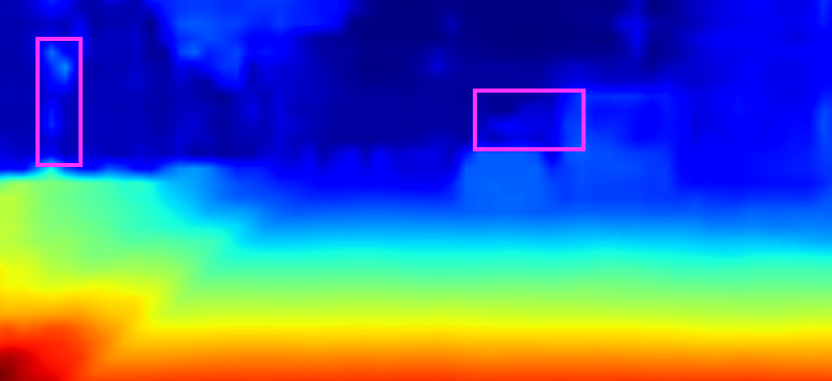}\\
        \end{minipage}%
    }%
    \subfigure{
        \begin{minipage}{0.18\linewidth}
            \centering
            \includegraphics[width=0.993\textwidth,height=0.7in]{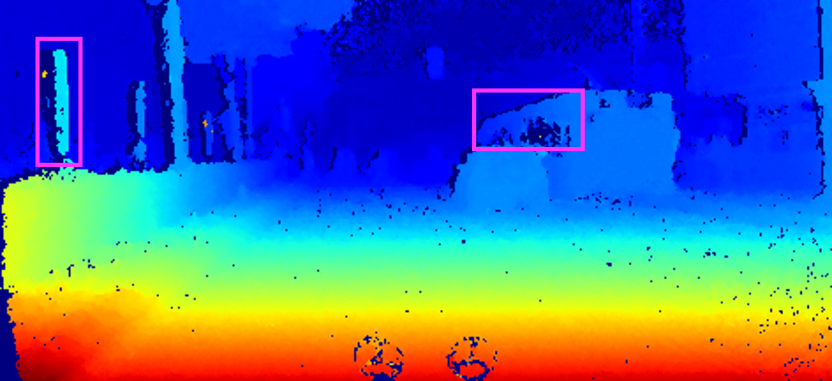}\\
        \end{minipage}%
    }%
    \vspace{-3mm}
    \setcounter{subfigure}{0}
    \subfigure[Image]{
        \begin{minipage}{0.18\linewidth}
            \centering
            \includegraphics[width=0.993\textwidth,height=0.7in]{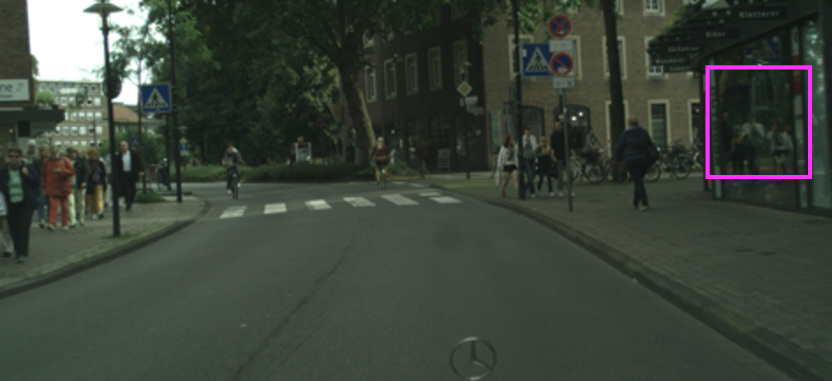}\\
            \vspace{1mm}
        \end{minipage}%
    }%
    \subfigure[seg.(ours)]{
        \begin{minipage}{0.18\linewidth}
            \centering
            \includegraphics[width=0.993\textwidth,height=0.7in]{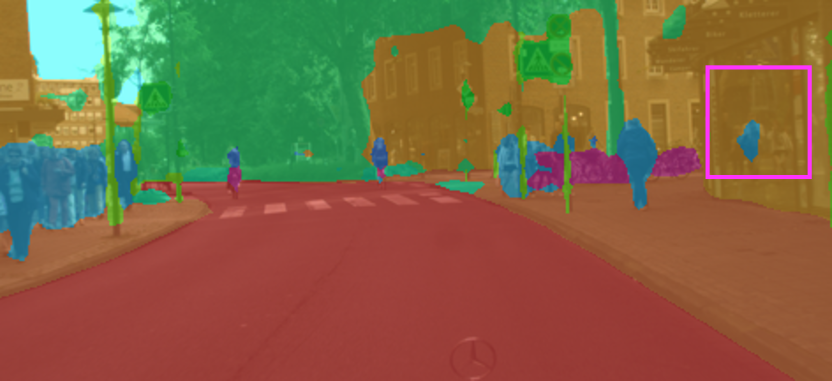}\\
            \vspace{1mm}
        \end{minipage}%
    }%
    \subfigure[seg.(GT)]{
        \begin{minipage}{0.18\linewidth}
            \centering
            \includegraphics[width=0.993\textwidth,height=0.7in]{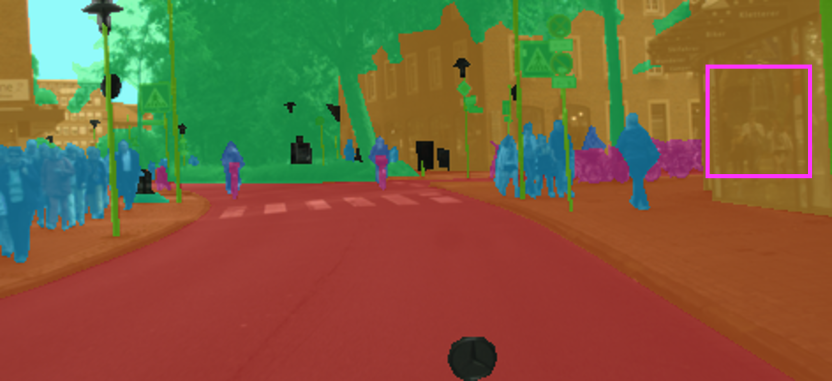}\\
            \vspace{1mm}
        \end{minipage}%
    }%
    \subfigure[disparity(ours)]{
        \begin{minipage}{0.18\linewidth}
            \centering
            \includegraphics[width=0.993\textwidth,height=0.7in]{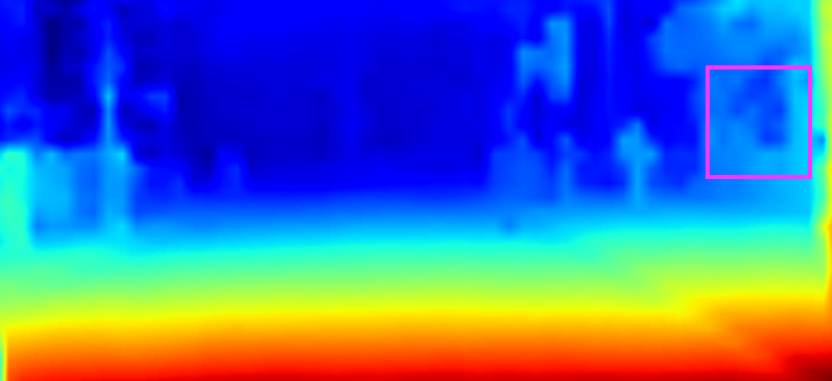}\\
            \vspace{1mm}
        \end{minipage}%
    }%
    \subfigure[disparity(GT)]{
        \begin{minipage}{0.18\linewidth}
            \centering
            \includegraphics[width=0.993\textwidth,height=0.7in]{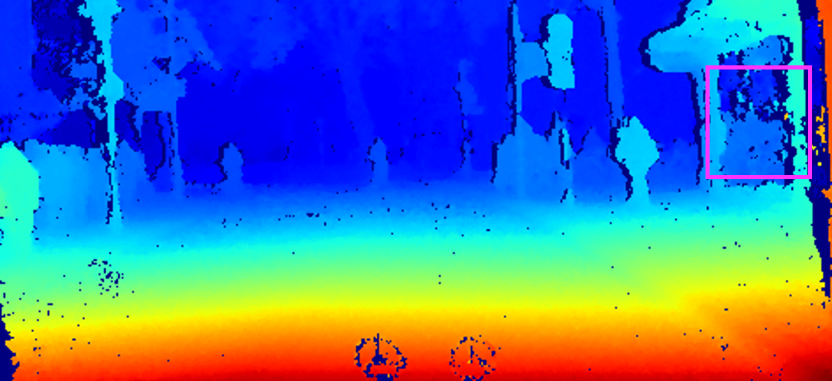}\\
            \vspace{1mm}
        \end{minipage}%
    }%
    \vspace{-3mm}
    \caption{Visualization of Fail cases on \textbf{Cityscapes}. ''GT`` indicates ground truth. ''ours`` refers to a model that uses semantic segmentation as its primary task and depth regression as an auxiliary task, which is trained using the method we proposed.}
    \label{fig:fail_cases}
    \vspace{-6mm}
\end{figure*}

In this section, we primarily delve into the ramifications of employing equal training for these tasks. 
As shown in Table~\ref{table:dense_prediction} and Table~\ref{table:nyu_noise}, the integration of two pseudo auxiliary tasks results in performance decreases of 13.41\%, 8.6\%, 5.28\%, and 3.72\% for IMTL, GradNorm, UW, and Uniform, respectively. 
Secondly, Fig.~\ref{fig:fail_cases} showcases the instances of failure in employing equal training with semantic segmentation as the primary task and disparity as the auxiliary task on the Cityscapes dataset. These examples elucidate the potential misguidance inflicted upon the model due to the negative transfer between main and auxiliary tasks. 
For example, situations where the same object at varying depths is depicted in the first row, along with erroneous annotations shown in the third row, may lead to semantic segmentation mistakenly dividing a single object into multiple parts, due to being misled by the training of disparity estimation. 
Both observations conform to the definition of negative transfer, that is, an improvement in performance on one task leads to a decline in performance on another task.
Hence, properly training and focusing on auxiliary tasks can unintentionally cause negative transfer, disrupting the balance between main and auxiliary tasks.
In order to resolve this problem, our approach mitigates the adverse effects by applying equal training exclusively to the parameters unique to the auxiliary tasks, thereby reducing the negative impact.

\subsection{Results on Multi-domain Classification.}
We further evaluate our methods for image classification tasks in a multi-domain setting. We trained on CIFAR-$100$~\cite{cifar100} and treated each of the $20$ ‘coarse’ classes as one domain, thus creating a dataset with $20$ tasks, where each task is a $5$-class classification over the dataset’s ‘fine’ classes. For the \texttt{Normal} and \texttt{Auxiliary setting}s, we trained all methods on a VGG-$16$ network~\cite{vgg} with standard hard-parameter sharing, where each task has a task-specific prediction layer.

In Table~\ref{table:mdc}, we show the classification accuracy of the five most challenging domains along with the average performance across all $20$ domains. 
Multi-tasking learning in this dataset is more challenging because the number of parameters per task is reduced by a factor of $20$ compared to single-tasking. Under this setting, we observed that all multi-tasking baselines achieved overall performance similar to single-tasking learning due to the limited parameter space for each task.
However, our method still improves the overall performance by a non-trivial margin. Similarly, our method can further improve performance in the \texttt{Auxiliary setting}, with higher per-task performance in challenging domains of around $2$\%.

\begin{table}[ht]
\caption{Comparison of the multi-task performance under the \texttt{Auxiliary setting} on the multi-domain CIFAR-100 dataset, where the performance of the worst $5$ single tasks and the average accuracy across all $20$ coarse classes are reported. All results are reimplemented.\label{table:mdc}}
\vspace{-2mm}
\centering
\renewcommand\arraystretch{1.2}
\resizebox{\linewidth}{!}{
\begin{tabular}{l ccccc c}
\toprule
\multirow{2}{*}{Method} & \multirow{2}{*}{People} & Aquatic & Smalls & \multirow{2}{*}{Trees} & \multirow{2}{*}{Reptiles} & Avg. \\ 
 & & Animals & Mamals & & &(20 classes)\\
\cmidrule(r){1-1} \cmidrule(lr){2-2} \cmidrule(lr){3-3} \cmidrule(lr){4-4} \cmidrule(r){5-5} \cmidrule(r){6-6} \cmidrule(r){7-7}

Single task    &  \multicolumn{1}{|c}{56.60~\stdvu{±0.31}}  & 67.40~\stdvu{±0.42} & 71.80~\stdvu{±0.33} & 75.40~\stdvu{±0.21} & 75.80~\stdvu{±0.44} & \multicolumn{1}{|c}{82.30} \\
Uniform    &  \multicolumn{1}{|c}{57.56~\stdvu{±0.35}}  & 73.57~\stdvu{±0.37} & 74.14~\stdvu{±0.33} & 74.55~\stdvu{±0.22} & 76.71~\stdvu{±0.40} & \multicolumn{1}{|c}{81.45} \\
Uncert.~\cite{uw}    &  \multicolumn{1}{|c}{55.99~\stdvu{±0.38}}  & 71.00~\stdvu{±0.40} & 72.20~\stdvu{±0.34} & 73.00~\stdvu{±0.21} & 76.40~\stdvu{±0.40} & \multicolumn{1}{|c}{82.62} \\
Auto-$\lambda$~(MTL)~\cite{autolambda}    &  \multicolumn{1}{|c}{57.57~\stdvu{±0.45}}  & 74.00~\stdvu{±0.37} & 75.05~\stdvu{±0.42} & 75.15~\stdvu{±0.31} & 77.55~\stdvu{±0.50} & \multicolumn{1}{|c}{83.92} \\
\hline
GCS~\cite{gcs}    &  \multicolumn{1}{|c}{57.66~\stdvu{±0.32}}  & 73.33~\stdvu{±0.34} & 73.98~\stdvu{±0.33} & 76.83~\stdvu{±0.24} & 77.45~\stdvu{±0.45} & \multicolumn{1}{|c}{82.77} \\
OL-AUX~\cite{ol_aux}    &  \multicolumn{1}{|c}{56.45~\stdvu{±0.32}}  & 74.41~\stdvu{±0.33} & 73.40~\stdvu{±0.34} & 76.32~\stdvu{±0.30} & 77.21~\stdvu{±0.44} & \multicolumn{1}{|c}{82.02} \\
Auto-$\lambda$~\cite{autolambda}    &  \multicolumn{1}{|c}{60.89~\stdvu{±0.39}}  & 75.70~\stdvu{±0.44} & 75.64~\stdvu{±0.45} & 77.38~\stdvu{±0.35} & 81.75~\stdvu{±0.50} & \multicolumn{1}{|c}{84.92} \\
{\cellcolor{mypink}\texttt{\textbf{Ours}}}    &  \multicolumn{1}{|c}{{\cellcolor{mypink}\textbf{61.54~\stdvu{±0.34}}}}  & {\cellcolor{mypink}\textbf{76.33~\stdvu{±0.36}}} & {\cellcolor{mypink}\textbf{77.22~\stdvu{±0.35}}} & {\cellcolor{mypink}\textbf{77.91~\stdvu{±0.25}}} & {\cellcolor{mypink}\textbf{81.92~\stdvu{±0.44}}} & \multicolumn{1}{|c}{{\cellcolor{mypink}\textbf{86.87}}} \\
\bottomrule
\end{tabular}}
\end{table}

%------------------------------------------------------------------------
\subsection{Ablation Study}

\subsubsection{Two type of balance in the encoder stage.}
Our weighting method in the encoder stage is formulated in two parts, \ie,  the gradient re-normalization and robust uncertainty weights. To explore the influence of these two parts, we split our weighting strategy into two modules, \ie,  \texttt{\textbf{gradient-only}} and \texttt{\textbf{uncertainty-only}} modules.
However, the \texttt{\textbf{gradient-only}} modules use gradient norm to balance and need an additional weighting mechanism under the \texttt{Auxiliary setting}. We first adopt the mechanism of fixed weights, \ie,  $0.1$ for each auxiliary task and $1.0$ for the primary tasks. Additionally, we provide grid search results that pick the best result within the grid search for auxiliary tasks weighted between [$0.1$, $0.5$).

Note that we also report the grid search of all task weights under the \texttt{Auxiliary setting}. 
As in shown in Table~\ref{table:parts}, gradient parts with grid search achieve the highest performance compared to their fix-weight and loss versions. However, the loss part still outperform the baseline by $2$\% and shows the comparability of commonly used gradient based methods. Finally, combining both parts gets the highest improvement across all variants. The results indicate that both loss and gradient parts can contribute to the \texttt{Auxiliary setting}.

\begin{table}[ht]
\caption{Ablation studies on both balancing strategies in the encoder stage with regards to different variations, including 'GS': the best performance with grid search of task-specific weights between [$0.1$, $0.5$); 'Grad.(Fix)' / 'Grad.(GS)': only the gradient re-norm module with manually chosen and searched auxiliary weights is applied; 'Uncer.': only the uncertainty module is reported. Experiments are on \textbf{NYUv2} benchmark.  
\vspace{-2mm}
\label{table:parts}}
\centering
\renewcommand\arraystretch{1.2}
\resizebox{\linewidth}{!}{
\begin{tabular}{l cc cccc}
\toprule
  Method & uncer. & grad.& Depth        & Segm. & Normals & $ \Delta MTL $  \\
\cmidrule(r){1-1} \cmidrule(lr){2-3} \cmidrule(lr){4-4} \cmidrule(lr){5-5} \cmidrule(r){6-6} \cmidrule(r){7-7}
Single Task  &  \multicolumn{1}{|c}{} &   &  \multicolumn{1}{|c}{0.5877~\stdvu{±0.0006}}  & 43.58~\stdvu{±0.05}   & 19.49~\stdvu{±0.03}   & + 0.00\% \\
\hline
Uniform      &  \multicolumn{1}{|c}{} &   &  \multicolumn{1}{|c}{0.5933~\stdvu{±0.0008}} & 43.47~\stdvu{±0.03}& 21.91~\stdvu{±0.03} & - 4.45\%   \\
GS  &  \multicolumn{1}{|c}{} &   &  \multicolumn{1}{|c}{0.5846~\stdvu{±0.0008}} & 43.99~\stdvu{±0.03} & 21.35~\stdvu{±0.04} & - 2.69\% \\
Grad.(Fix) &  \multicolumn{1}{|c}{} & \checkmark  &  \multicolumn{1}{|c}{0.5811~\stdvu{±0.0009}} & 43.98~\stdvu{±0.03} & 21.22~\stdvu{±0.03} & - 2.28\% \\
Grad.(GS) &  \multicolumn{1}{|c}{} & \checkmark &  \multicolumn{1}{|c}{0.5807~\stdvu{±0.0007}} & 44.21~\stdvu{±0.03} & 21.03~\stdvu{±0.03} & - 1.75\%  \\
Uncer.     &  \multicolumn{1}{|c}{\checkmark} &   &  \multicolumn{1}{|c}{0.5832~\stdvu{±0.0006}} & 44.01~\stdvu{±0.03} & 21.41~\stdvu{±0.03} & - 2.69\%  \\
{\cellcolor{mypink}\texttt{\textbf{Ours}}}             &  \multicolumn{1}{|c}{{\cellcolor{mypink}\checkmark}} & {\cellcolor{mypink}\checkmark}  &  \multicolumn{1}{|c}{{\cellcolor{mypink}0.5751~\stdvu{±0.0009}}} & {\cellcolor{mypink}44.60~\stdvu{±0.11}} & {\cellcolor{mypink}20.62~\stdvu{±0.03}} & {\cellcolor{mypink}- 0.44\% }\\
\bottomrule
\end{tabular}}
\end{table}

\begin{table}[ht]
\caption{Ablation studies on the influence of the impartial learning of the decoders on multi-task learning, where both the partial~(Fixed, Grid Search, Auto-$\lambda$) and impartial learning~(including GradNorm, DWA, RLW-Normal and ours with uncertainty) are reported. Experiments are on \textbf{NYUv2} benchmark. All results are reimplemented. 
\vspace{-2mm}
\label{table:impart}}
\centering
\renewcommand\arraystretch{1.2}
\resizebox{\linewidth}{!}{
\begin{tabular}{l c cc}
\toprule
  \multirow{2}{*}{Method}& Segmentation & Depth & Normals \\
  & mIoU[\%]($\uparrow$) & RMSE[m]($\downarrow$)  & Mean Error($\downarrow$) \\
\cmidrule(r){1-1} \cmidrule(lr){2-2} \cmidrule(lr){3-4}
Auto-$\lambda$(impartial)                & \multicolumn{1}{|c}{44.54~\stdvu{±0.10}}  & \multicolumn{1}{|c}{0.5818~\stdvu{±0.0013}}  & 21.34~\stdvu{±0.03}  \\
Auto-$\lambda$(partial)                  & \multicolumn{1}{|c}{44.07~\stdvu{±0.11}}  & \multicolumn{1}{|c}{0.5896~\stdvu{±0.0014}}  & 21.89~\stdvu{±0.04}  \\
\hline \hline
Uniform                            & \multicolumn{1}{|c}{44.12~\stdvu{±0.05}}   & \multicolumn{1}{|c}{0.5942~\stdvu{±0.0007}}  & 21.80~\stdvu{±0.03}  \\
Fixed                             & \multicolumn{1}{|c}{43.77~\stdvu{±0.05}} & \multicolumn{1}{|c}{0.6045~\stdvu{±0.0008}}  & 22.78~\stdvu{±0.03}  \\
Grid Search                       & \multicolumn{1}{|c}{43.90~\stdvu{±0.07}}  & \multicolumn{1}{|c}{0.5977~\stdvu{±0.0006}}  & 22.02~\stdvu{±0.03}  \\
\textit{w}. GradNorm                    & \multicolumn{1}{|c}{43.99~\stdvu{±0.09}} & \multicolumn{1}{|c}{0.5910~\stdvu{±0.0011}}  & 21.66~\stdvu{±0.03} \\
\textit{w}. DWA                         & \multicolumn{1}{|c}{44.21~\stdvu{±0.10}}  & \multicolumn{1}{|c}{0.5898~\stdvu{±0.0005}}  & 21.95~\stdvu{±0.03} \\
\textit{w}. RLW-Normal                  & \multicolumn{1}{|c}{44.38~\stdvu{±0.11}} & \multicolumn{1}{|c}{0.5855~\stdvu{±0.0007}}  & 21.72~\stdvu{±0.03} \\
{\cellcolor{mypink}\texttt{\textbf{Ours}}~(\textit{w}. Uncertainty)}  & \multicolumn{1}{|c}{{\cellcolor{mypink}44.60~\stdvu{±0.11}}} & \multicolumn{1}{|c}{{\cellcolor{mypink}0.5799~\stdvu{±0.0011}}}  & {\cellcolor{mypink}21.00~\stdvu{±0.03}} \\
\bottomrule
\end{tabular}}
\end{table}

\subsubsection{Impartial Learning in the Task-specific Decoders}

Another essential part of our framework is impartial decoder learning. By default, we utilize Uncertainty Weight~\cite{uw} to weight each task separately during training decoders instead of setting a uniform weight, \textit{e}.\textit{g}.  $1.0$ for all tasks. To illustrate the superior performance boosted by this choice, we replace our default choice with three default settings in multi-task learning, \ie,  a) Uniform: the weight of all the tasks is set to $1.0$. b) Fixed: the weights are set to $1.0$ for primary tasks and $0.1$ for auxiliary tasks. c) Grid Search: search optimal weights for each task by grid. We also compare our method with other commonly used weighting methods, \ie,  GradNorm~\cite{gradnorm}, DWA~\cite{mtan_dwa} and RLW with Normal Distribution~\cite{rlw}~(shorted as RLW-Normal). Details are reported in the first subtable in Table~\ref{table:impart}. As can be seen from this subtable, our method not only promotes the performance of primary tasks but also obviously helps the training of auxiliary tasks~(depth regression and surface normal prediction). 

To illustrate the gain achieved by impartial learning, we also modify Auto-$\lambda$~\cite{autolambda} with impartial learning. Similarly, we separate the training objectives of both the encoder and decoder, where the encoder is still optimized by the meta-learning objective of Auto-$\lambda$~\cite{autolambda} while the decoders are trained with our impartial objective, \ie,  Eq.~\ref{eq:ug_loss}. As shown in the second subtable in Table~\ref{table:impart}, we observe an obvious improvement over the original Auto-$\lambda$~\cite{autolambda} on the \textbf{NYUv2} benchmark.
As can be seen from this subtable, our impartial learning can improve the performance of auxiliary tasks, \ie,  for depth regression from $0.5896$ to $\textbf{0.5818}$, for surface normals prediction from $21.89$ to $\textbf{21.34}$, thereby promoting the primary tasks, \ie,  $44.07\%$ $\rightarrow$ $\textbf{44.54}\%$.
For impartial learning, Uniform weighting obtains values of $44.12$\% on the semantic segmentation task. 
This is attribute to our task-specific learning with more attention on the auxiliary tasks that can in turn improve the performance of primary tasks. 
However, GradNorm~\cite{gradnorm} performs poorly on semantic segmentation, while DWA~\cite{mtan_dwa} and RLW~\cite{rlw} with normal distribution outperform the uniform baseline. 
For partial learning, all the baseline methods, including Fixed weight, Grid Search and Auto-$\lambda$ in the partial mode all fail to reach the comparable performance of the uniform baseline. Comparing between the different modes of Auto-$\lambda$, we can see an obvious gap between impartial and partial learning, \ie,  $44.54\%$ \textit{vs.} $44.07\%$. 
Instead, our method with uncertainty weights still achieves the best performance across all counterparts.

\subsubsection{Influence of Worse Auxiliary Tasks}
\begin{figure}[!t]
    \centering
    \subfigure[Depth as auxiliary]{
        \begin{minipage}{0.5\linewidth}
            \centering
            \includegraphics[width=0.993\textwidth,height=1.2in]{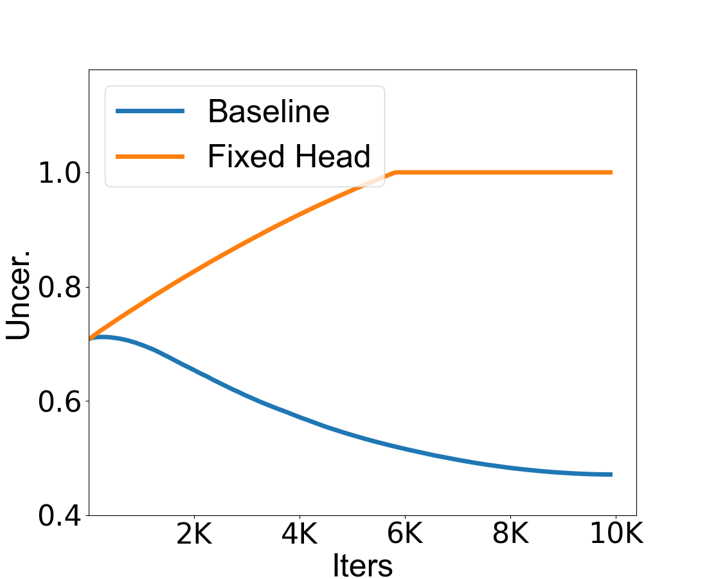}\\
            \vspace{1mm}
        \end{minipage}%
    }%
    \subfigure[Segmentation as primary]{
        \begin{minipage}{0.5\linewidth}
            \centering
            \includegraphics[width=0.993\textwidth,height=1.2in]{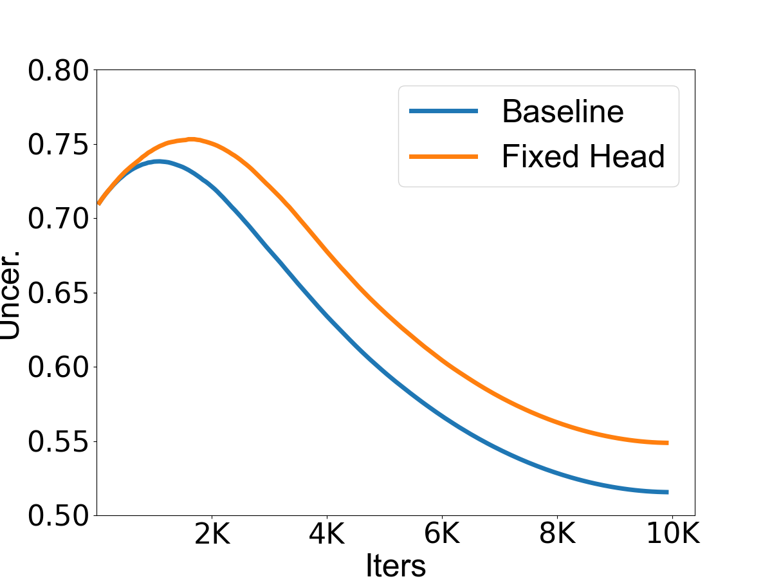}\\
            \vspace{1mm}
        \end{minipage}%
    }%
    % \vspace{-5pt}
    \vspace{-3mm}
    \caption{Uncertainty curve of baseline and fixed head on NYUv2~\cite{nyuv2} under both (a) Depth as the auxiliary task and (b) Segmentation as the primary task. }
    \label{fig:ablation2}
    \vspace{-6mm}
\end{figure}

Here, we investigate the influence of a broken auxiliary task on the primary task. In order to show this, we designed an experiment that fixed the head of depth regression on the standard \textbf{NYUv2} benchmark. As shown in Fig.~\ref{fig:ablation2}, the uncertainty of depth regression obviously drops compared to normal training. It indicates that uncertainty can be a good indicator of the training state. Additionally, we plot the uncertainty of the primary task in Fig.~\ref{fig:ablation2}. From this figure, we can see that the uncertainty drops during the whole process while still maintaining the main trend, \ie,  first dropping and then going higher.

We also summarize the performance of these experiments in Table~\ref{table:fix_head}. 
The results illustrate that a broken auxiliary task may perform negative transfer to the primary task, and previous methods cannot resolve this problem.
Comparing to Auto-$\lambda$ and Uncertainty Weights, our method still keeps acceptable performance except for depth regression, \ie,  drops by about $1.89$\% on segmentation and surface normals by weakening the weights of depth regression. 

\begin{table}[ht]
\caption{Ablation studies of the multi-task results with fixed and unfixed decoders of the \texttt{Auxiliary tasks}. Experiments are on \textbf{NYUv2} benchmark. All results are reimplemented.
\vspace{-2mm}
\label{table:fix_head}}
\centering
\renewcommand\arraystretch{1.2}
\resizebox{\linewidth}{!}{
\begin{tabular}{l cccc}
\toprule
  \multirow{2}{*}{Method}& Depth        & Segmentation & Normals & $ \Delta MTL $  \\
  & RMSE[m]($\downarrow$) & mIoU[\%]($\uparrow$) & Mean Error($\downarrow$) & [\%]($\uparrow$) \\
\cmidrule(r){1-1} \cmidrule(lr){2-2} \cmidrule(lr){3-3} \cmidrule(lr){4-4} \cmidrule(r){5-5}
Single    & \multicolumn{1}{|c}{0.5877~\stdvu{±0.0006}}  & 43.58~\stdvu{±0.05}   & 19.49~\stdvu{±0.03}   & + 0.00\% \\
Uniform   & \multicolumn{1}{|c}{0.5933~\stdvu{±0.0008}} & 43.47~\stdvu{±0.03} & 21.91~\stdvu{±0.03} & - 4.45\%   \\
Uncer.    & \multicolumn{1}{|c}{0.5874~\stdvu{±0.0004}} & 43.97~\stdvu{±0.06} & 21.61~\stdvu{±0.03} & - 3.31\% \\
Uncer. (Fix) &\multicolumn{1}{|c}{1.8050~\stdvu{±0.0015}} & 32.77~\stdvu{±0.10} & 23.89~\stdvu{±0.05} & - 47.38\% \\
\hline
Auto-$\lambda$                        & \multicolumn{1}{|c}{0.5803~\stdvu{±0.0010}} & 43.86~\stdvu{±0.10} & 21.57~\stdvu{±0.04} & - 1.09\%  \\
Auto-$\lambda$ (Fix)                   & \multicolumn{1}{|c}{1.8110~\stdvu{±0.0009}} & 44.05~\stdvu{±0.11} & 21.23~\stdvu{±0.03} & - 7.85\%  \\
{\cellcolor{mypink}\texttt{\textbf{Ours}}}   & \multicolumn{1}{|c}{{\cellcolor{mypink}0.5732~\stdvu{±0.0009}}} & {\cellcolor{mypink}44.82~\stdvu{±0.08}} & {\cellcolor{mypink}19.36~\stdvu{±0.03}} & {\cellcolor{mypink}+ 2.43\%} \\
{\cellcolor{mypink}\texttt{\textbf{Ours}} (Fix)}               & \multicolumn{1}{|c}{{\cellcolor{mypink}1.7990~\stdvu{±0.0013}}} & {\cellcolor{mypink}44.56~\stdvu{±0.12}} & {\cellcolor{mypink}20.77~\stdvu{±0.05}} & {\cellcolor{mypink}- 4.32\%} \\
\bottomrule
\end{tabular}}
\end{table}

\subsubsection{Inplacement of the uncertainty in the encoder stage}
\label{sec:decoder_weighting}
Loss-based methods always use different kinds of indicators to weight, \textit{e}.\textit{g}.  uncertainty in UW~\cite{uw}, expected gradient norm in GradNorm~\cite{gradnorm}, relative training speed in DWA~\cite{mtan_dwa}, and commonly used cosine similarity.
We replace the uncertainty balance with these metrics in the encoder stage, in order to examine our choice of uncertainty-base balance. Note that the task-specific parameters are still trained via the impartial multi-task objective~(Eq.~\ref{eq:mtl_loss}) to make a fair comparison between experiments. In a word, we only change the training process in the encoder stage of our method.

We summarize the results in Table~\ref{table:replace_uncertainty}. It can be seen from this table that, only with the expected gradient norm proposed by GradNorm, the relative performance improves compared to the baseline without loss-based part. Cosine Similarity achieves the best performance within the rest metrics and is comparable with the baseline. Uncertainty improves the performance by a large gap, \ie,  $1.68$\% to GradNorm version and $1.84$\% to the baseline. These results show that the utilization of uncertainty in the balance of encoder stage serves as a favorable metric for evaluating the efficacy of auxiliary task training.

\begin{table}[ht]
\caption{Comparison of multi-task performance on \textbf{NYUv2} between different alternatives of 'Uncertainty' measurement in robust uncertainty weighting (Sec. \ref{sec:decoder_weighting}), where 'Fixed' uses grid to estimate 'uncertainty', relative Training Speed~\cite{mtan_dwa} and Expected Gradient Norm~\cite{gradnorm} are referred to RTS and EGN respectively, Cos. indicates cosine similarity between task-specific gradients. 
\vspace{-2mm}
\label{table:replace_uncertainty}}
\centering
\renewcommand\arraystretch{1.2}
\resizebox{\linewidth}{!}{
\begin{tabular}{c cccc}
\toprule
  \multirow{2}{*}{Method}& Depth        & Segmentation & Normals & $ \Delta MTL $  \\
  & RMSE[m]($\downarrow$) & mIoU[\%]($\uparrow$) & Mean Error($\downarrow$) & [\%]($\uparrow$) \\
\cmidrule(r){1-1} \cmidrule(lr){2-2} \cmidrule(lr){3-3} \cmidrule(lr){4-4} \cmidrule(r){5-5}
Single    & \multicolumn{1}{|c}{0.5877~\stdvu{±0.0006}}  & 43.58~\stdvu{±0.05}   & 19.49~\stdvu{±0.03}   & + 0.00\% \\

Uniform        & \multicolumn{1}{|c}{0.5933~\stdvu{±0.0008}} & 43.47~\stdvu{±0.03} & 21.91~\stdvu{±0.03} & - 4.45\%   \\

Fixed  & \multicolumn{1}{|c}{0.5811~\stdvu{±0.0007}} & 43.98~\stdvu{±0.12} & 21.22~\stdvu{±0.03} & - 2.28\% \\
RTS   & \multicolumn{1}{|c}{0.5847~\stdvu{±0.0010}} & 43.78~\stdvu{±0.10} & 21.47~\stdvu{±0.04} & - 3.05\% \\
EGN & \multicolumn{1}{|c}{0.5803~\stdvu{±0.0009}} & 44.02~\stdvu{±0.09} & 21.17~\stdvu{±0.03} & - 2.12\%  \\
Cos.  & \multicolumn{1}{|c}{0.5823~\stdvu{±0.0012}} & 43.95~\stdvu{±0.11} & 21.27~\stdvu{±0.04} & - 2.46\%  \\

{\cellcolor{mypink}\texttt{\textbf{Ours}} (Uncer.)}        & \multicolumn{1}{|c}{{\cellcolor{mypink}0.5751~\stdvu{±0.0011}}} & {\cellcolor{mypink}44.60~\stdvu{±0.11}} & {\cellcolor{mypink}20.62~\stdvu{±0.03}} & {\cellcolor{mypink}- 0.44\%} \\
\bottomrule
\end{tabular}}
\end{table}

\subsubsection{Weight changing over training}
\label{sec:dynamic}
\begin{figure}[!t]
    \centering
    \subfigure{
        \rotatebox[origin=c]{90}{\centering\scriptsize{Decoder}}
        \begin{minipage}{0.3\linewidth}
            \centering
            \includegraphics[width=0.993\textwidth,height=1.2in]{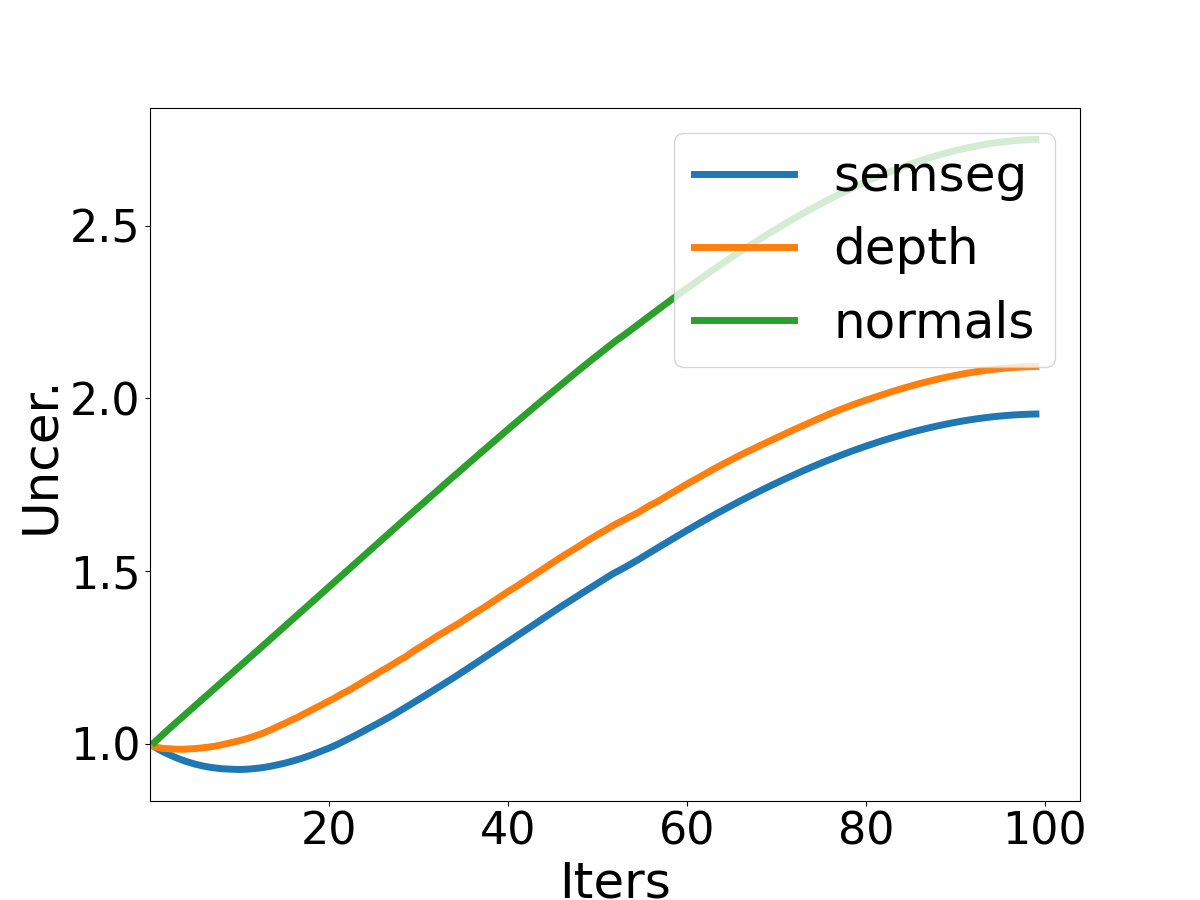}\\
        \end{minipage}%
    }%
    \subfigure{
        \begin{minipage}{0.3\linewidth}
            \centering
            \includegraphics[width=0.993\textwidth,height=1.2in]{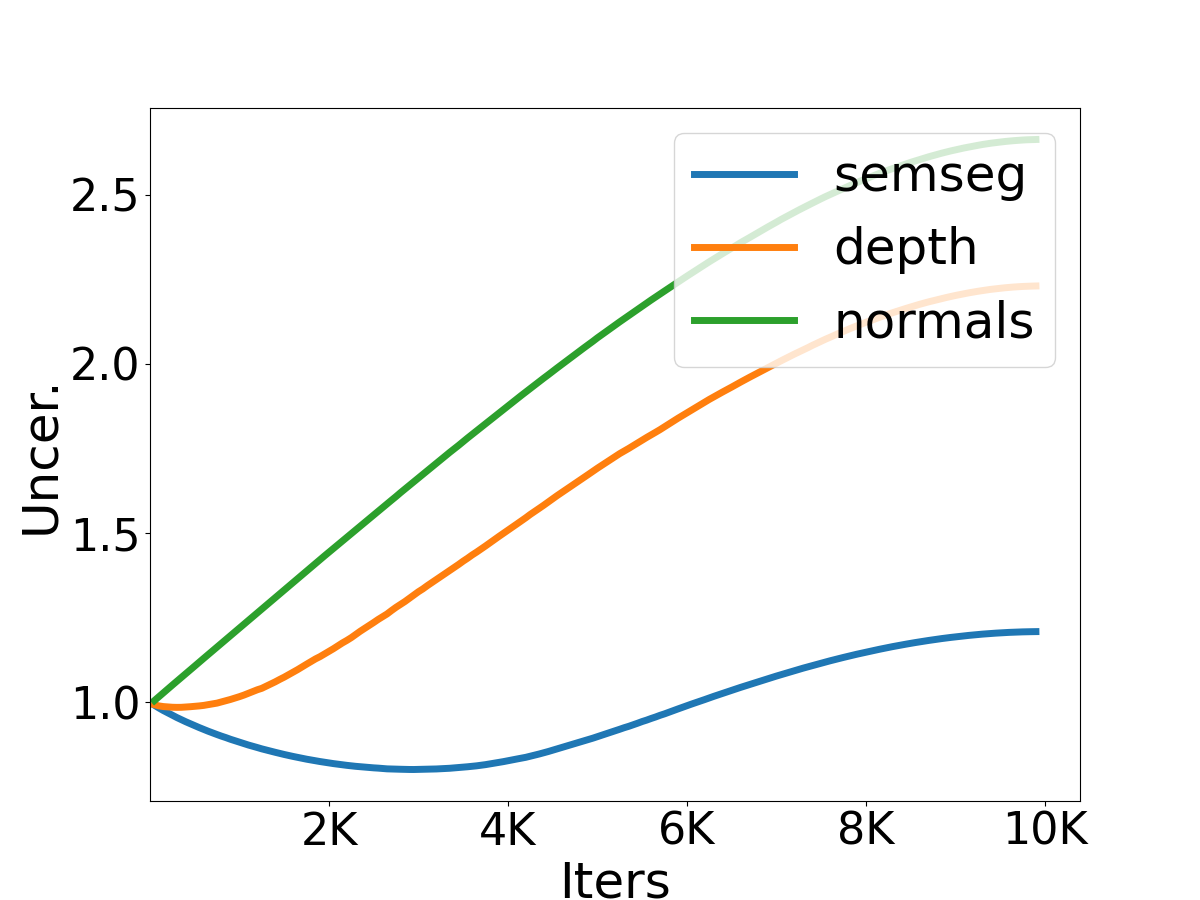}\\
        \end{minipage}%
    }%
    \subfigure{
        \begin{minipage}{0.3\linewidth}
            \centering
            \includegraphics[width=0.993\textwidth,height=1.2in]{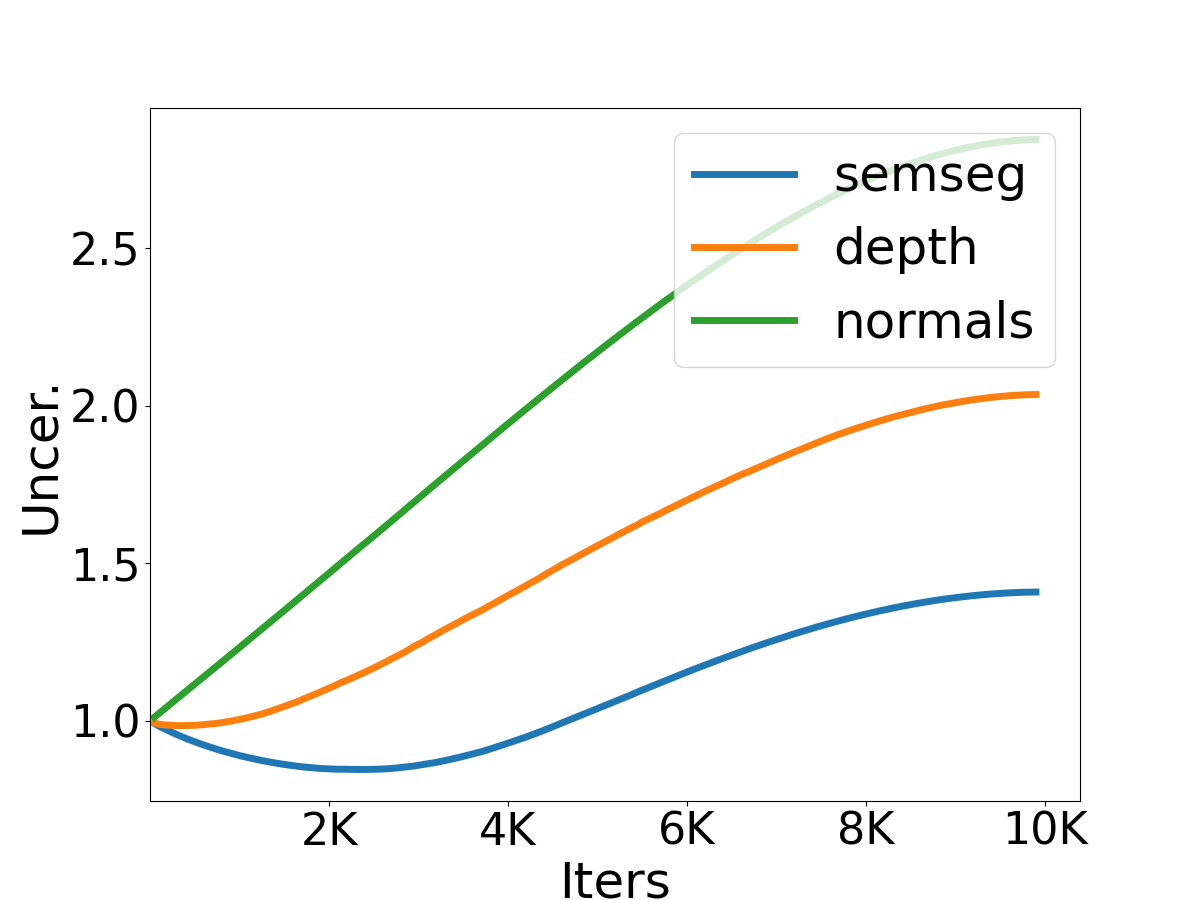}\\
        \end{minipage}%
    }%
    \vspace{-3mm}
    \subfigure{
        \rotatebox[origin=c]{90}{\centering\scriptsize{Encoder}}
        \begin{minipage}{0.3\linewidth}
            \centering
            \includegraphics[width=0.993\textwidth,height=1.2in]{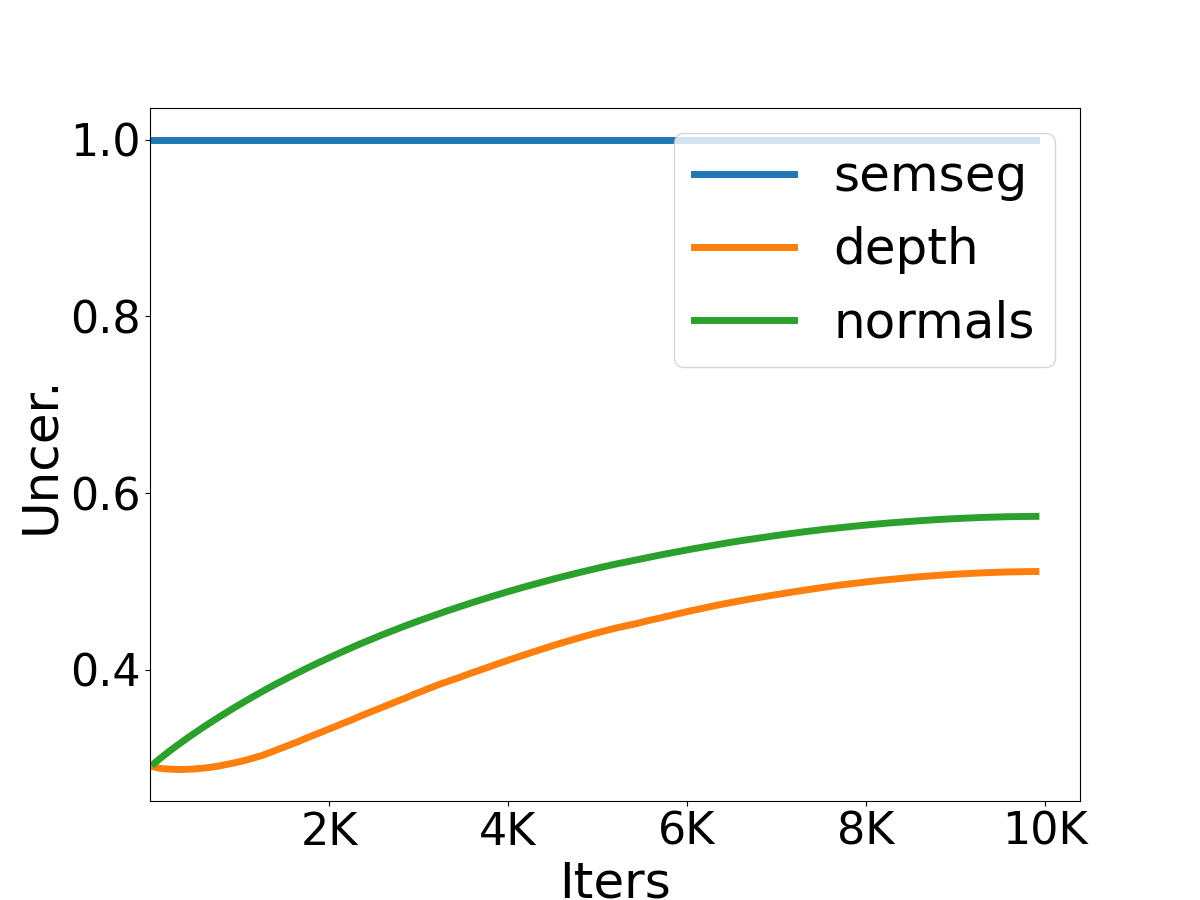}\\
        \end{minipage}%
    }%
    \subfigure{
        \begin{minipage}{0.3\linewidth}
            \centering
            \includegraphics[width=0.993\textwidth,height=1.2in]{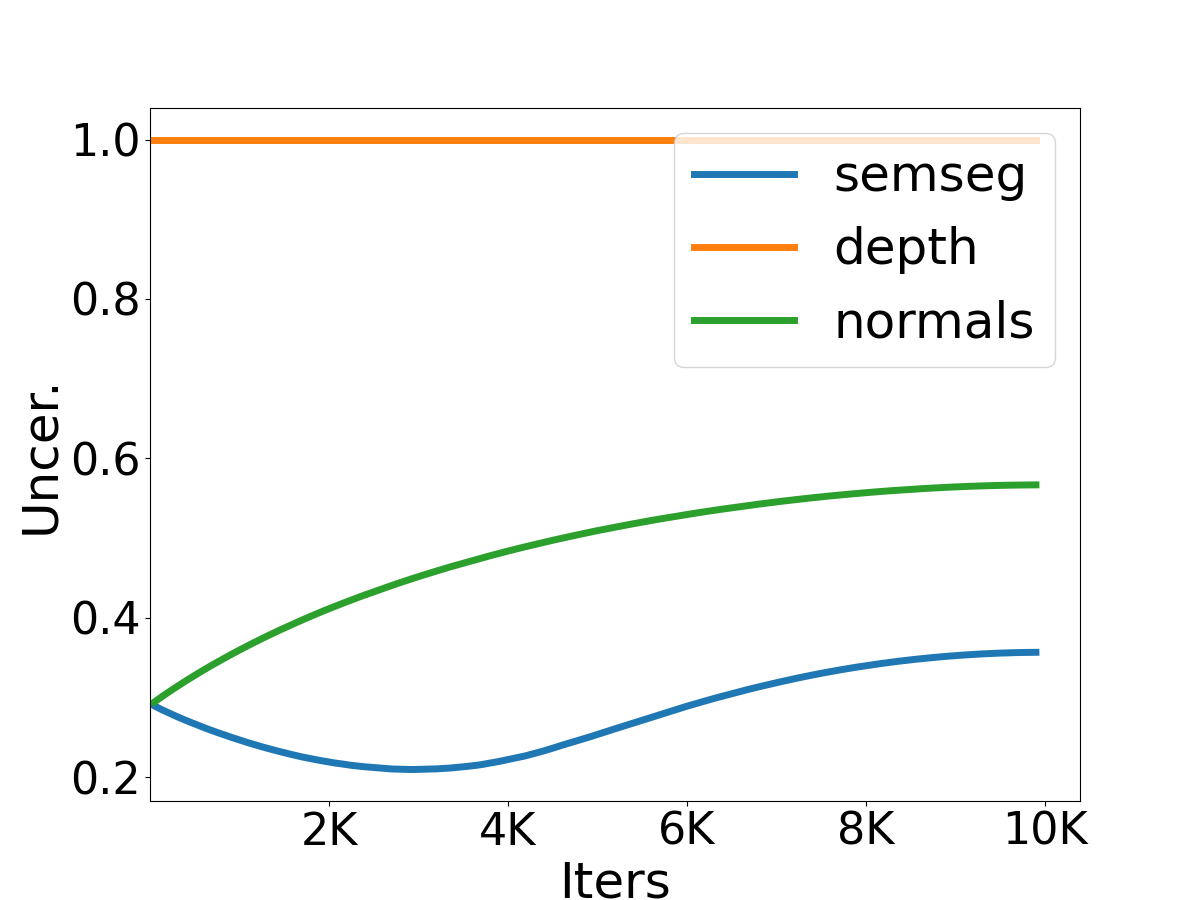}\\
        \end{minipage}%
    }%
    \subfigure{
        \begin{minipage}{0.3\linewidth}
            \centering
            \includegraphics[width=0.993\textwidth,height=1.2in]{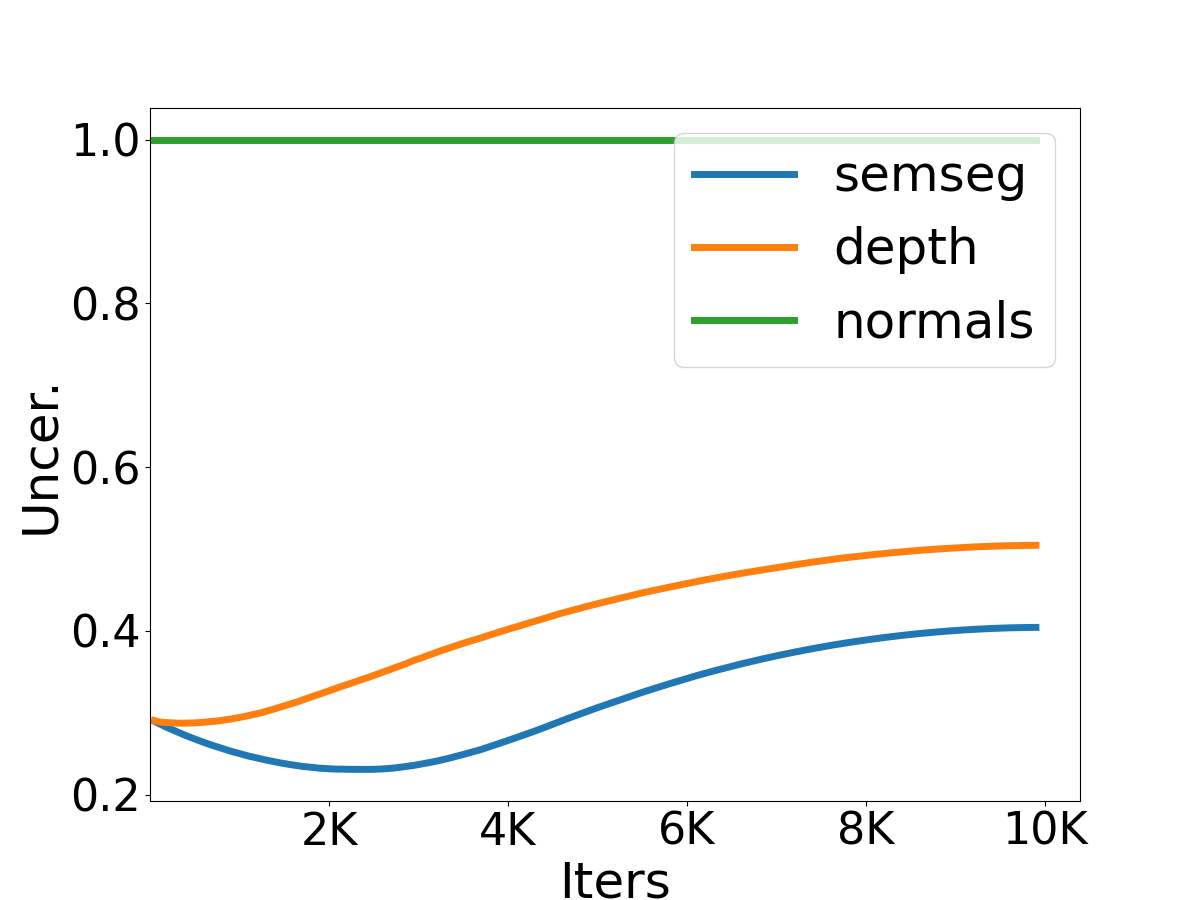}\\
        \end{minipage}%
    }%
    \vspace{-3mm}
    \setcounter{subfigure}{0}
    \subfigure[Seg.]{
        \rotatebox[origin=c]{90}{\centering\scriptsize{Gradient Norm}}
        \begin{minipage}{0.3\linewidth}
            \centering
            \includegraphics[width=0.993\textwidth,height=1.2in]{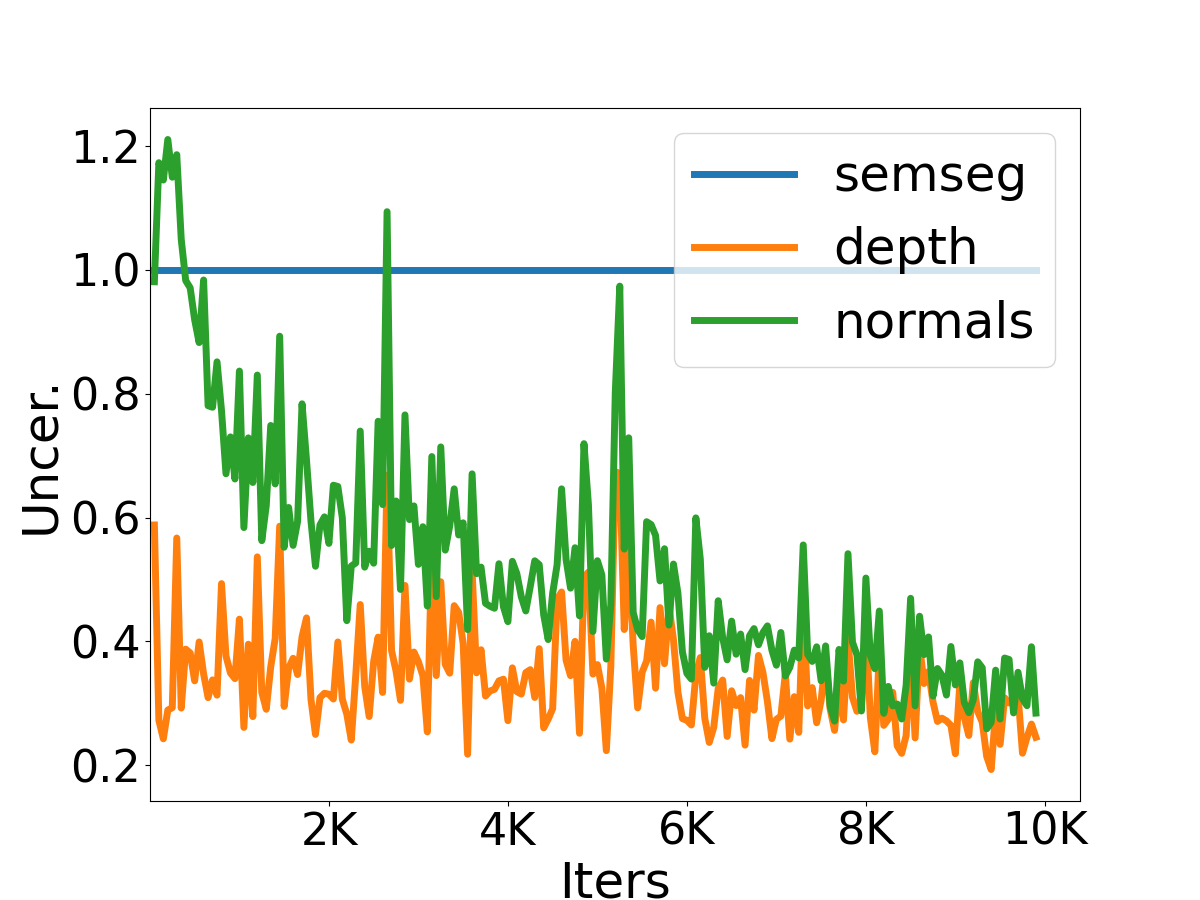}\\
            \vspace{1mm}
        \end{minipage}%
    }%
    \subfigure[depth]{
        \begin{minipage}{0.3\linewidth}
            \centering
            \includegraphics[width=0.993\textwidth,height=1.2in]{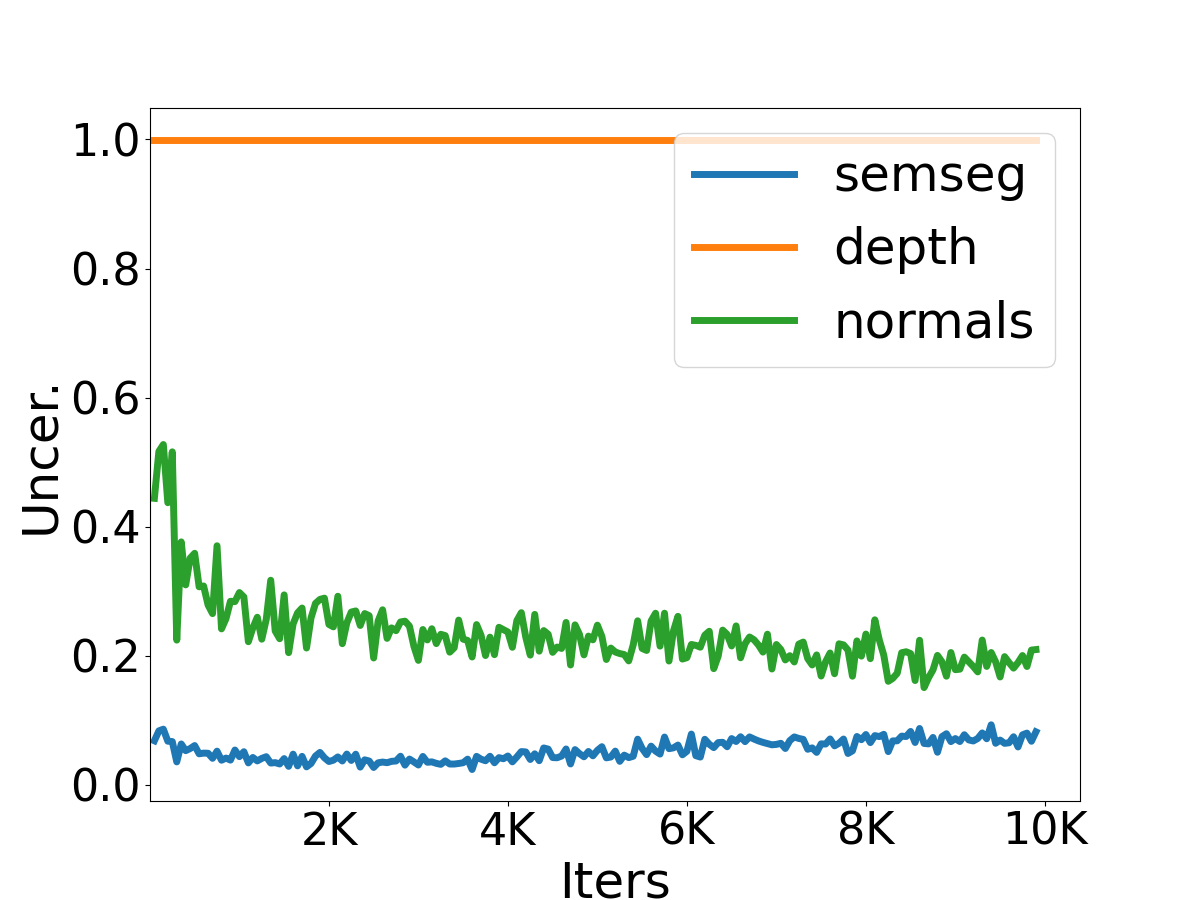}\\
            \vspace{1mm}
        \end{minipage}%
    }%
    \subfigure[Norm.]{
        \begin{minipage}{0.3\linewidth}
            \centering
            \includegraphics[width=0.993\textwidth,height=1.2in]{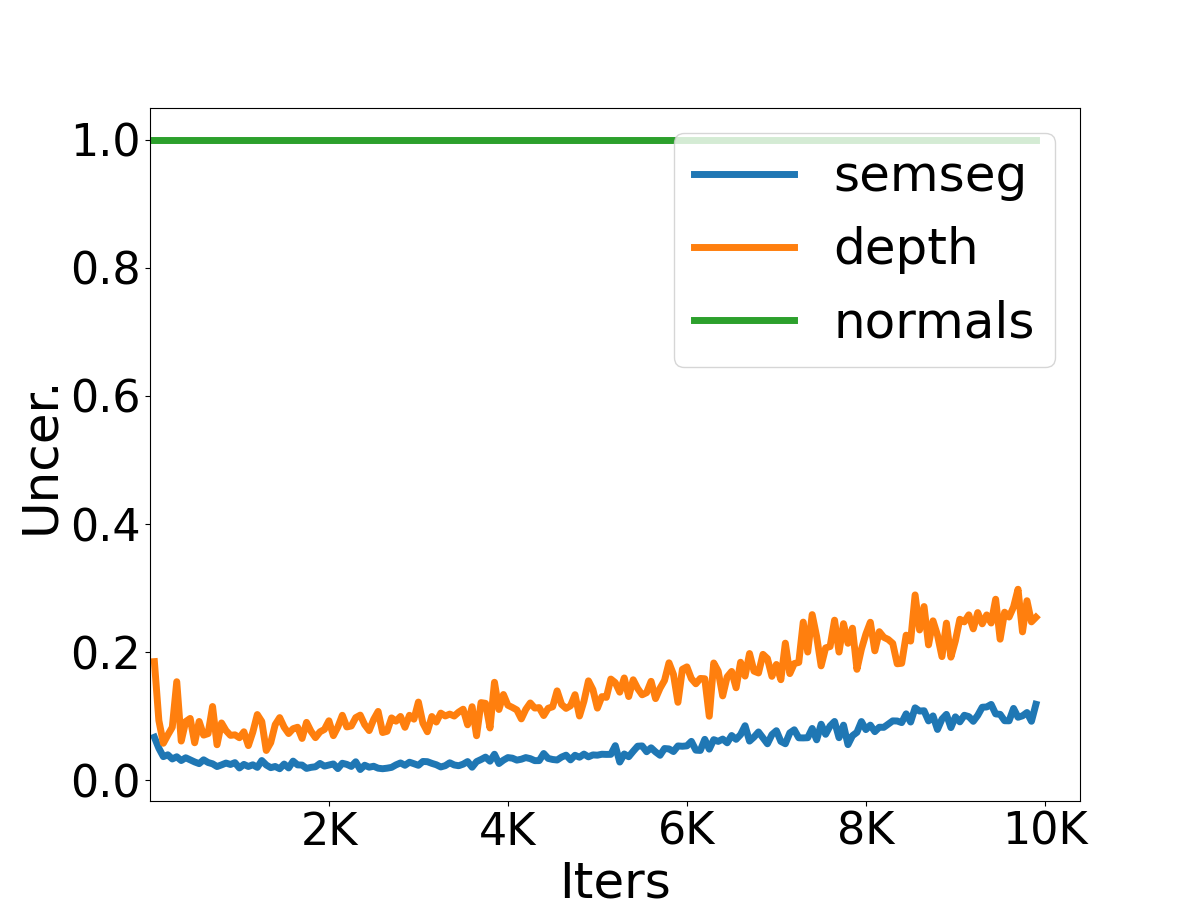}\\
            \vspace{1mm}
        \end{minipage}%
    }%
    % \vspace{-5pt}
    \vspace{-3mm}
    \caption{Visualization of weight changing over training on \textbf{NYUv2}. Each column shows the weight changes during training with semantic segmentation, depth estimation, and normals prediction as the primary tasks, respectively. The first row displays the weights trained in the Decoder stage, the second row shows the auxiliary weights~(Eq.~\ref{eq:ug_weight}) in the encoder stage, and the third row presents the change of gradient norms~(Eq.~\ref{eq:grad_norm}) for different tasks.}
    \label{fig:weight_changing}
    \vspace{-6mm}
\end{figure}

We show the weight changing over training on NYUv2 in Fig.~\ref{fig:weight_changing}. In the decoder stage, we equally optimize the task-specific decoders across both primary task and auxiliary tasks. As shown in the first row of Fig.~\ref{fig:weight_changing}, the weights are learnt via the uncertainty loss~(Eq.~\ref{eq:uw_loss}). The weight values for each task dynamically change based on the selection of the primary task and the progress of training. However, the relationship of task weights remains consistent.
Additionally, these tasks always achieve lower uncertainty~(higher weights) when they are treated as the primary tasks.

In the encoder stage, we weight the task-specific normalized gradient with Eq.~\ref{eq:weight_grad}, as shown in the middle row of Fig.~\ref{fig:weight_changing}. The weight of the primary task is always set to 1., while the weights of auxiliary tasks is related to their uncertainty. 
We also show $ Norm(\nabla_{z} L_t) $~(Eq.~\ref{eq:grad_norm}) in the third row of Fig.~\ref{fig:weight_changing}. With gradient weights, we balance the gradient norm of auxiliary tasks to ensure the dominance of the primary tasks.

\section{Conclusions}

This paper focuses on multi-task learning with auxiliary tasks. 
It empirically discovers that existing works often overlook the training of auxiliary tasks, which subsequently harms the performance of the primary tasks. 
To address this issue, a new multi-task framework is proposed that effectively balances the training of auxiliary tasks with the priorities of the primary tasks, enhancing overall performance.
The proposed method assigns uncertainty weights to ensure independent and effective training for task-specific decoders. 
It also uses estimated homoscedastic uncertainty and gradient norms as weights for auxiliary tasks to optimize encoder training for the primary tasks.
Experiments reveal that the proposed method achieves the best performance across state-of-the-art approaches in multi-task learning.
It can also be applied in the presence of certain noisy tasks. 
The method primarily trains improved decoders for auxiliary tasks.
However, the underexplored enhancement of auxiliary task learning on a shared-parameter encoder is a limitation. 
Additionally, the proposed method is based on an encoder-decoder network. 
Its applicability to other multi-task networks, such as PAP-Net, requires further investigation.

Future research may focus on advancing the joint optimization of auxiliary and primary tasks in networks with shared parameters, potentially employing approaches akin to null space learning theory. 
Furthermore, the applicability of this method to other multi-task networks can also be further explored.

\ifCLASSOPTIONcompsoc
\section*{Acknowledgments}
\else
\section*{Acknowledgment}
\fi
This work was supported in part by the National Key R\&D Program of China under Grant No. 2021ZD0112100. Guanglei Yang was also supported by the Postdoctoral Fellowship Program of CPSF under Grant Number GZC20233458. 
% I sincerely appreciate Dr. Chunpu Liu for the inspiration, discussions, and attention to code details provided during the completion of this work.

% \ifCLASSOPTIONcaptionsoff
%   \newpage
% \fi



%\bibliographystyle{IEEEtran}
%\begin{thebibliography}{1}
%\end{thebibliography}
% \clearpage
% \balance
\bibliographystyle{IEEEtran}
\bibliography{arxiv}


% \begin{IEEEbiography}[{\includegraphics[width=1in,height=1.25in,clip,keepaspectratio]{Images/elisa}}]
% {Elisa Ricci}
% received the PhD degree from the
% University of Perugia in 2008. She is an assistant
% professor at the University of Perugia and a
% researcher at Fondazione Bruno Kessler. She
% has since been a post-doctoral researcher at
% Idiap, Martigny, and Fondazione Bruno Kessler,
% Trento. She was also a visiting researcher at the
% University of Bristol. Her research interests are
% mainly in the areas of computer vision and
% machine learning. She is a member of the IEEE.
% \end{IEEEbiography}


%\vfill

% \enlargethispage{-5in}

\begin{IEEEbiography}[{\includegraphics[width=1in, height=1.25in, clip,keepaspectratio]{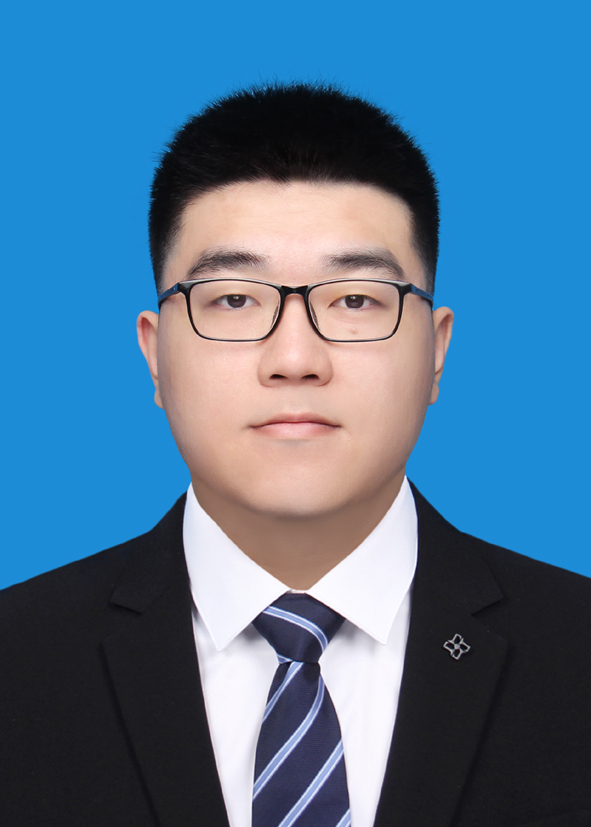}}]{Yuanze Li}
received a B.S. degree in computer science and technology from Harbin Institute of Technology (HIT), Harbin, China, in 2018. He is currently pursuing a Ph.D. degree at the Faculty of Computing, Harbin Institute of Technology (HIT), Harbin, China.  His research interests include multi-task learning and large multimodal models.
\end{IEEEbiography}

\begin{IEEEbiography}
[{\includegraphics[width=1in,height=1.25in,clip,keepaspectratio]{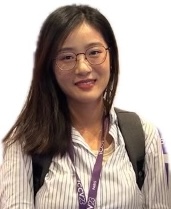}}]{Chun-Mei Feng}
	received the Ph.D. degree from Harbin Institute of Technology, Shenzhen. She is currently a research scientist in Institute of High Performance Computing (IHPC), Agency for Science, Technology and Research (A*STAR), Singapore. Her research interests include Federated learning, Medical image analysis, and Multi-modal foundation learning.
\end{IEEEbiography}

\begin{IEEEbiography}[{\includegraphics[width=1in, height=1.25in, clip,keepaspectratio]{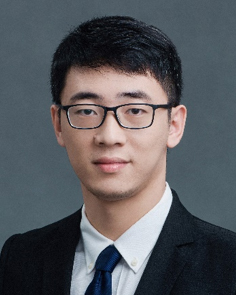}}]{Qilong Wang}
 (Member, IEEE) received the PhD degree from the School of Information and Communication Engineering, the Dalian University of Technology, China, in 2018. He currently is a professor at Tianjin University. His research interests include computer vision and deep learning, particularly deep architectures design and optimization based on high-order statistical modeling, multimodal large language model for specific tasks. He has published more than 40 academic papers in top conferences and referred journal including ICCV/CVPR/NeurIPS/ECCV and the IEEE Transactions on Pattern Analysis and Machine Intelligence/the IEEE Transactions on Image Processing/the IEEE Transactions on Circuits and Systems for Video Technology. He is served as an Area Chair of CVPR 2024.
\end{IEEEbiography}

\begin{IEEEbiography}[{\includegraphics[width=1in, height=1.25in, clip,keepaspectratio]{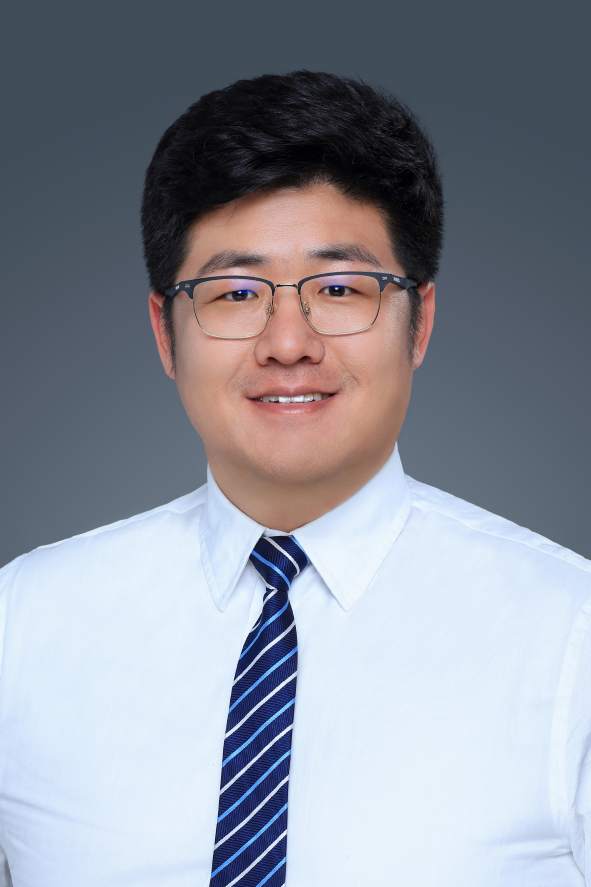}}]{Guanglei Yang}
is an Assistant Professor at the Faculty of Computer Science, Harbin Institute of Technology(HIT), China. He belongs to Integrative Intelligence and Learning(IIL) lab, where principal investigator is Wangmeng Zuo. He received my Ph.D. degree in 2023 from the School of Instrumentation Science and Engineering at HIT. He was also a joint Ph.D. student in 2020-2022 at University of Trento under the co-supervision of Prof. Elisa Ricci. His research interests mainly include the design and development of robust and scalable visual recognition systems, with a keen focus on their application in real-world scenarios. Central to my academic pursuit is the exploration of contrastive learning, unsupervised and semi-supervised learning methodologies, and domain adaptation techniques.
\end{IEEEbiography}

\begin{IEEEbiography}[{\includegraphics[width=1in, height=1.25in, clip,keepaspectratio]{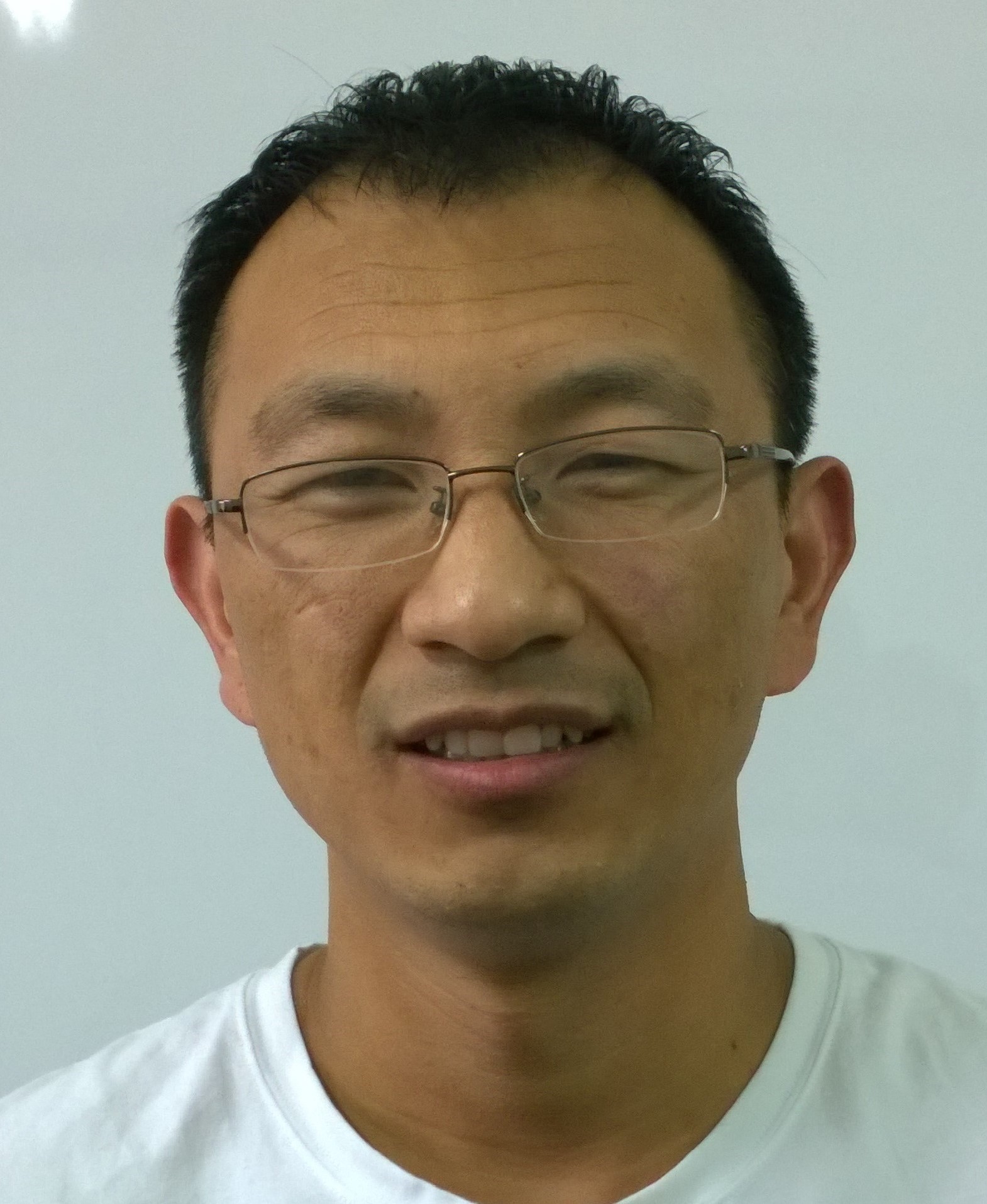}}]{Wangmeng Zuo}
received the Ph.D.degree in computer application technology from the Harbin Institute of Technology, Harbin, China, in 2007. He is currently a Professor in the Faculty of Computing, Harbin Institute of Technology. His current research interests include low level vision, image and video generation, multimodal understanding. He has published over 200 papers in top tier academic journals and conferences. He has served as an Associate Editor of IEEE Trans. Pattern Analysis and Machine Intelligence, IEEE Trans. Image Processing, and SCIENCE CHINA Information Sciences.
\end{IEEEbiography}
\enlargethispage{-5in}

% \begin{IEEEbiography}[{\includegraphics[width=1in, height=1.25in, clip,keepaspectratio]{images/biography/yangguanglei.pdf}}]{Guanglei Yang}
% received a B.S. degree in instrument science and technology from Harbin Institute of Technology (HIT), Harbin, China, in 2016. He is currently pursuing a Ph.D. degree at the School of Instrumentation Science and Engineering, Harbin Institute of Technology (HIT), Harbin, China. He has been working at the University of Trento as a visiting student since 2020. His research interests include domain adaptation and pixel-level prediction .
% \end{IEEEbiography}

\end{document}